\documentclass[sigconf, nonacm]{acmart}

\newcommand{\paperTitleFull}{Serving and Optimizing Machine Learning Workflows on Heterogeneous Infrastructures}

\usepackage{anyfontsize}
\usepackage[hyphens]{url}

\definecolor{linkcolor}{HTML}{647382}
\definecolor{citecolor}{HTML}{647382} %
\definecolor{urlcolor}{rgb}{0.4,0.2,0.2}
\definecolor{sqlcolor}{HTML}{965d67}
\definecolor{smtcolor}{HTML}{5d968c}
\definecolor{webblue}{rgb}{0,0,.7}
\definecolor{webgreen}{rgb}{0,.5,0}
\definecolor{webbrown}{rgb}{.6,0,0}
\usepackage{array}
\usepackage{tikz}
\usepackage{frame}
\usepackage{mathtools}
\usepackage{mdwlist}
\usepackage{listings}
\usepackage{amsmath}
\usepackage{float}
\usepackage[short]{optidef}
\usepackage{algpseudocode}
\usepackage{color}
\usepackage{hyperref}
\usepackage{multirow,mathrsfs}

\usepackage{amsmath}
\usepackage{subcaption}

\usepackage{endnotes,microtype,xspace,graphicx,fancyvrb,multirow}
\usepackage{supertabular,booktabs}
\usepackage{array,relsize}
\usepackage{fancyhdr}
\usepackage{enumitem}
\usepackage{balance}
\usepackage{booktabs}
\usepackage{pifont}
\usepackage{listings}
\usepackage{multirow}
\usepackage[scaled]{beramono}
\usepackage{tabularx}
\makeatletter
\newcommand\BeraMonottfamily{%
  \def\fvm@Scale{0.85}
  \fontfamily{fvm}\selectfont
}
\makeatother

\usepackage{aliascnt}

\usepackage[ruled,linesnumbered]{algorithm2e}
\usepackage{setspace}
\usepackage{enumitem}
\usepackage{ hyperref }
\usepackage{dsfont}

\SetAlFnt{\small}
\SetAlCapFnt{\small}
\SetAlCapNameFnt{\small}

\usepackage{semantic}
 
\usepackage{stmaryrd}
\usepackage{ltablex}
\usepackage{mathtools}
\usepackage{adjustbox}
\usepackage[capitalize,noabbrev,nameinlink]{cleveref}

\crefname{lstlisting}{listing}{listings}
\Crefname{lstlisting}{Listing}{Listings}

\newcommand{\cut}[1]{}
\newcommand{\name}{{\sf JellyBean}\xspace}

\newcommand{\xref}[1]{\S\ref{#1}}

\title{\paperTitleFull} 
\author{Yongji Wu}
\affiliation{%
  \institution{Duke University}
}
\email{yongji.wu769@duke.edu}

\author{Matthew Lentz}
\affiliation{%
  \institution{Duke University}
}
\email{mlentz@cs.duke.edu}

\author{Danyang Zhuo}
\affiliation{%
  \institution{Duke University}
}
\email{danyang@cs.duke.edu}

\author{Yao Lu}
\affiliation{%
  \institution{Microsoft Research}
}
\email{luyao@microsoft.com}

\begin{document}

\newcommand{\mail}[1]{\href{mailto:#1}{#1}}

\begin{abstract}
With the advent of ubiquitous deployment of smart devices and the Internet of Things, data sources for machine learning inference have increasingly moved to the edge of the network. Existing machine learning inference platforms typically assume a homogeneous infrastructure and do not take into account the more complex and tiered computing infrastructure that includes edge devices, local hubs, edge datacenters, and cloud datacenters. On the other hand, recent AutoML  efforts have provided viable solutions for model compression, pruning and quantization for heterogeneous environments; for a machine learning model, now we may easily find or even generate a series of models with different tradeoffs between accuracy and efficiency.

We design and implement {\name}, a system for serving and optimizing machine learning inference workflows on heterogeneous infrastructures. Given service-level objectives (e.g., throughput, accuracy), {\name} picks the most cost-efficient models that meet the accuracy target and decides how to deploy them across different tiers of infrastructures. Evaluations show that {\name} reduces the total serving cost of visual question answering by up to 58\%, and vehicle tracking from the NVIDIA AI City Challenge by up to 36\% compared with state-of-the-art model selection and worker assignment solutions. {\name} also outperforms prior ML serving systems (e.g., Spark on the cloud) up to 5x in serving costs.
\end{abstract}

\maketitle

\begin{figure}[t]
\centering
\includegraphics[width=.95\linewidth]{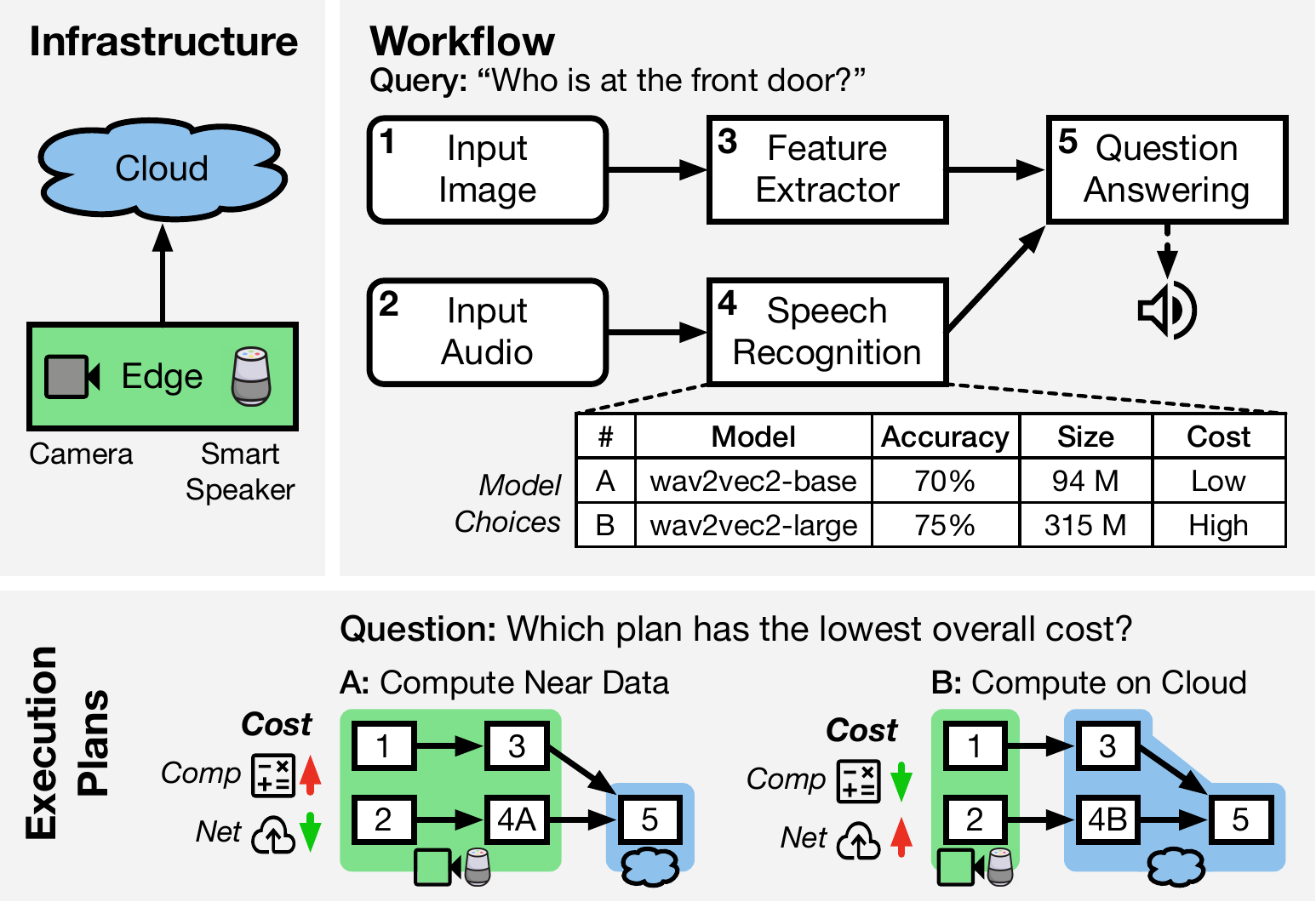}
\caption{An example ML workflow of Visual Query Answering (VQA) on heterogeneous infrastructures. Execution plans vary by model selection and worker assignment for each operator (in the boxes) and result in different serving costs, i.e., compute and network.}
\label{fig:q1}
\end{figure}

\section{Introduction}

There is a growing complexity in machine learning (ML) inference workloads both
in terms of the workloads themselves as well as the computing and networking
infrastructures.
These workloads often involve multiple ML operators
that together form a larger \emph{ML workflow}$^1$; each can be a directed acyclic graph (DAG) of ML or relational operators. For each ML operator, there are often choices of models (e.g., YOLO~\cite{redmon2016you}, Faster R-CNN~\cite{ren2015faster}) or the same model architectures with different hyperparameters (e.g., number of layers, neural network size, choice of activation functions); 
inputs to the ML workflows are often collected by sensors deployed at the edge, including video cameras and an ever-expanding array of Internet-of-Things (IoT) devices.
These devices may have varying on-board compute~\cite{dnncam} and are connected to more powerful edge-local and cloud computing services over the network.  

\vspace{0.05in}
\noindent\textbf{Example}. Consider the visual question answering (VQA) workflow in Figure~\ref{fig:q1} for the query ``\emph{Who is at the front door?}''. The workflow uses multiple ML models for feature extraction and model inference. The infrastructure includes edge devices (e.g., cameras) as well as cloud datacenters. To deploy ML workflows on heterogeneous infrastructures, the following decisions must be made:

\begin{itemize} [leftmargin=0.2in]

\item \emph{Model selection}. With advances of AutoML and model compression techniques (e.g., pruning, quantization~\cite{jacob2018quantization,sanh2020movement}),  each ML operator in the workflow\footnote{Workflows are generated using a standard parser~\cite{johnson1975yacc} or a natural language interface~\cite{kim2020natural}, which are orthogonal to this paper. See \xref{sec:overview} for more details.} can use various structures or hyperparameters; e.g., the speech recognition operator in Figure~\ref{fig:q1} may use the \texttt{base} variant for a faster execution or \texttt{large} for a better accuracy. To provide a viable accuracy-efficiency tradeoff, picking individual models in the workflow is non-trivial.

\item \emph{Worker assignment}. Each operator must be assigned to a worker for execution. Figure~\ref{fig:q1} demonstrates two execution plans - placing compute near the data source to reduce communication, or moving them to the cloud to take advantage of more powerful (and likely cheaper) compute resources. Choosing an appropriate plan depends on resource availability and costs.

\end{itemize}

\vspace{0.05in}
\noindent\textbf{Goals, challenges, and prior solutions}. Given the ML workflow, resource availability, input throughput, and target accuracy, we aim to optimize the total serving costs that consist of both compute and networking. It is easy to see that model selection and worker assignment formulate a complex search space. 

Current ML serving platforms such as Ray~\cite{moritz2018ray}, Clipper~\cite{crankshaw2017clipper},
PyTorch~\cite{paszke2019pytorch}, and Spark~\cite{zaharia2012resilient} focused on
homogeneous infrastructures (namely cloud datacenter environments).
Unfortunately, ignoring resource heterogeneity (e.g., compute, network) often leads to
sub-optimal deployments and even feasibility issues given the infrastructure constraints
 (e.g., on links shared among many high data rate sensors like video cameras).
Some prior systems solve this problem in an ad-hoc manner for specific ML workflows, individual models, and fixed infrastructure configurations~\cite{crankshaw2017clipper,jiang2018chameleon,li2020reducto,zhang2019mark,shen2019nexus,romero2019infaas,abadi2016tensorflow,paszke2019pytorch}. Chameleon~\cite{jiang2018chameleon} considers video analytics with one model on a single GPU; Nexus~\cite{shen2019nexus} considers workflows on a homogeneous GPU cluster with no model choices.
To our best knowledge, there is currently no off-the-shelf system that optimizes the deployments of ML workflows on heterogeneous infrastructures.
As a result, users determine how to best deploy ML workflows often manually.

\vspace{0.05in}
\noindent\textbf{{\name} ideas and approaches}.
We address some initial problems for optimizing ML
workflows on heterogeneous infrastructures, and 
propose a system {\name}. Given an ML workflow and specifications of the infrastructures, the {\name} optimizer quickly finds a cost-efficient execution plan with model choices and worker assignments using the following insights:

First, we formulate the problem within a cost-based optimization~\cite{chaudhuri1998overview},
minimizing the compute and network costs while meeting the input throughput
and accuracy constraints.
However, optimizing ML workflows
poses novel challenges.
In the above example, even though we can profile the
accuracy and cost for every single model, understanding how different models
interact for estimating the overall query accuracy is non-trivial.
We leverage a simple but effective model profiling strategy that relies on sampled measurements of interactions between models to estimate query accuracy.

Next, simultaneously solving for optimal model choices and worker assignment is NP-hard
and results in an exponentially large search space.
We reduce the search space and provide a fast query optimization by 
(1) making two simplifying assumptions that hold for many real-world scenarios,
and (2) identifying key parts that are amenable to greedy approaches.
Our evaluations in \xref{s:eval} show the efficacy in practice.

Lastly, to serve and optimize ML workflows on heterogeneous infrastructures, a flexible runtime is critical such that \emph{the optimizer may explore plans in which models are placed in different workers and locations}. Due to the lack of an existing system to support this, we implemented the {\name} processor upon Naiad~\cite{murray2013naiad} and Timely Dataflow~\cite{timelydataflow}, modifying them to enable operator-level parallelism -- each worker may handle a subset of the overall workflow. 
Such a processor and optimizer decide \emph{where to run what}; for \emph{how} to execute each individual operator, we use a containerized runtime with  virtualization and ML compiler techniques~\cite{nvidiavirtualization, chen2018tvm} such that {\name} can cope with the infrastructure heterogeneity.

We performed experiments on various real-world use cases, 
including the Nvidia AI City Challenge~\cite{aicity} and Visual Question Answering (VQA)~\cite{antol2015vqa}. Compared with running the
ML workflows (1) with all data pushing to the cloud, (2) with all
computations staying on the edge, and (3) with optimizations carried out by several worker assignment heuristics, better assigning different parts of the workload to different
infrastructure is significantly more effective. We also compared with a few
recent ML serving platforms and found that {\name} is significantly better
to achieve the user-specified query-level goal. {\name} achieves close to equivalent performance compared with an exhaustive brute force search on a small-scale experiment and can still generate efficient physical plans when brute force is infeasible on larger-scale experiments. {\name} can reduce the total serving cost for VQA by up to 58.1\%, and for vehicle tracking in AICity by up to 36.3\% compared to the best baselines. {\name} also outperforms prior ML serving systems (e.g., Spark on the cloud) up to 5x in total serving costs.
 
\vspace{0.05in}
\noindent\textbf{Contributions} of this paper can be summarized as follow:
\begin{itemize}[leftmargin=0.2in]

	\item The {\name} optimizer to derive highly effective execution plans for complex ML workflows on heterogeneous infrastructures given the infrastructure constraints and model choices. 

	\item A flexible \name processor based on a graph dataflow to execute
		the optimized plans and enable operator-level parallelism on heterogeneous 
		infrastructures. 

	\item Evaluations on real datasets show significant performance
		improvements over state-of-the-art ML serving platforms as well as
		running the workflows using heuristics.

\end{itemize} 

\section{Background}\label{sec:bg}
We discuss some popular ML workflows, followed by the challenges of running them across heterogeneous infrastructures.

\vspace{0.05in}
\noindent\textbf{ML Workflows}
There are many other use cases of ML queries for intelligent Internet of Things (IoT). In addition to the VQA query introduced above, we name a few interesting scenarios for instance: 

\begin{figure}[t]
\centering
\includegraphics[width=.95\linewidth]{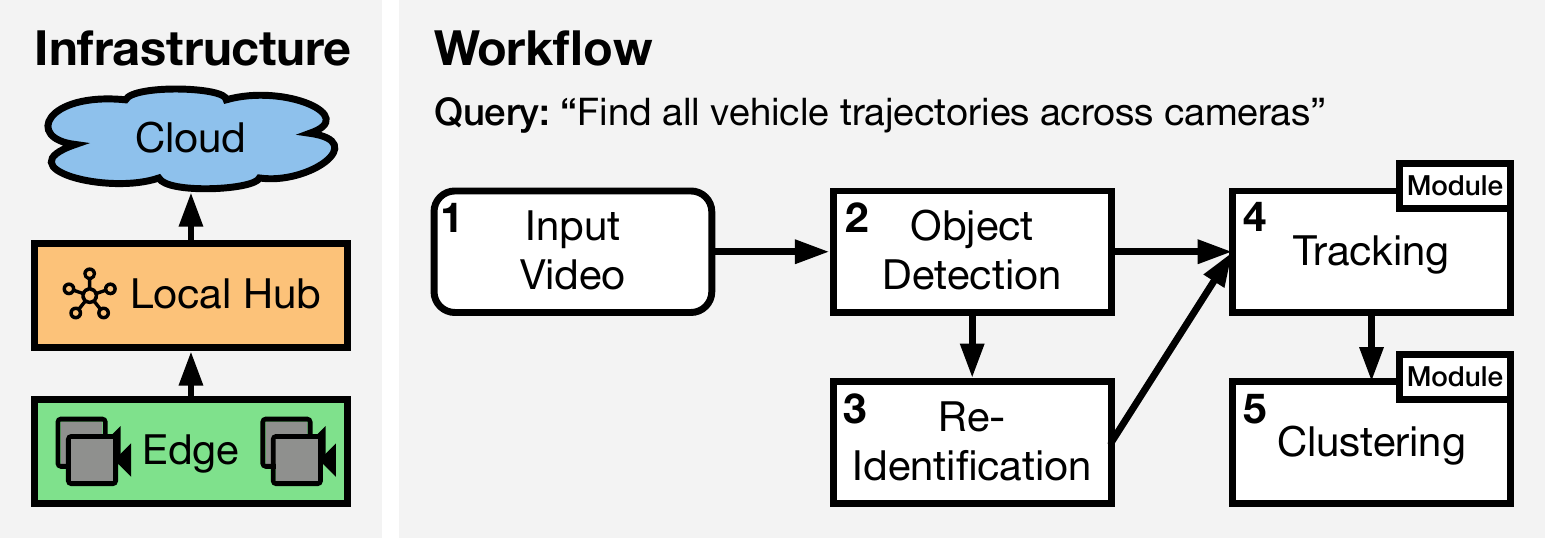}
\caption{NVIDIA AI City Challenge for Vehicle Tracking. Some pair-wise operators are omitted for simplicity.}
\label{fig:ai_city}
\end{figure}

\begin{itemize} [leftmargin=0.2in]
    \item \textit{NVIDIA AI City challenge:} Tracking vehicles across neighboring intersections is an important ML query that allows people to understand and improve transportation efficiency~\cite{aicity}. The workflow is shown in Figure~\ref{fig:ai_city} with video inputs from multiple cameras of neighboring traffic intersections. It first detects objects on each individual video stream, and then performs an object re-identification (ReID) step to extract key features per detected car. A tracking module is used to find car traces in each video stream, followed by a clustering module to trace cars across different video streams.
    \item \textit{Wearable health:} detecting anomalous heart signals. 
    \item \textit{Personal assistant:} answering complex human voice commands using Internet data.  
\end{itemize}

One common characteristic is that they all rely on a set of loosely-coupled operators (i.e., operators that do not share global states but only depend on prior outputs), each of which uses an ML model or a traditional data processing module; e.g., a model to tokenize the text or relational operators such as reduce and join~\cite{bird2006nltk,manning2014stanford}. The output of a previous operator is the input of the next, therefore formulating a workflow or logic plan in directed acyclic compute graph (DAG).
Breaking down an ML query into workflows that consist of independent operators has been highly leveraged in prior research and production~\cite{shen2019nexus,lu2018accelerating}. Doing so promotes the reuse of trained models and operators to ease the development of the serving system as well as to boost performance due to shared computations~\cite{wolf2020transformers,le2019multimodal,ben2017mutan,lu2016optasia}; each module also can be improved independently to accelerate the application development.

\begin{figure}[t]
\centering
\includegraphics[width=.7\linewidth]{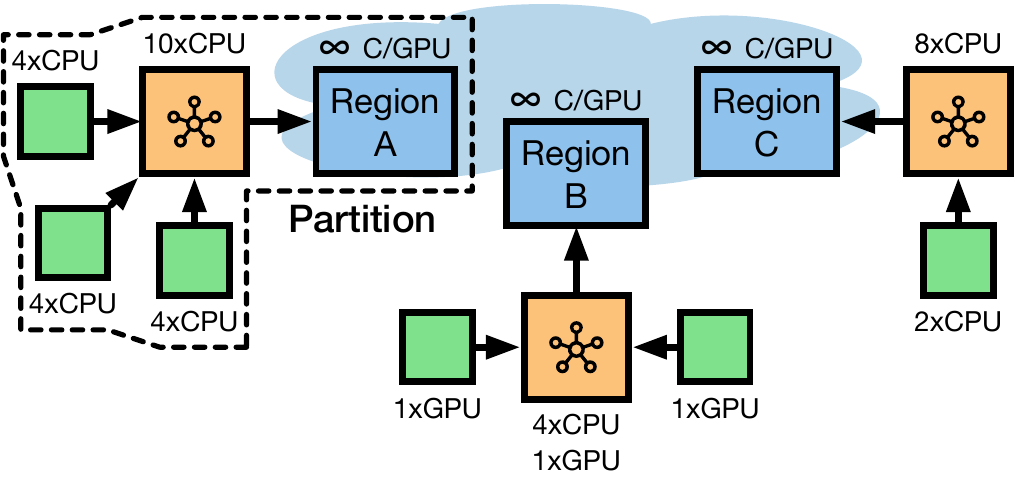}
\caption{Deploying ML workflows on heterogeneous infrastructure requires designing physical plans for different partitions.} 
\label{fig:partition}
\end{figure}

\begin{figure*}[ht]
\centering
\includegraphics[width=0.96\linewidth]{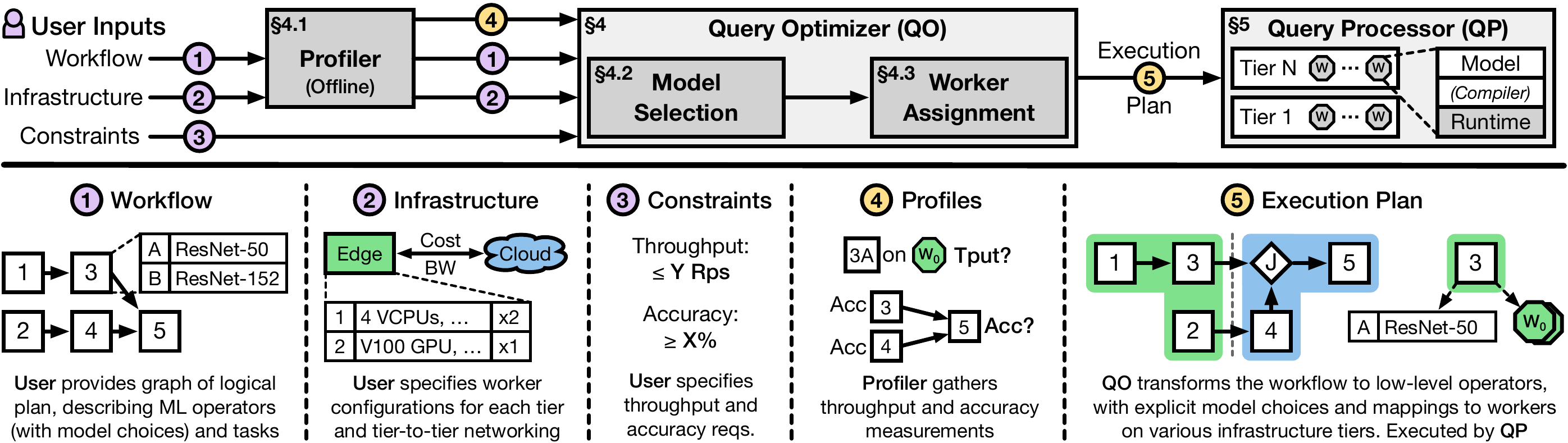}
\caption{Overview of the \name architecture. There are three main components: Profiler, Query Optimizer (QO), and Query Processor (QP).}
\label{fig:sys}
\end{figure*}

\vspace{0.05in}
\noindent\textbf{Serving ML on Heterogeneous Infrastructures}.
The above examples also show that many application scenarios have input data injected from edge devices. To deploy ML workflows upon these inputs, one way is to put them in cloud datacenters. Clearly, this can often be suboptimal since raw inputs (e.g., images and videos) can be large and data movement can be costly. 

Moving compute to near the data source is a well-known topic in the big-data systems literature and has been proven to be effective in many use cases~\cite{huang2019yugong,pu2015low}. However, today developers still have to hard code or manually tune the physical execution plans for each ML workflow depending on the amount of resources on the edge and costs of various types of resources~\cite{jiang2018chameleon,merenda2020edge,kang2017neurosurgeon}. We believe this manual approach cannot scale with the rapid development of edge data centers and IoT devices. 

Figure \ref{fig:partition} shows a cloud with three regional datacenters, several local hubs, and edge compute devices. A different execution plan is needed for each partition. For example, different local hubs can have different numbers and types of workers.
The cost of running models at different locations can also be different, depending on the cloud region and the resource availability at local hubs. 
We use the term \textit{partition} to denote the tiered infrastructure where different locations within a tier have similar resources. If a partition contains multiple local hubs, they must have similar worker configurations. 
\name can be used to generate a physical plan per partition.

\begin{table}[t]
	\caption{Comparing current ML systems. MS: model selection. WA: worker assignment.}\label{tab:mlsys}
	\begin{small}
		\centering
		\begin{tabular}{@{}c|c|cc|c|cc@{}}
		\toprule
		\multirow{2}{*}{System} & \multirow{2}{*}{Parallelism} & \multicolumn{2}{c|}{QO} & \multirow{2}{*}{Usage} & \multicolumn{2}{c}{Heterogeneity} \\
		 &  & MS & WA & & Worker & Infra.\\
		\midrule
		PyTorch~\cite{paszke2019pytorch} & Data & $\times$ & $\times$ & Both & $\times$ & $\times$ \\
		TF~\cite{abadi2016tensorflow} & Data & $\times$ & $\times$ & Both & $\times$ & $\times$ \\
		Spark~\cite{zaharia2012resilient} & Data & $\times$ & $\times$ & Infer & $\times$ & $\times$ \\
		Clipper~\cite{crankshaw2017clipper} & Data & $\times$ & $\times$ & Infer & $\times$ & $\times$ \\
		Ray~\cite{moritz2018ray} & Data, Model & $\times$ & $\times$ & Both & $\times$ & $\times$ \\
		Optasia~\cite{lu2016optasia} & Data, Op & $\times$ & $\checkmark$ & Infer & $\times$ & $\times$ \\
		Pathways~\cite{barham2022pathways} & Data, Model & $\times$ & $\times$ & Train & $\checkmark$ & $\times$ \\
		Llama~\cite{romero2021llama} & Data, Op & $\times$ & $\checkmark$ & Infer & $\checkmark$ & $\times$ \\
		Scrooge~\cite{hu2021scrooge} & Data, Op & $\times$ & $\checkmark$ & Infer & $\checkmark$ & $\times$ \\
		\midrule
		\name (Ours) & Data, Op & $\checkmark$ & $\checkmark$ & Infer & $\checkmark$ & $\checkmark$  \\
		\bottomrule
		\end{tabular}
	\end{small}
\end{table}  

\vspace{0.05in}
\noindent\textbf{ML Serving Systems}. In order to partially move the ML workflow to the edge devices, besides being able to break it into modules or operators, another necessary condition is a serving system that supports operator-level parallelism on heterogeneous infrastructures. Prior ML systems focused on data, model (i.e., breaking large DNNs into operators) and operator (i.e., breaking workflows into operators) parallelism on homogeneous infrastructures~\cite{crankshaw2017clipper,paszke2019pytorch,lu2016optasia,moritz2018ray}, or on heterogeneous workers within a datacenter~\cite{romero2021llama,hu2021scrooge,barham2022pathways}. We demonstrate a qualitative comparison in Table~\ref{tab:mlsys}.  Recently, Google's Pathways~\cite{barham2022pathways} has started to investigate operator-level parallelism for training large deep neural networks with hybrid cloud infrastructures of CPUs, GPUs, and TPUs. There still lacks an off-the-shelf system for serving and optimizing ML workflows with model choices on heterogeneous and especially IoT infrastructures.
We provide a more detailed comparison with related systems in \xref{sec:related}.

\section{Overview}\label{sec:overview}
We discuss our \name design and scope in this section.

\vspace{0.05in}
\noindent\textbf{System scope}.
{\name} aims at serving and optimizing ML inference workloads that can be decomposed into multiple operators deployed on heterogeneous infrastructures.
We target infrastructures that exhibit resource heterogeneity across tiers and resource  homogeneity  within a tier.
\name operates over an infrastructure configuration that describes a single partition of a potentially larger infrastructure.
The optimization takes into account input throughput, resource cost,  availability and efficiency, and targets scenarios in which compute and communication are important factors in the total serving cost. The {\name} processor provides a flexible runtime and decouples resource heterogeneity using a containerized runtime with virtualization and ML compilers, hence targeting a wide spectrum of edge and cloud devices.

\vspace{0.05in}
\noindent\textbf{System overview}.
In Figure~\ref{fig:sys}, we present an overview of our {\name} system architecture and the workflow for processing an ML workflow. There are two main components: the query optimizer (QO) and the query processor (QP). The query optimizer generates an execution plan for the ML workflow, while the query processor runs the execution plan across heterogeneous infrastructure. 

{\name} takes the following inputs:

\begin{itemize}[leftmargin=0.2in]

\item \emph{Workflow}. Each input workflow is a directed acyclic graph (DAG) with compute operators on the nodes and input-output relationships between operators on the edges. The operators can be ML models or relational operations. Declarative queries can be parsed into workflows~\cite{johnson1975yacc,kim2020natural} as is done in~\cite{lu2018accelerating,kang2018blazeit}.

\item \emph{Model choices for each ML operator}. Each ML operator may use different models with the same semantics but different structures or hyperparameters. These models have different accuracy and cost profiles. \name may profile these models offline if necessary.

\item \emph{Infrastructure specifications}. We consider infrastructures that consist of heterogeneous resources (i.e., compute, storage and networking) in multiple tiers - each tier is a group of efficiently interconnected resources that share common specifications.

\item \emph{Input throughput and target accuracy}. Users provide a target accuracy on the query output; meanwhile, {\name} must keep up with the input throughput.  The target accuracy restricts the model selection to generate a low-cost physical plan.

\end{itemize}

Our query optimizer generates the physical plan in two steps. First, it selects models that satisfy the target accuracy with the least costs (\xref{sec:model}). Here we do not have worker assignments yet, so the exact costs of deploying the selected models are unknown. We approximate  the costs based on the characteristics of the models (e.g., model sizes, the latency of inference on a standard CPU/GPU) and use beam search to select the best K configurations. Each configuration includes the model selection for all models in the workflow.

The second step is to determine the worker assignment (\xref{sec:assign}). We again use a beam search method. We progressively determine the worker assignment by choosing a set of workers for each operator to achieve the lowest compute and networking costs. More than one workers may be assigned to an operator to consolidate the costs. The best worker assignment is derived then for each of the K configurations and choose the best physical execution plan for both model selections and worker assignment.

The {\name} processor is a distributed query processing engine upon Naiad~\cite{murray2013naiad} and Timely dataflow~\cite{timelydataflow} to provide a low-overhead dataflow abstraction. However, Naiad and Timely Dataflow use a homogeneous datacenter setup with data parallelism only. {\name} augmented their codebase to incorporate operator-level parallelism, allowing different workers to run different portions of the workflow. Each worker leverages a containerized runtime with virtualization or ML compilers~\cite{chen2018tvm,tensorrt} to offset heterogeneity (\xref{sec:qp}).

\begin{table}[t]
	\caption{\label{t:param} Set of common notations used in our description.}
	\begin{footnotesize}
		\centering
		\begin{tabular}{@{}ccl}
		    \toprule
			& {\bf Notation} & {\bf Definition}\\
			\midrule
            \parbox[t]{1.5mm}{\multirow{8}{*}{\rotatebox[origin=c]{90}{User Input}}}
			& $G$ & Graph of logical plan $(G = \langle V, E, M, m \rangle)$ \\
			& & \hspace{1.5em} Vertices $V$, Edges $E$, Models $M$ \\
			& & \hspace{1.5em} Model Choices $m: V \rightarrow \mathcal{P}(M)$ \\
			& $I$ & Set of infrastructure tiers \\
			& $W, W_i$ & Set of workers overall [or for tier $i \in I$] \\
			& $C_B$ & Worker-to-worker communication cost $(C_B: WxW \rightarrow \frac{\$}{\text{byte}})$ \\
			& $T, T_v$ & Input throughput overall [or for node $v \in V$] \\
			& $A$ & Target overall accuracy \\
			\midrule
            \parbox[t]{1.5mm}{\multirow{3}{*}{\rotatebox[origin=c]{90}{Profiler}}}
            & $C_C$ & Unit compute cost for model on worker $(C_C: MxW \rightarrow \$)$ \\
            & $t_u^w$ & Throughput for model $u$ on worker $w$ \\
            & $r$ & Unit input size at $v$ from $u$ $(r: VxV \rightarrow \text{byte})$ \\
			\midrule
            \parbox[t]{1.5mm}{\multirow{2}{*}{\rotatebox[origin=c]{90}{QO}}}
			& $s$ & Model selection $(s: V \rightarrow M)$ \\
			& $a$ & Worker assignment $(a: V \rightarrow \mathcal{P}(W))$ \\
		    \bottomrule
		\end{tabular}
	\end{footnotesize}
\end{table}

\section{Query optimizer} \label{sec:qo}

\subsection{Problem formulation} \label{sec:qoprob}
We consider our infrastructure to be composed of a number of workers with diverse computing capability distributed across multiple tiers (e.g., edge, hub, and cloud). Data sources are located on the lowest tier (i.e., $W_1$), often with some limited compute resources. Workers on higher tiers tend to have more computing capability but are far away from the data sources. We assume a set of workers $W$, which are partitioned into $|I|$ tiers. 

Let the input of our optimizer be a logical plan graph $G$ in which each node $v \in V$ corresponds to an ML or regular relational operator.
For each ML operator, the user specifies a list of candidate models $m(v)$, each having a different accuracy and runtime performance. These models can be developed independently or can be variants of other well-known models through quantization~\cite{esser2019learned,jacob2018quantization}, distillation~\cite{gou2021knowledge}, and pruning~\cite{liu2017learning,sanh2020movement}. \xref{sec:discussion} discusses techniques to generate a diverse set of model choices. A model's accuracy and performance can be either provided by the user or profiled by \name.
We use $s(v)$ to denote the model choice for $v$. Meanwhile, we assign for each logical operator $v$ a list of workers $a(v)$ in the heterogeneous infrastructure. The infrastructure specification contains sets of each type of worker a tier has, the cost of each type of worker, and the communication costs between different tiers. Note here our formulation only considers a single partition. This is because each partition (shown in Figure \ref{fig:partition}) requires a different physical execution plan. 
Table~\ref{t:param} illustrates the notations used in this paper as well as inputs to our query optimizer.
Note that the compute and communication costs here as unit monetary costs; the former is the hourly price per worker, and the latter is based on network traffic (i.e., data movement on the DAG edges).

We aim to solve worker assignment\footnote{We note that assigning for each operator a list of workers is equivalent to picking the model to execute for each worker.} $a:V\rightarrow \mathcal{P}(W)$ and model selection $s: V \rightarrow M$ simultaneously, such that the overall query accuracy ($acc$) is beyond a user-specified target $A$, and that the system's throughput ($t_{v_{out}}^{a(v_{out})}$ at the output node $v_{out}$) is no less than a target $T$. 
We describe our target cost function and our query optimization as:

\begin{align}\label{eq:problem}
\arg\min_{\substack{a,s}} \enspace &
\sum_{v\in V} \sum_{w\in a(v)} C_c(s(v), w) \enspace +
\\
& \sum_{(u,v) \in E} \sum_{\substack{(w_u, w_v) \in \\ a(u) \times a(v)}} C_B(w_u, w_v) R(u, v) \nonumber
\\
s.t. \enspace acc \ge A, \enspace t_{v_{out}}^{a(v_{out})} \ge T, \nonumber
\end{align}

\noindent where $R(u,v)$ denotes the consumed network bandwidth from $u$ to $v$.
The formulation above minimizes the ML workflow's combined compute (first term) and networking  (second term) costs 
and is NP-hard, because the sub-problem of solving only the worker assignment is already a combinatorial optimization that can be reduced to a binary knapsack problem (which is NP-complete~\cite{garey1979computers}).

\vspace{0.05in}\noindent\textbf{Assumptions}.
We make two assumptions in our optimization to reduce the problem complexity without losing generality, as these assumptions hold for many realistic use cases: 

\begin{itemize}[leftmargin=0.2in]

\item \texttt{A1}: We assume that communication costs $C_B(w_1, w_2)$ to have the following properties: 1) set to 0 if $w_1$ and $w_2$ are on the same infrastructure tier and are in the same location, 2) otherwise set to a positive value. This is common in many use cases, as workers in the same tier either do not inter-communicate (e.g., among edge devices at different locations) or use high-speed networking (e.g., among datacenter nodes) with negligible costs.

\item \texttt{A2}: We assume that all workers only communicate with peers in the same infrastructure tier or any higher tier, thus making information flow in one direction\footnote{We note that the final result from the workflow may be transferred back to the lowest tier (e.g., user's device), but we don't model this}. This assumption implies that for all edges $(u, v) \in E$, the set of workers $a(v)$ are all on tiers greater than or equal to the highest tier of any worker in $a(u)$. This is reflected in $C_B$ by values of $+\infty$ for pairs of workers that violate this one-way flow assumption.

\end{itemize}

\vspace{0.05in}
\noindent\textbf{Model Profiling}.
\name needs to understand the impact of selecting different models on accuracy and throughput in order to meet the constraints specified by the user for the overall workflow.
While users can optionally specify the accuracy and performance of models for different infrastructure workers, \name supports automatic profiling using validation datasets provided by the user. 
If a worker cannot run a particular model (e.g., model requires a GPU but the worker is CPU-only), we set both the accuracy and the throughput to be zero.
Otherwise, \name measures the runtime performance in terms of the  throughput for the model on every worker type in the infrastructure. Note that we use the mean throughput of each model (and thus compute cost) relative to the input throughput   during cost calculation, since operators in ML workflows may have different output-to-input ratios. 
For model accuracy, we need to understand the accuracy response of a model with respect to the accuracy of upstream models whose outputs are fed into it.
\name varies the input accuracy by selecting different upstream models (with different accuracy profiles) and measures the output accuracy response of the model under test.
For example, consider a model with two inputs and exhibits the following accuracy profile: $(60\%, 50\%) \rightarrow 55\%$, $(50\%, 60\%) \rightarrow 60\%$, $(70\%, 90\%) \rightarrow 65\%$.
This profile enables us to conservatively estimate the output accuracy by identifying the row that is closest to (but not higher than) the accuracy of all inputs; for example, if the input accuracy is (55\%, 83\%), then we can conclude the output accuracy is at least 60\%.
One assumption we make here is that the output accuracy is monotonically increasing with respect to each input accuracy (with the others fixed). 
In \xref{sec:model}, we demonstrate how we use the accuracy profile to select models that satisfy the user's target end accuracy.

Next, we describe our solution that finds highly effective execution plans as well as components to derive query-level accuracy and, assign workers across the heterogeneous infrastructure.

\subsection{Model Selection}
\label{sec:model}
Model selection balances the inference cost and model accuracy: 

\vspace{0.1in}
\noindent\textbf{Satisfying Accuracy Constraints}. One challenge in our model selection is to estimate the query-level accuracy given profiles of individual ML models, which can be non-trivial due to the dependency among them. So far, this has not been discussed in any prior work, and we propose a solution here as follows. 

We consider the dependency graph of the ML operators in the logical plan $G$. For each operator, we can assign (choose) a model variant; the final accuracy for the model selections $s$ should satisfy a user specified accuracy threshold $A$. We use the model profiles to determine whether a model configuration satisfy the accuracy constraint. In each model's accuracy profile, we need to choose a row such that the output accuracy of a model is larger than a downstream node's required input accuracy. Also, the final output model's accuracy has to be above the target end-to-end accuracy.

\vspace{0.1in}
\noindent\textbf{Reducing the Total Cost}.
Another problem during model selection is that we do not know worker assignments yet and thus we cannot use a concrete cost. Thus, we need to choose models based on a different cost definition. We can use the execution latency on a single GPU or the number of parameters in the model. In our current prototype, we use a simple notion of cost: the latency for model inference on the most powerful infrastructure worker (e.g., NVIDIA V100 GPU in our evaluation).

We use the accuracy profiles and perform a \emph{beam-search} to find the model assignments that can attain user's specified end-to-end accuracy threshold. We traverse the graph in reverse topological order, and assign the model for each node. Each candidate is a combination of partial model assignment and the accuracy requirements for upstream nodes. Specifically, we first extract the accuracy requirement for a node that we are currently assigning, and then iterate through all the candidate models for the node and find models whose output accuracy is greater than the threshold from the downstream models. 
When there are multiple models satisfying the output accuracy, we pick the ones that have the lowest cost. There can be many model configurations that satisfy the accuracy constraint, and we maintain the best $B_{MS}$ number of model configurations based on their costs. After one model selection is found for this node, we then update the model assignment to propagate the accuracy constraints to upstream nodes until all nodes have a model assignment. We have to maintain more than a single candidate model configuration because our cost estimation can be not accurate. The real cost should be the actual cost of deploying this model on a particular worker type in the edge or cloud; here we simply use the latency or model size as the cost. 

\subsection{Worker Assignment}
\label{sec:assign}

The goal here is to take the set of candidate model
selections from the previous step and determine the best mapping from models
to available infrastructure workers that minimizes the overall cost while meeting the input throughput to our system.
We will first present an overview of our worker assignment algorithm, which
makes greedy choices along two dimensions to reduce the large search space
for worker assignment: 1) the order of assigning nodes $v \in V$
to workers, and 2) the workers $w \in W$ to be assigned.
Next, we will describe our approach for determining the per-input cost of
assigning the execution of a model to a given worker, which enables our
greedy selection of workers.
We also discuss key refinements that improve optimality in practice.

\vspace{0.1in}
\noindent\textbf{Computing Assignments}.
We present our solution in Algorithm~\ref{alg:assign}. 
We consider as input a specific candidate model selection (out of the
top-$K$ candidates produced by the previous phase).
The output consists of a mapping between nodes in the logical graph and sets
of available workers.

\cut{
\begin{algorithm}[ht]

\SetAlgoLined
\SetAlgoNoEnd
\DontPrintSemicolon
\SetKwInOut{Input}{Input}
\SetKwInOut{Output}{Output}
\SetKwComment{Comment}{// }{}
\SetKwProg{Func}{Function}{}{}

\Input{Model selection $s: V \rightarrow M$}
\Output{Worker assignment $a: V \rightarrow \mathcal{P}(W)$} 

\Func{Avail($W$, $a$) {\small // Returns set of unassigned workers}}{}
\Func{MinCost($W$, $s$, $v$) {\small // Returns worker with minimum cost}}{}
\Func{TCoeff($w$) {\small // Returns throughput coefficient based on tier}}{}

\For{$v \in \text{Topo}(V)$ \label{line:topo}}{
    $T_{\text{rem}} = T_v$\;
    \While{$T_{\text{rem}} > 0$ \text{\bf and} $|\text{Avail}(W, a)| > 0$}{
        $a[v] \cup= \text{MinCost}(\text{Avail}(W, a), s, v)$\;
        $T_{\text{rem}} -= (t_v^w \times \text{TCoeff}(w))$\;
    }
    \lIf{$T_{\text{rem}} > 0$}
    {
        \Return ERROR
    }
}

\caption{\label{alg:assign} Worker assignment.}
\end{algorithm}
}

\begin{algorithm}[ht]

\SetAlgoLined
\SetAlgoNoEnd
\DontPrintSemicolon
\SetKwInOut{Input}{Input}
\SetKwInOut{Output}{Output}
\newcommand\myCommentStyle[1]{\small\textcolor{gray}{#1}}
\SetCommentSty{myCommentStyle}
\SetKwComment{Comment}{// }{}
\SetKwProg{Func}{Function}{}{}
\newcommand\FuncImpl[1]{\small\textit{$\rightarrow$ #1}}

\let\oldnl\nl 
\newcommand{\nonl}{\renewcommand{\nl}{\let\nl\oldnl}} 

\newcommand{\minuseq}{\mathrel{-}=}
\newcommand{\cupeq}{\mathrel{\cup}=}

\Input{Model selection $s: V \rightarrow M$}
\Output{Worker assignment $a: V \rightarrow \mathcal{P}(W)$} 
\vspace{0.3em}

\nonl \Func{Avail($W$, $a$, $i$) \FuncImpl{Returns unassigned workers in tier $\geq i$}}{}
\vspace{-0.5em}
\nonl \Func{MinCost($W$, $s$, $v$) \FuncImpl{Returns worker with min cost (Eq~\ref{eq:worker_pick})}}{}
\vspace{-0.5em}
\nonl \Func{TCoeff($w$) \FuncImpl{Returns throughput coefficient based on tier}}{}
\vspace{-0.5em}
\nonl \Func{Top($a$, $k$) \FuncImpl{Returns top-$k$ best assignments in set}}{}

$a_B = \{ \varnothing \}$ \tcp*[h]{Current set of assignments in beam}\;
\For{$v \in \text{Topo}(V)$ \label{line:topo}}{
    $a_B' = \{ \}$ \tcp*[h]{Next set of assignments in beam}\; 
    \For{$a_b \in a_B$ \tcp*[h]{Iterates over current set of assignments in beam}}{
        \For{$i \in I$}{
            $T_{\text{rem}} = T_v$, $a_{cur} = a_b$\;
            \Comment{Greedily assign workers up to throughput req.}
            \While{$T_{\text{rem}} > 0$ \text{\bf and} $|\text{Avail}(W, a_{cur}, i)| > 0$ \label{line:assignStart}}{
                $a_{cur}[v] \cupeq \text{MinCost}(\text{Avail}(W, a_{cur}, i), s, v)$\;
                $T_{\text{rem}} \minuseq (t_v^w \times \text{TCoeff}(w))$ \label{line:assignEnd}\;
            }
            \lIf{$T_{\text{rem}} \leq 0$}
            {
                $a_B' = a_B' \cup \{ a_{cur} \}$
            }
        }
    }
    $a_B = \text{Top}(a_B', B_{WA})$ \tcp*[h]{Keep only top assignments in beam} \label{line:topk}\;
}
$a = \text{Top}(a_B, 1)$\;

\caption{\label{alg:assign} Worker assignment.}
\end{algorithm}

In Line~\ref{line:topo}, we start by iterating over each node $v \in V$, using a topological
ordering such that parent nodes are assigned before their downstream child
nodes.
While an optimal solution would need to consider the assignment of all nodes
jointly, this is computationally intractable.
However, due to the nature of realistic workflows and our assumption \texttt{A2} that
limits communication in one direction between tiers (i.e., from lower to
higher), greedily computing worker assignments based on the topological
ordering is a reasonable approximation.
For any particular $V$ and $E$, there may be many valid topological
orderings; therefore, we extend our approach to also iterate over a constant
number of different, randomly-selected topological orderings to improve the optimality.

For a given node $v$, we need to assign a set of workers to execute ML
operator (or task), such that we limit the cost while meeting the input throughput.
Each worker can be assigned to a node $v$\footnote{We use one-to-one mapping due to the low overhead of our processor. See \xref{sec:qp} and \xref{sec:discussion}.}, and such assignments formulate a combinatorial optimization which is NP-hard~\cite{garey1979computers}.

We use a greedy approximation for worker assignment by
considering the cost of assigning a worker $w$ to handle the execution of
node $v$ (with the assignment cost defined at the end of this section).
We assign workers based on availability (i.e., not already assigned) and
ordering from lowest to highest cost until the
input throughput is met, or until we run out of workers to
assign (Lines~\ref{line:assignStart}-\ref{line:assignEnd}).
Given our assumption of one-way communication between infrastructure
tiers (\texttt{A2}), if a node $u$ is greedily assigned to a worker on a
higher-tier, then all nodes $v \in V$, where there exists an edge from $u$
to $v$, are unable to be placed on lower tiers.
We modify this by computing the greedy assignment over expanding pools of
available workers, where the number of pools is equal to the number of tiers $|I|$ and
the i$^\text{th}$ pool contains all workers in the $i^\text{th}$ tier or lower.
We use a beam search to reduce the search space by keeping the best $B_{WA}$
candidate assignments (i.e., those with the lowest cost) out of the
$B_{WA} |I|$ considered at each step (Line~\ref{line:topk}).

Since each tier may be distributed among one or more locations, we cannot simply consider the remaining throughput based on that achieved by a candidate worker $w$ for node $u$ (i.e., $t_u^w$).
Instead, we need to multiply this by the $\text{TCoeff}(w)$, which computes the factor based on the number of locations from the tier of $w$ up to the root of the partition (e.g., cloud tier). 
Consider an example infrastructure that consists of the cloud, hub (2 locations), and edge (5 locations); TCoeff$(.)$ is 1, 2, and 10 for workers on the cloud, hub, and edge (respectively).

\vspace{0.1in}
\noindent\textbf{Assignment Cost}.
To greedily pick workers with minimal unit (or per-input) cost, we need to
take both computation and communication costs into account.
Considering the cost for a node $v \in V$,
with model selection $s$, running on a worker $w$, our overall cost equation is:
\begin{equation}
C_C(s(v), w) +
\sum_{(u,v) \in E}
\sum_{x\in a(u)}
C_B(x, w)
\left( \frac{t_u^x}{T_u} \right)
r(u, v),
\label{eq:worker_pick}
\end{equation}

\noindent containing the unit cost for computation
(first term) and communication (second term).
$s(v)$ is the selected model out of all choices for node $v$, and the unit computation 
cost is derived from the profiler using the cost of each worker
and the throughput of the worker while executing the selected model.

For the unit communication cost, we leverage all previous assigned nodes $u \in V$ that have edges to the
current node $v$.
Hence, the second term involves summing the costs
across all workers assigned to $u$ (i.e., $x \in a(u)$) and the worker
$w$ that is being considered.
Note that we only consider the parents to $v$ and not it's children, since
our greedy algorithm operates in the topological ordering of the nodes, such that
the assignments $a(u)$ for all child nodes $u$ are already known.
If $w_u$ and $w$ are on the same tier, the communication cost between the
workers will be zero (\texttt{A1}); otherwise, there is some bandwidth-based cost for the
traffic between the infrastructure tiers for $x$ and $w$.
This bandwidth cost is multiplied by the amount of communication for $w_u$,
which is based on the unit input size $r(u, v)$ and the fraction of that input
which is handled by $x$.
The fraction of input is equivalent to the ratio of the throughput for $u$ on $x$
compared to the input throughput $T_u$.
For instance, a node $v$ takes a single input from $u$ that is assigned to a worker edge worker $x_1$ (40\% inputs) and cloud worker $x_2$ (60\% inputs). If we assign a worker $w$ at the cloud, the communication cost has to include the split linkages. The term $t_u^x/T_u$ is the fraction of the $u\rightarrow v$ traffic contributed from $x$.

\section{Query processor}\label{sec:qp}
We prototype {\name} upon Naiad~\cite{murray2013naiad} and Timely Dataflow ~\cite{timelydataflow} code base, which offered a low-overhead dataflow abstraction. However, there are additional features that {\name} requires. We outline the challenges and our implementations in the following. 

\vspace{0.05in}
\noindent\textbf{{Operator-level parallelism.}} The code base is designed for data parallelism. Instead, {\name} aims for operator-level parallelism, spanning the workflow and compute nodes across different workers; hence they can execute different portions of the plan. The challenges here are two-fold: (1)  all workers in the code base must execute the same set of operators with different data inputs; (2) the code base uses all-to-all communications for progress tracking, causing unnecessary overhead. 

In the prior sections, we described our optimizer to assign workers to operators, where each worker is responsible for one operator in the graph. Indeed, executions of pipelines or workers that are assigned with multiple nodes are used in production database systems~\cite{ozsu1999principles}. Our solution is simple but effective; as our experiments will show, we may put multiple workers on a single device, since the compute and network overhead of our processor is low.

Therefore, each worker only acquires its input data from upstream workers and sends its outputs to the downstream workers. We build a relay mechanism to serve as a "broker" between adjacent workers. There can be one or more relays in each worker; each receives input data from the relay nodes in the upstream workers. It also collects the outputs and sends them to the relay nodes in the downstream worker. To implement this, we use a thread for each upstream worker that keeps pulling data from the upstream worker's relays through TCP streams and maintaining proper buffers.  There is also a thread for each downstream worker that pulls output data and sends it to the relays of the downstream workers. 
In such a manner, operator-level parallelism is achieved by properly parallelizing independent workers (which can be on the same device), tracking their progress and syncing by treating each worker in our compute graph as a Naiad node. Lastly, we modified the progress tracking algorithm to support node-to-node progress updates.

\vspace{0.05in}
\noindent\textbf{{Networking protocals.}} The code base supports communication among the worker nodes only by relaying on the master node; this results in unnecessary data movements. We augment the networking protocols to enable peer-to-peer communications among the workers; a low networking overhead is essential in a dataflow engine that supports operator-level parallelism.

\cut{
\begin{table}
\begin{small}
\caption{Some AICity models/operators used in our experiments. }
 \label{tab:model_table}
 \begin{tabular}{lcc}
     \toprule
     Model & Backbone NN & \#Parameters (M) \\ \midrule
     Object Re-identification & ResNet-18 & 11.7 \\
     & ResNet-34 & 21.8 \\ 
     & ResNet-50 & 25.6 \\
     & ResNet-101 & 44.5 \\         
     & ResNet-152 & 60.2 \\      
     \midrule
     Object Detection &  YOLOv5n & 1.9 \\                       
     & YOLOv5s & 7.2 \\                         
     & YOLOv5m & 21.2\\                         
     & YOLOv5l & 46.5 \\                         
     & YOLOv5x & 86.7 \\                         
     \midrule
     Speech Recognition & wav2vec2-base & 94.4 \\
     & wav2vec2-large & 315.5 \\
     \bottomrule
 \end{tabular}
\end{small}
\end{table}
}

\newcommand{\tblmodel}[1]{\multicolumn{1}{c}{\textbf{#1}}}

\begin{table}
\caption{Some AICity models/operators used in our experiments. }
\begin{small}
 \begin{tabular}{cccccc}
     \toprule
     Model & \multicolumn{5}{c}{\#Parameters (Millions)} \\
     \midrule
     \tblmodel{resnet} & \tblmodel{18} & \tblmodel{34} & \tblmodel{50} & \tblmodel{101} & \tblmodel{152} \\
     Object Re-identification & 11.7 & 21.8 & 25.6 & 44.5 & 60.2 \\
     \midrule
     \tblmodel{YOLO} & \tblmodel{v5n} & \tblmodel{v5s} & \tblmodel{v5m} & \tblmodel{v5l} & \tblmodel{v5x} \\
     Object Detection & 1.9 & 7.2 & 21.2 & 46.5 & 86.7 \\
     \midrule
     \tblmodel{wav2vec2} & \tblmodel{base} & \tblmodel{large} & & & \\
     Speech Recognition & 94.4 & 315.5 & & & \\
     \bottomrule
 \end{tabular}
\end{small}
\label{tab:model_table}

\end{table}

\vspace{0.05in}
\noindent\textbf{{Containerized worker runtime.}} The code base supports homogeneous runtime only.
To offset runtime and hardware heterogeneity in {\name}, each compute node deploys a containerized runtime with a Linux virtual machine to hold one or more Naiad workers. Table~\ref{tab:model_table} illustrates part of the operators and models used in our experiments; each may contain a feature extraction or classification model.  Within each container, {\name} optionally applies ML compilers~\cite{tensorrt,chen2018tvm} to adapt the model assigned by the QO to the worker hardware. By default, the ML models are implemented in PyTorch within the Naiad map functions.

\vspace{0.05in}
\noindent\textbf{{Relational operators support.}} The code based did not support relational opeartors including filters, join and group-by-aggregation upon columnar inputs. We hence implement these operators in {\name}. The metadata is packaged with the data being transmitted in-between the workers to facilitate relational operations.

\vspace{0.05in}
\noindent\textbf{{Remark}}.
The runtime backend of our prototype system consists of ~12K lines of new code in Rust beyond the Timely Dataflow code base v0.12. While our query optimizer design is independent to the runtime engine, supporting broader runtime backends can be interesting future work.

\begin{table*}[t]
\centering
\caption{Four workload and infrastructure setups. We use $m \times n$ to denote that there exists $m$ servers, each has $n$ vCPUs. We show here the input throughput in frame/request per second (FPS/RPS); we use the mean per-frame/audio size from the input dataset in our cost model. }
\label{tab:setups}
\begin{small}
\begin{tabular}{c|c|c|c|c}
\toprule
Dataset & \multicolumn{2}{c|}{VQA} & \multicolumn{2}{c}{AICity} \\\hline
Setups & Objectives & Infras & Objectives & Infras\\\midrule
\texttt{small}: & Accuracy: 0.55, & Edge: 1x4, 1x8, 1x16, & Accuracy: 0.65, & Edge: 1x4, 1x8, Hub: 1x16, 1xV100,\\
(5 nodes)						& Throughput: 9 rps. & Cloud: 2xV100. & Throughput: 3.5 fps. & Cloud: 1xV100. \\\hline
\texttt{medium}: & Accuracy: 0.56, & Edge: 1x2, 1x4, 2x8, 1x16, & Accuracy: 0.70, & Edge: 1x2, 1x4, 1x8, Hub: 1x8, 1x16, 1xV100,\\
(9 nodes)								 & Throughput: 40 rps. & Cloud: 1x48, 3xV100. & Throughput: 8 fps. & Cloud: 1x48, 2xV100.\\\hline
\texttt{large}: & Accuracy: 0.56 & Edge: 2x2, 6x4, 1xV100, & Accuracy: 0.70 & Edge: 2x2, 2x4, Hub: 4x4, 2xV100,\\
(15 nodes)								 & Throughput: 60 rps & Cloud: 3x8, 3xV100.     & Throughput: 11 fps & Cloud: 3x8, 2xV100.\\ \hline
\texttt{xlarge}: & Accuracy: 0.57 & Edge: 6x2, 10x4, 2xV100 & Accuracy: 0.75  & Edge: 6x2, 3x4, 1xV100, Hub: 8x4, 2x8, 2xV100,\\
(30 nodes)								 & Throughput:	100 rps & Cloud: 2x4, 6x8, 4xV100 & Throughput: 20 fps & Cloud: 1x4, 4x8, 3xV100. \\\bottomrule
\end{tabular}
\end{small}
\label{tabsize}
\end{table*}

\section{Evaluation}
\label{s:eval}
We evaluate {\name} against state-of-the-art techniques for machine learning model serving with the following goals.
 
\begin{itemize}[leftmargin=0.2in]
\item[\texttt{G1}] Is it beneficial to use {\name} for serving ML inference workloads on heterogeneous infrastructures? We showcase end-to-end accuracy and cost measurements comparing with relative systems on two real-world use cases. 

\item[\texttt{G2}] We measure the effectiveness and cost overhead of the {\name} processor on various cloud and physical runtime. 

\item[\texttt{G3}] To show that our optimizer is near optimal, we tease apart the usefulness of various aspects of the {\name} optimizer in an ablation study and compare with alternative ML  model selection and placement strategies as well as lower bounds.

\item[\texttt{G4}] We study the robustness and flexibility of {\name} in a sensitivity analysis by varying the systems and workload settings.

\end{itemize}
 
\subsection{Experiment Setup}\label{sec60}

\noindent\textbf{Datasets}.
We consider two realistic machine learning workflows (and associated datasets) for model inference:

\vspace{0.05in}
\noindent\emph{NVIDIA AI City Challenge (AICity)}~\cite{aicity} is a public dataset and benchmark to evaluate tracking of vehicles across multiple cameras. The dataset is divided into 6 traffic intersection scenarios in a mid-sized US city, which in total contains 3.58 hours of videos collected from 46 cameras.  The frame has 1.1MP (megapixels), and a frame has 22 objects (cars) on average. The ReID models are trained on the  CityFlowV2-ReID dataset~\cite{tang2019cityflow}, while the object detection models are pre-trained on the COCO image dataset~\cite{lin2014microsoft}. We leverage their testing scenario in our system evaluations. Figure~\ref{fig:ai_city} demonstrates a typical workflow upon this dataset with an object detection  model, an object Re-identification (ReID) model and the subsequent tracking modules to derive cross-camera vehicle trajectories.

\vspace{0.05in}
\noindent\emph{Visual Question Answering (VQA)}~\cite{antol2015vqa} is another public dataset containing open-ended questions about images from the COCO image dataset~\cite{lin2014microsoft}. The task is to generate an answer (from a large set of candidate responses) for an image-question pair. This dataset has 614,163 questions on 204,721 images. The mean input image resolution is 0.3MP and the mean input speech length is 1.5sec. The validation set from the original dataset split is used in our evaluation. 
Figure~\ref{fig:q1} demonstrates a typical workflow for VQA.

\vspace{0.05in}
In our offline profiling, we measure the accuracy of 10 model combinations on the VQA validation set with 121,512 samples, taking 10-20 minutes depending on the model combinations, and 20 model combinations for AI City. For the latter, the test labels are not available hence we use the official benchmarking API~\cite{aicity} to get the IDF1 scores, and the profiling takes 1-2 hours depending on the model combinations. We also use reported accuracy on standard benchmarks whenever available~\cite{yolov5,torchmodelzoo,huggingfacemodels}. We note that these are \emph{one-time, per-database} costs and can be amortized among different ML workflows later on. We use P75 efficiency numbers as input to our optimizer to offset runtime variance; our sensitivity analysis in~\xref{sec63} discusses using other percentiles. 

\vspace{0.05in}
\noindent\textbf{Workload and infrastructure settings}.
We conduct our experiments on the IBM cloud where the workload and infrastructure setups are detailed in Table~\ref{tab:setups}. We evaluate four setups ranging from \texttt{small} to \texttt{xlarge} by varying the number and type of available workers for each infrastructure tier as well as the throughput and accuracy targets. Each compute node represents a virtual machine as described in \xref{sec:qp} with the number of vCPUs specified (2-48), while each GPU compute node represents a VM with a 16GB NVIDIA V100 GPU.
The memory of each node ranges from 4GB to 192GB and the bandwidth ranges from 3Gbps to 25Gbps. In \xref{sec63}, we show experiments when the bandwidth is limited.
While the absolute infrastructure tier configurations may not capture all real-world infrastructure setups (e.g., IoT devices with compute $<$2 vCPUs), we note that the \emph{relative} compute power difference between tiers does capture this. Using these settings straightens our evaluations as our processor offsets hardware heterogeneity by virtualization and ML compilers (\xref{sec:qp}). 

We strive to echo to real-world scenarios when setting up the base resource costs in our experiments; nevertheless, there can be orthogonal factors such as dynamic pricing models~\cite{ha2012tube}. Hence, we use the unit compute and networking costs based on the pricing catalog of the IBM Cloud as of April 2022~\cite{ibmcloudprice}. The unit costs increases sub-linearly along with the resources used (e.g., 1 and 1.5 unit costs for 2 and 8 vCPUs respectively, and 3 for V100). The communication costs among different tiers (e.g., from edge to cloud) range from 0.1 to 0.3 unit cost per GB; for example, direct communication from edge to cloud bypassing local hubs is more expensive. 
 
We also leverage prior VQA and AICity solutions on top of the benchmarks from~\cite{ben2017mutan,liu2021city} and set up the accuracy and throughput targets used in our experiments based on the profiles of these state-of-the-art solutions. The virtual machines are chosen such that \texttt{small} and \texttt{medium} aim for low serving costs without edge GPUs, while the larger setups aim for low latency with edge GPUs available. The later cases also demonstrate how compute can be moved to the cloud when the edge has not enough compute power.

\vspace{0.05in}
\noindent\textbf{Evaluation metrics} used in our experiment include: 

\vspace{0.05in}
\noindent\emph{Throughput, accuracy and overhead}. We wonder if the systems to be evaluated can timely execute various ML inference workloads, have low overhead and  can keep up with the input throughput. We report both estimated and actually achieved throughput in one hour, as well as various overheads incurred by our query optimizer and processor.
We  also aim for a system that provides viable trade-offs between accuracy and throughput; we report the actual accuracy scores on the validation sets described earlier.

\vspace{0.05in}
\noindent\emph{Serving costs}.
We report the compute and networking costs of executing the ML workload on the infrastructures specified in Table~\ref{tab:setups}. We evaluate the costs while varying the target accuracy and input throughput. For \name and all baselines (described next), we report the serving costs and other metrics when the system saturates, excluding model loading, system startup and shutdown time. 

\begin{figure}
    \centering 
    \begin{subfigure}{0.9\linewidth}
	\raisebox{0.55in}{\rotatebox[origin=t]{90}{\small{\textbf{(a) VQA}}}}
      \includegraphics[width=\linewidth]{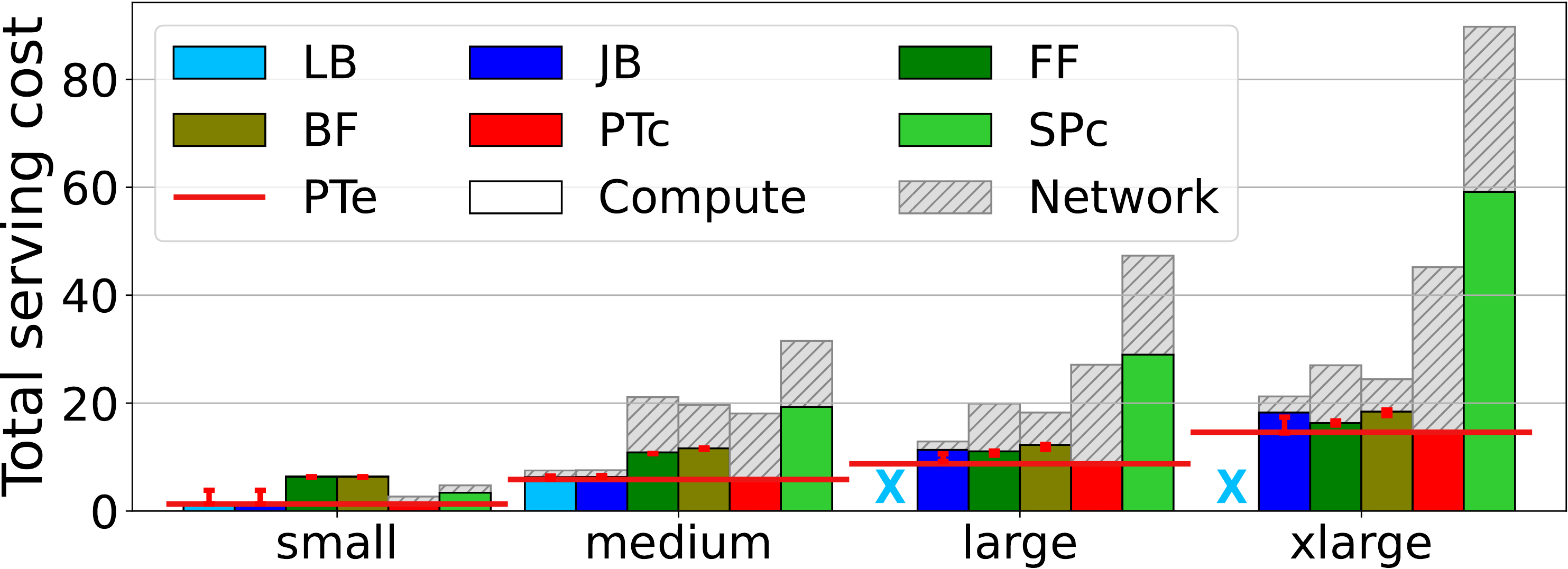}
    \end{subfigure} 
    \begin{subfigure}{0.9\linewidth}
	\raisebox{0.6in}{\rotatebox[origin=t]{90}{\small{\textbf{(a) AICity}}}}
      \includegraphics[width=\linewidth]{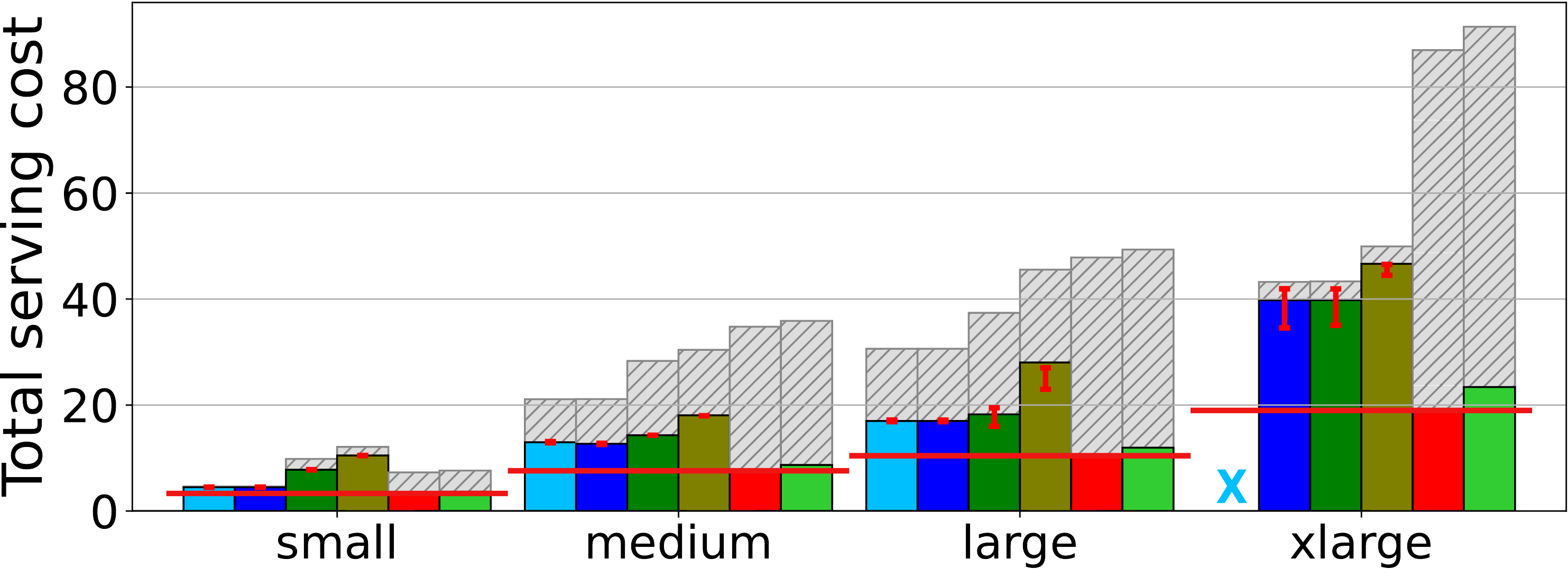}
    \end{subfigure}
\caption{End-to-end evaluations of ML serving. We showcase the actual total serving costs using the P75 profiles; the error bars illustrate the estimated costs using the P50 and P90 profiles (see \xref{sec63}). `X' indicates unsolvable given 1h of QO time.}
\label{fig:end_to_end}
\end{figure}

\begin{figure}
	\begin{subfigure}{0.45\linewidth}
      \includegraphics[width=\linewidth]{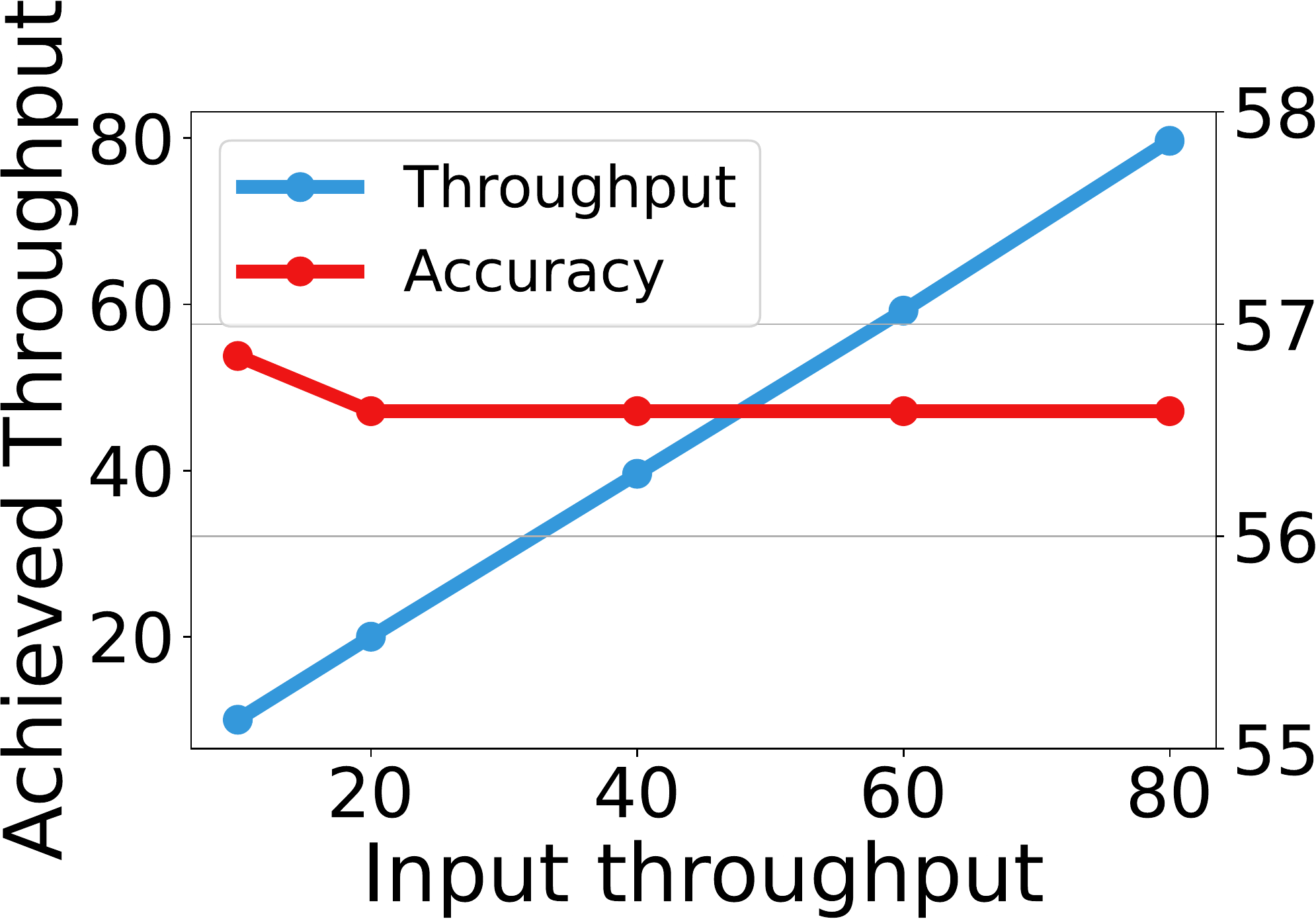}
       \caption{VQA}
    \end{subfigure} \hfil
	\begin{subfigure}{0.45\linewidth}
      \includegraphics[width=\linewidth]{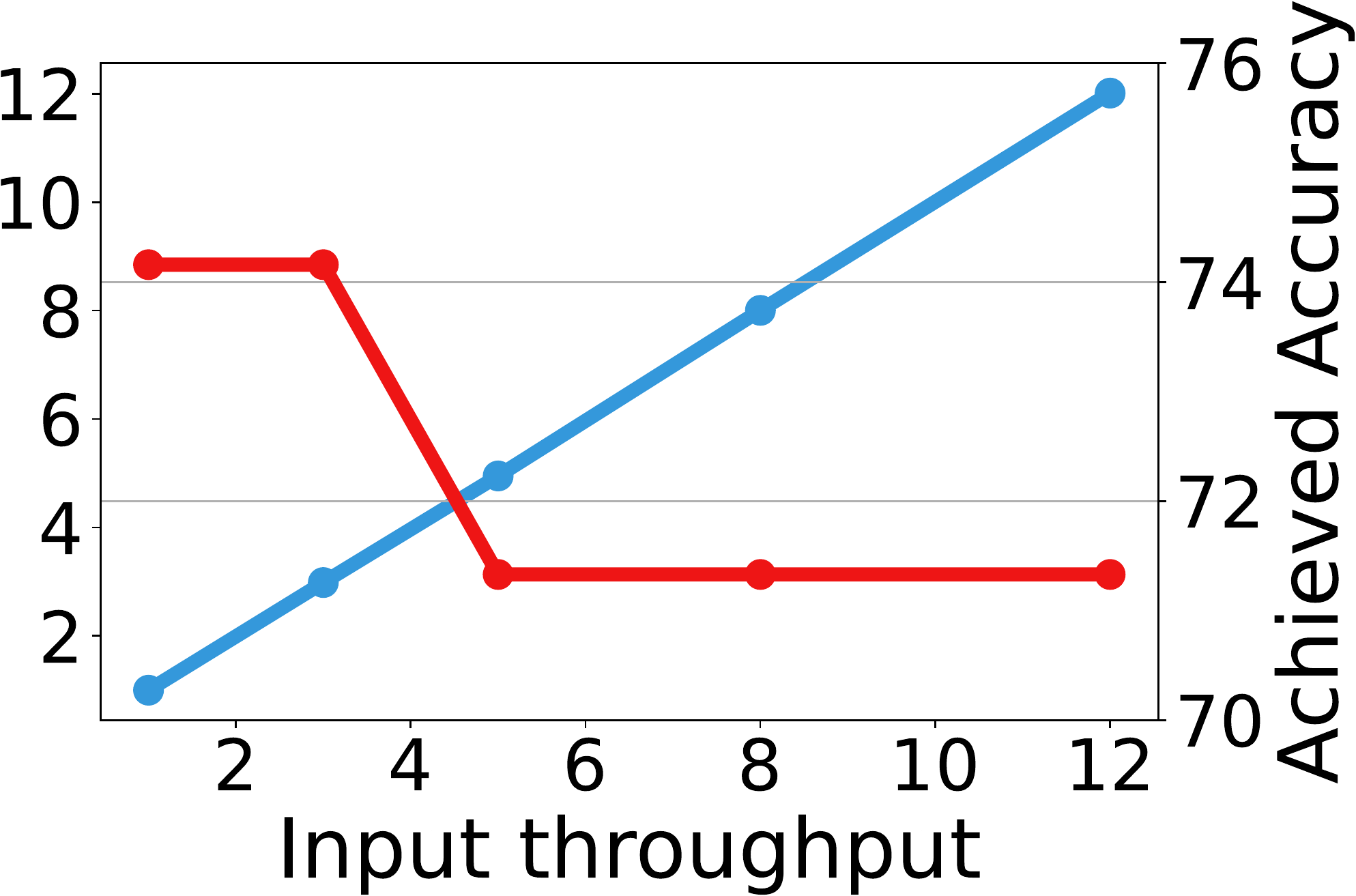}
      	  \caption{AICity}
    \end{subfigure}\vspace{-0.1in}
\caption{Achieved throughput and accuracy given different input throughput on the \texttt{medium} setup.}
\label{fig:actual}
\end{figure}
 
\vspace{0.05in}
\noindent\textbf{Baselines and comparisons}. To compare {\name} (\texttt{JB}) over state-of-the-art ML serving solutions on heterogeneous infrastructures, we consider the following baselines in our experiments: 

\vspace{0.05in}
\noindent\emph{Worker assignment strategies}.
Inspired by geo-distributed database placement~\cite{huang2019yugong,pu2015low,panigrahy2011heuristics} and VM placement strategies~\cite{grandl2014multi}, we compare with using the following model selection and worker assignment strategies while using the {\name} processor\footnote{We note that Worst Fit placement~\cite{panigrahy2011heuristics} that greedily puts models on the most expensive location does not fit in our context.}: (1) Best Fit (\texttt{BF}) is inspired by geo-distributed database optimizers~\cite{huang2019yugong,pu2015low} to reduce the networking costs; it uses the most accurate model and greedily assigns jobs to the cheapest worker on the same infrastructure tier. 
(2) First Fit (\texttt{FF})  follows a classic VM placement strategy~\cite{panigrahy2011heuristics} in which each operator uses the most accurate model and assign jobs to the cheapest worker regardless of their location. (3) Lower bound (\texttt{LB}): we compute a lower bound of the serving cost by enumerating over all possible model choices and worker assignments when keeping the placement constraints (\texttt{A2}). This baseline showcases the optimality of our solution and it is worth noting that \texttt{BF} and \texttt{FF} may not follow the networking constraints used in \texttt{JB} and \texttt{LB}. 

\begin{table}
    \caption{Cost analysis on the AICity dataset. We show the costs for one hour of input data with input throughput specified in Table~\ref{tab:setups} and the corresponding query optimizing time (QO).}
	\begin{small}
    \begin{tabular}{cccccccc}
        \toprule
        \multicolumn{2}{c}{Model} & \multicolumn{3}{c}{\texttt{medium}} & \multicolumn{3}{c}{\texttt{large}} \\
        \cmidrule(lr){3-5} 
        \cmidrule(lr){6-8}
        Select. & Assign. & Comp. & Net & QO & Comp. & Net & QO \\
        \midrule
        JB & JB & 12.7 & 8.4 & 6.5ms & 17.0 & 13.6 & 7.8ms \\
        \midrule
        LB & LB & 13.0 & 8.1 & 2.1\textbf{s} &  17.0 & 13.6 & 27\textbf{min}  \\ 
        JB & FF & 9.0 & 14.1 & 3.9ms  &  12.0 & 19.1 & 5.7ms\\
        JB & BF & 16.0 & 8.3 & 3.4ms  &  20.0 & 13.5 & 3.4ms \\
        \bottomrule
    \end{tabular}
	\end{small}
    \label{tab:costs}
\end{table}

\vspace{0.05in}
\noindent\emph{End-to-end ML serving}.
To our best knowledge, there lacks an off-the-shelf solution for serving ML on heterogeneous infrastructures while supporting the functionalities that {\name} can provide. We use the following variants of existing systems to echo the real-world ML deployments. (1) We perform all computation on a single GPU worker using native PyTorch to handle the entire workflow. Doing so has the minimum compute overhead from the software stack beyond PyTorch but has to pay potentially large networking costs if the workers are on the cloud. By default, we use the most accurate models that are available and denote \texttt{PTe} as running PyTorch on the edge, \emph{pretending} that there is a V100 GPU and counting the GPU costs; \texttt{PTc} runs PyTorch on a cloud V100 GPU, which is equivalent to \texttt{PTe} plus networking costs. (2) We assume the data is transferred to the cloud and use the most accurate models in a Spark. This baseline leverages all the cloud GPU workers in each infrastructure setup (Table~\ref{tab:setups}) and performs data parallelism upon native PyTorch wrapped in a map function (\texttt{SPc}).

\vspace{0.05in}
\noindent\emph{Model selection}. The baselines above use the most accurate models available, since none of them solves the model selection problem. We will perform in \xref{sec62} an ablation study to examine the effectiveness of our proposed model selection strategy, showing the optimality gap from using brute force.

\subsection{System Evaluations}\label{sec61}

\noindent\textbf{System efficiency.}
We showcase \texttt{G1} by the the end-to-end evaluations in Figure~\ref{fig:end_to_end} and Table~\ref{tab:costs} using various workload and infrastructure settings in Table~\ref{tab:setups}. We note a few observations here: 

\texttt{JB} demonstrates the best performance with different datasets and setups compared to the baselines. On VQA, \texttt{JB} saves the total serving cost up to 58.1\% compared to the best-performing baseline (\texttt{PTc}) and up to 5x compared to end-to-end ML systems \texttt{SPc}. On AICity, \texttt{JB} saves the total cost for up to 36.3\% compared to the best-performing baseline (\texttt{PTc}) and up to 2.1x comparing to \texttt{SPc}.

We showcase the \emph{actual} throughput and accuracy in Figure~\ref{fig:actual}. \texttt{JB} achieved near 1:1 for actual:expected throughput (diagonal line). While this result is demonstrated on the \texttt{medium} setting, we also show that such a trend holds for the \texttt{large} setting in Appendix~\xref{sec:app_se}. With increasing input throughput but fixed available infrastructures, {\name} successfully trades off throughput with accuracy by picking suitable models.

Comparing \texttt{JB} to \texttt{LB}, we observe a subtle difference in the overall serving costs -- with different input throughput, 94.2\% of the chances \texttt{JB} provides a total cost that has less than 1\% difference to that provided by \texttt{LB} on AICity. 
\texttt{LB} requires a large QO time as will be shown next and becomes unusable -- in AICity, \texttt{medium} has 8K choices while \texttt{large} has 7M choices. All these results indicate the optimality of the {\name} optimizer. 

\begin{figure}
    \centering 
     \includegraphics[width=\linewidth]{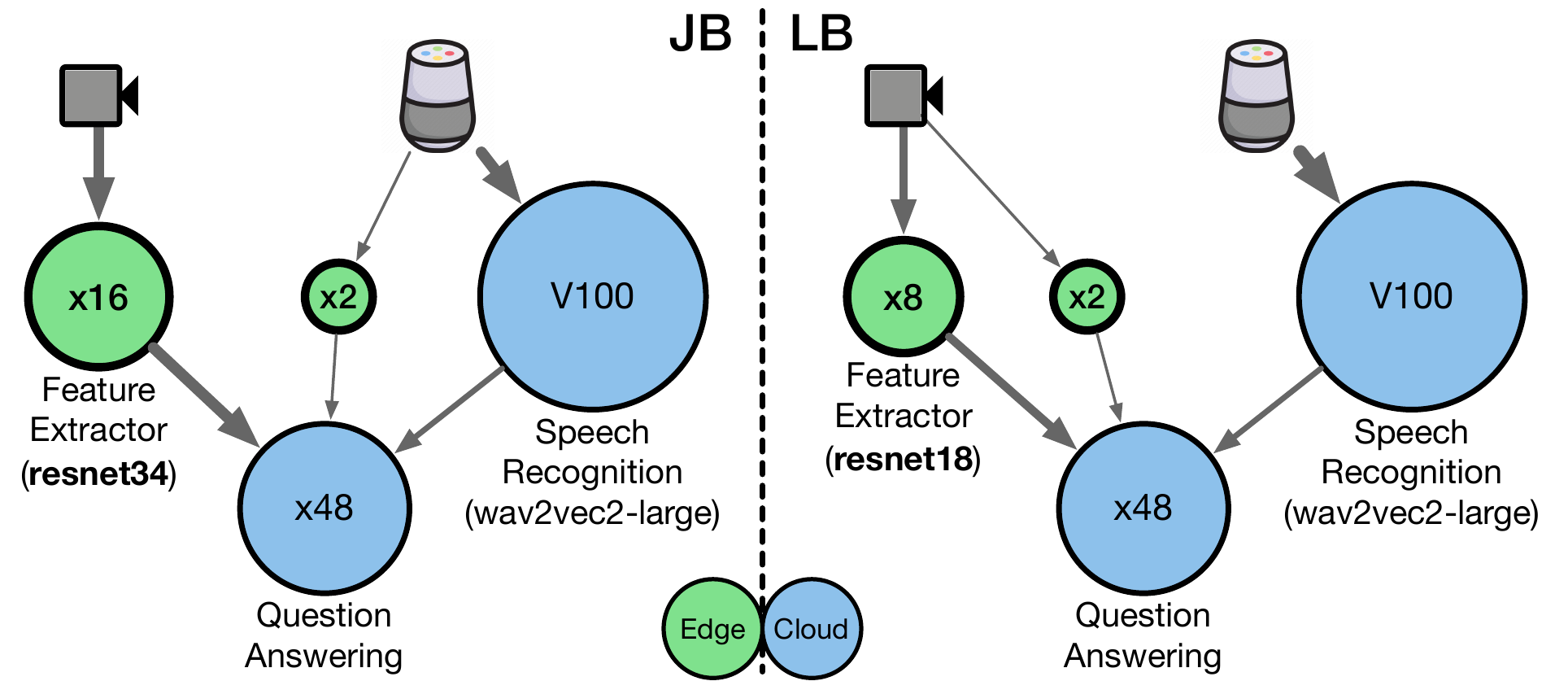}
\caption{Comparison of the execution plans of JB and LB on VQA using the \texttt{medium} setup modified to 20 rps.}
\label{fig:case_study_1}
\end{figure}

Figure~\ref{fig:case_study_1} illustrates a qualitative example of the execution plans of \texttt{JB} and \texttt{LB} when they do not match. JB uses 1x16 worker and a larger ResNet model for feature extraction, while LB uses 1x2 and 1x8 which leads to a lower cost. 
\texttt{BF} and \texttt{FF} failed to find overall optimal execution plans in our experiments; though in some cases, they find plans with low compute or low network costs solely (e.g., BF with low network cost while FF with low compute cost in Table~\ref{tab:costs}). This can be as expected since their heuristics ignore model accuracy-efficiency trade-offs and the resource availability on heterogeneous infrastructures.
In most cases, \texttt{BF} and \texttt{FF} have much higher costs than \texttt{JB}; using heuristics that consider network or compute cost solely is suboptimal. On the other hand, model selection greatly helps to reduce the overall costs, especially when the accuracy target is lower.
\texttt{PTc} and \texttt{SPc} use homogeneous GPU computing, which results in lower compute costs than \texttt{JB} yet larger networking costs since the raw data must be transferred from the edge. \texttt{SPc} exhibits more overhead as compared with \texttt{PTc}~\cite{meng2016mllib}.  
\texttt{PTe} is a hypothetical baseline that assumes strong GPUs on the edge, and thus leads to minimum compute costs at zero networking cost. In real-world applications, with no resource constraint, users should adopt this solution; however, this is often not true in practice.

Further evaluations in~\xref{sec:app_se} show that \name often yields serving costs equal to or close to the lower bound. We also show scenarios (e.g., when Assumption A2 is relaxed) when \name provides worse results.

We observe that the runtime variance is low across all setups; for example, the standard deviation from five runs on the \texttt{large} setup is 0.003\% for AICity and 0.020\% for VQA. The runtime variance on the \texttt{xlarge} setup is reported . 

\begin{table}[t]
    \caption{Costs of operators upon the \texttt{medium} setup. E: Edge. H: Hub. C: Cloud. Original: the latency with native PyTorch. QP exec: the overhead of executing the operator in the JellyBean query processor; QP network: the overhead of communication. }
    \resizebox{0.98\columnwidth}{!}{
    \begin{tabular}{lc|cccccc}
        \toprule
        VQA & \multirow{2}{*}{Node} &\multicolumn{2}{c}{Orignial (ms)} & \multicolumn{2}{c}{QP exec.} & \multicolumn{2}{c}{QP network}\\
        \cmidrule(lr){3-4}
        \cmidrule(lr){5-6}
        \cmidrule(lr){7-8}
        Operator  & & P50 & P90 & P50 & P90 & P50 & P90 \\
        \midrule
        ImgFeat  & x2E & 152.2 & 161.9 & +2.5\% & +4.1\% & +1.2\% & +1.3\% \\
        ImgFeat  & x4E & 76.6 & 82.3 & +7.4\% & +8.1\% & +3.9\% & +3.6\% \\
        ImgFeat  & x16E & 27.4 & 29.4 & +11.7\% & +19.4\% & +6.1\% & +6.0\% \\
        \midrule
        ASR  & x8E & 251.0 & 297.6 & +2.3\% & +1.0\% & +0.4\% & +0.4\% \\
        ASR  & V100C & 24.1 & 26.3 & +0.8\% & +0.8\% & +0.6\% & +0.5\% \\
        \midrule 
        VQA  & x48C & 6.3 & 8.6 & +9.7\% & +12.6\% & +4.8\% & +6.9\% \\
        \bottomrule
    \end{tabular}
    }
    \resizebox{0.98\columnwidth}{!}{
    \begin{tabular}{lc|cccccc}
        \toprule
        AICity & \multirow{2}{*}{Node} & \multicolumn{2}{c}{Orignial (ms)} & \multicolumn{2}{c}{QP exec.} & \multicolumn{2}{c}{QP network}\\
        \cmidrule(lr){3-4}
        \cmidrule(lr){5-6}
        \cmidrule(lr){7-8}
        Operator & & P50 & P90 & P50 & P90 & P50 & P90 \\
        \midrule
        ObjDet & x2E & 1412 & 1455 & +3.4\% & +8.2\% & +0.3\% & +0.4\% \\
        ObjDet & x4E & 721.0 & 732.0 & +6.4\% & +8.7\% & +0.4\% & +0.7\% \\
        ObjDet & x8E & 451.7 & 468.3 & +3.5\% & +6.1\% & +0.8\% & +0.8\% \\
        ObjDet & V100H & 19.7 & 20.6 & +13.7\% & +13.3\% & +8.2\% & +10\% \\
        \midrule
        ReID & V100C & 8.4 & 8.7 & +6.5\% & +7.5\% & +6.0\% & +6.2\% \\
        ReID & V100C & 8.5 & 8.8 & +6.1\% & +7.6\% & +20\% & +20\% \\
        \bottomrule
    \end{tabular}
    }
    \label{tab:latency}
\end{table}

\vspace{0.05in}
\noindent\textbf{System overhead.} Table~\ref{tab:costs} also illustrates \texttt{G3} -- the \texttt{JB} optimizer has a small overhead with the QO time of \texttt{JB} in a few milliseconds. In comparison, LB uses brute force, which incurs adverse QO time in larger infrastructure settings (e.g., 27 minutes for \texttt{large}). Other placement strategies have smaller QO time due to a smaller search space, but the total serving costs are larger.

We further demonstrate in Table~\ref{tab:latency} the compute overhead of the \texttt{JB} processor. We show the 50th and 90th percentile of various ML operators in native PyTorch and by the \texttt{JB} processor. The overhead caused by \texttt{JB} processor, as partially been discussed in~\cite{murray2013naiad}, contains that for metadata parsing, data (un)packing, network I/O, and task scheduling. The QO latency is reported on 1x8 virtual CPU node with a Python implementation. Results indicate a small overhead ranging from a few to 19\% upon the native PyTorch executions. This is significantly smaller that that of Spark which may take up to 300\% (as shown in Figure~\ref{fig:end_to_end}).

\vspace{0.05in}
\noindent\textbf{Remark.}
Our evaluations across various workload and infrastructure setups showed that {\name} efficiently computes and deploys execution plans and significantly reduces the total serving cost of real ML workloads. We believe it is beneficial to leverage {\name} for serving ML on heterogeneous infrastructures across a wide range of real-world applications.

\subsection{Ablation Study}\label{sec62}
We leverage the \texttt{medium} setting and evaluate the effectiveness of {\name} by sweeping different knobs used during query optimization. We also demonstrate similar experiments on other setups in Appendix~\xref{sec:app_ab}.

\begin{figure}
	\begin{subfigure}{0.475\linewidth}
	\centering
      \includegraphics[width=\linewidth]{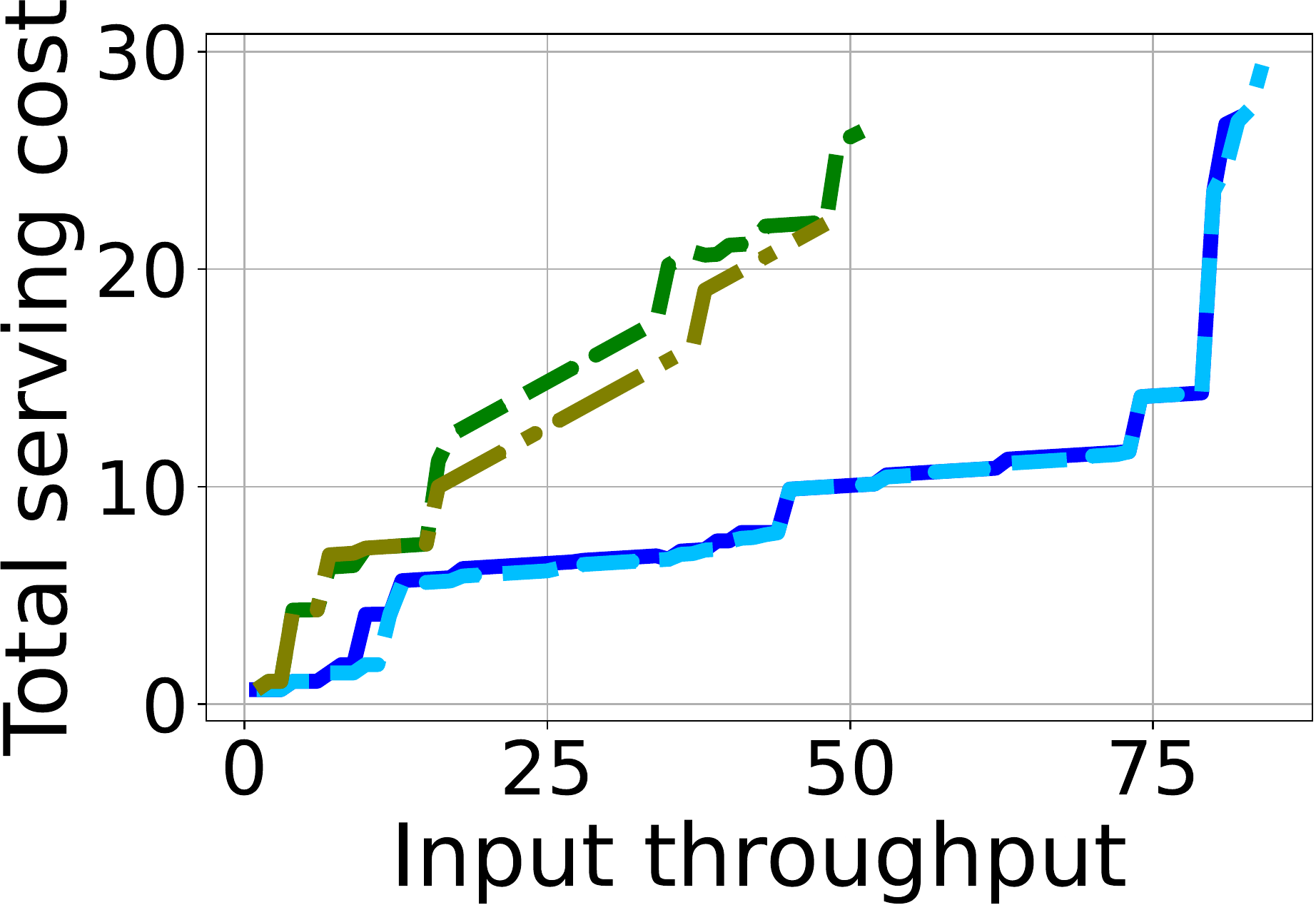}
     \caption{VQA}
     \label{fig:ab1-vqa}
    \end{subfigure} \hfil	
	\begin{subfigure}{0.475\linewidth}
	\centering
      \includegraphics[width=\linewidth]{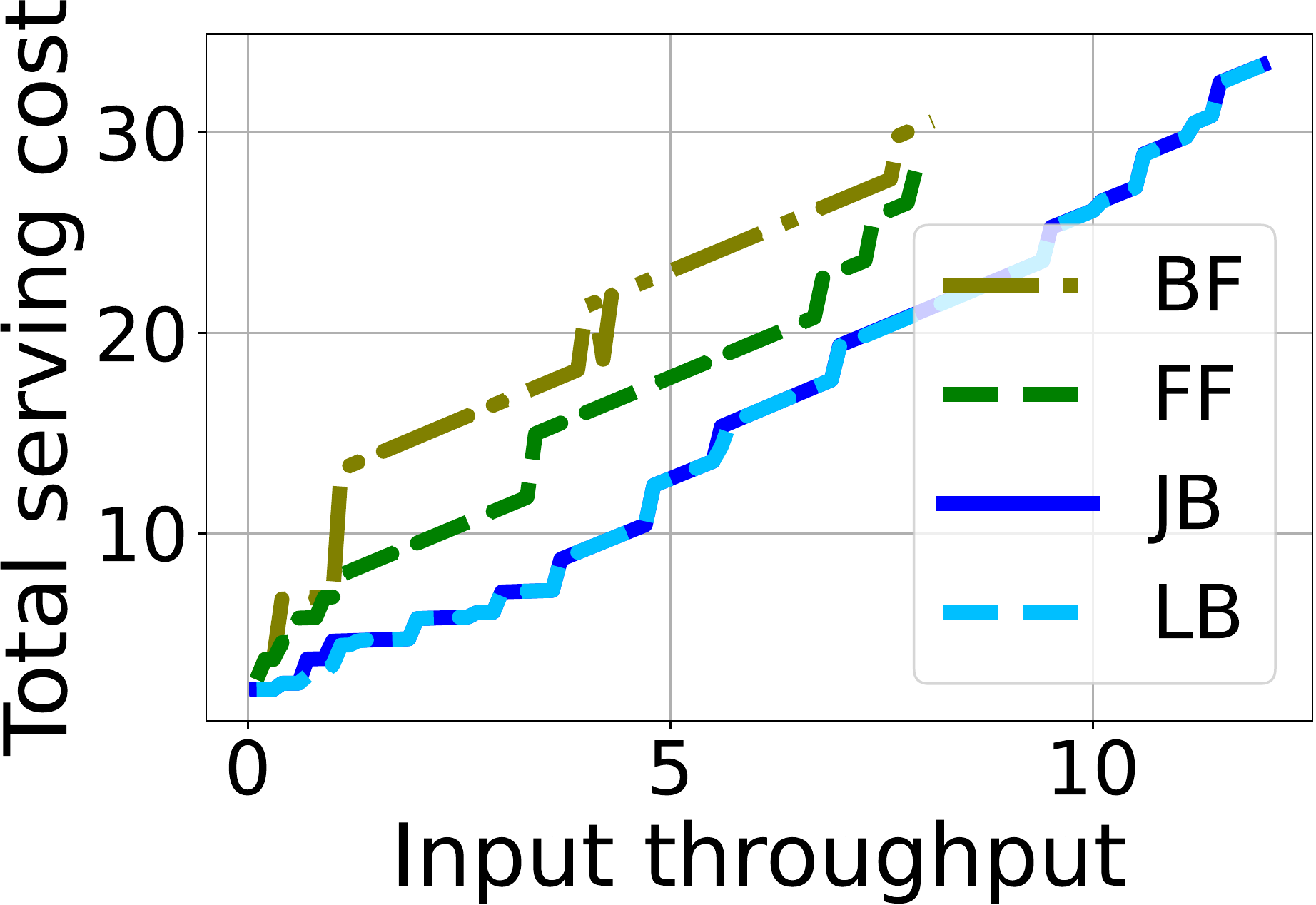}
	  \caption{AICity}
    \end{subfigure}\vspace{-0.1in}
\caption{Total serving cost w.r.t. input throughput} in \texttt{JB}. 
\label{fig:ab1}
\end{figure}

\vspace{0.05in}
\noindent\textbf{Input throughput.} To demonstrate the scalability of {\name} and to supplement Figure~\ref{fig:end_to_end}, we leverage a fixed target accuracy as in \texttt{medium} and demonstrate how the costs change when varying the input throughput. Figure~\ref{fig:ab1} shows the results. We observe that {\name} can keep up with increasing input throughput and is near optimal -- in most situations, \texttt{JB} achieves the same total serving costs as \texttt{LB}. For \texttt{BF} and \texttt{FF}, no valid execution plans can be found beyond 51 rps and 8 fps (VQA and AICity, respectively).
 
\begin{figure} 
	\begin{subfigure}{0.475\linewidth}
      \includegraphics[width=\linewidth]{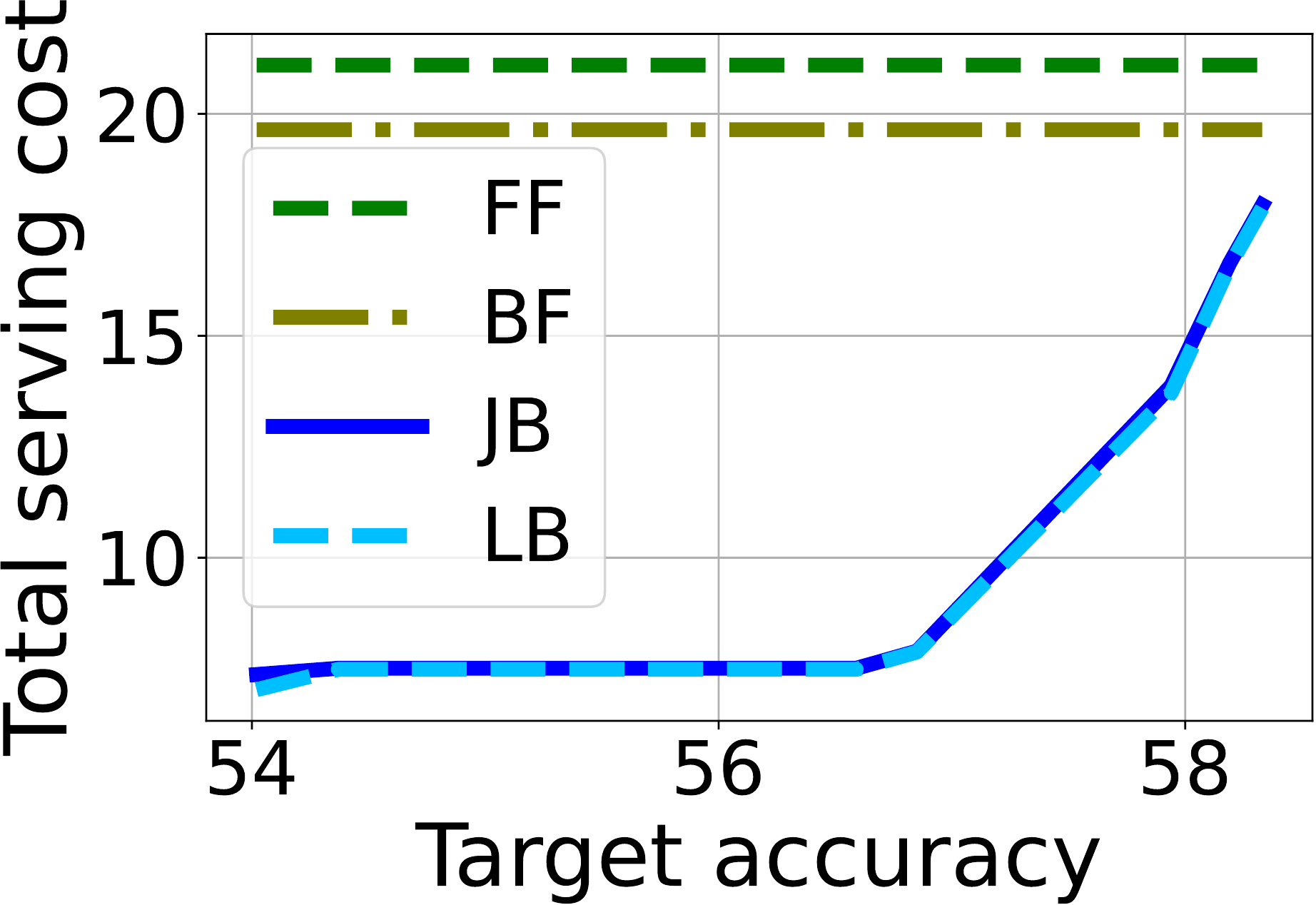}
          \caption{VQA}
    \end{subfigure} \hfil
	\begin{subfigure}{0.475\linewidth}
      \includegraphics[width=\linewidth]{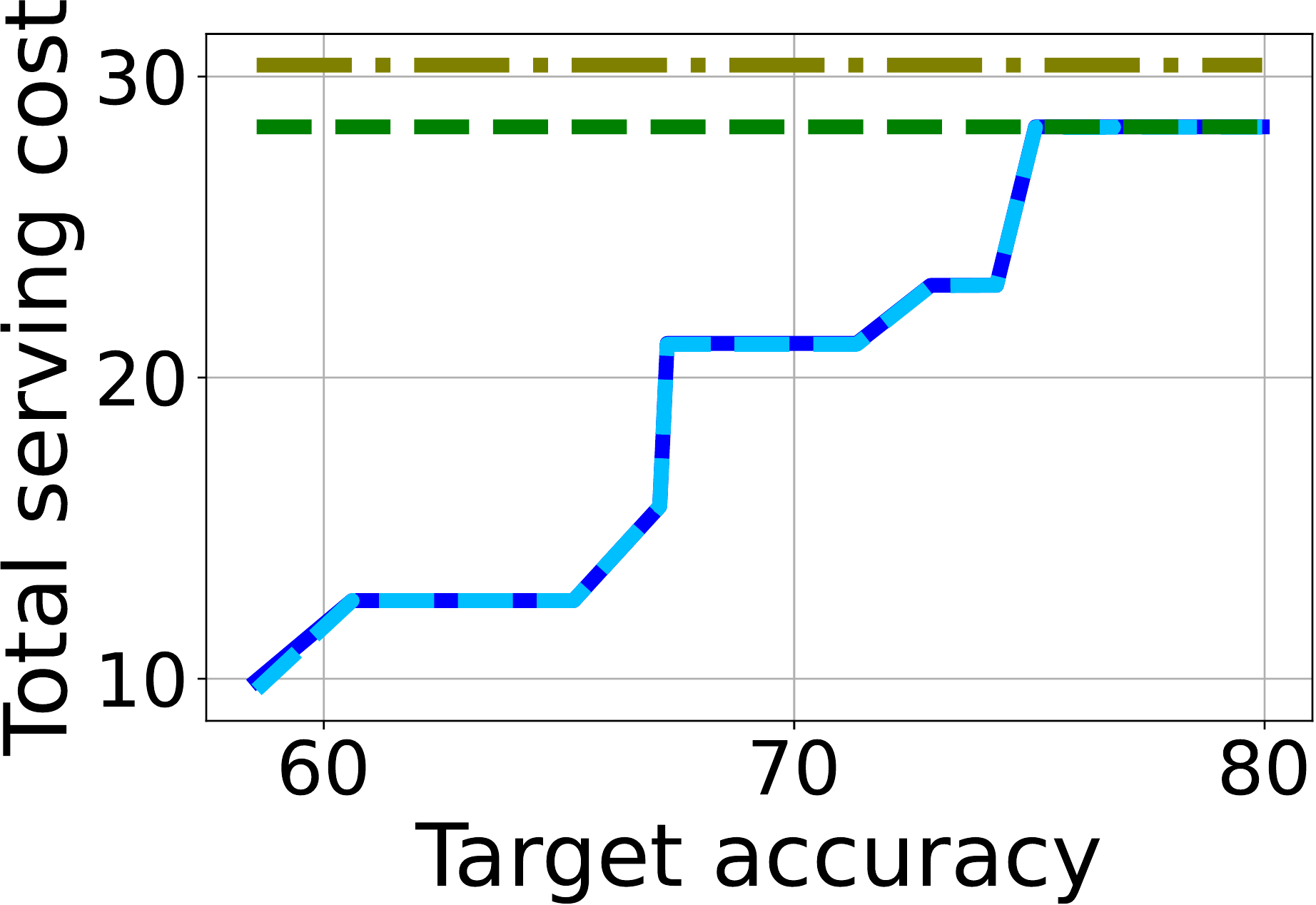}
     	  \caption{AICity} 
    \end{subfigure}\vspace{-0.1in}
\caption{Total serving cost w.r.t. target accuracy in \texttt{JB}.}
\label{fig:ab2}
\end{figure}

\vspace{0.05in}
\noindent\textbf{Target accuracy.} To show that {\name} provides viable accuracy-cost trade-offs, we fix the target throughput as in \texttt{medium} and demonstrate the total serving costs by varying the target accuracy. Figure~\ref{fig:ab2} shows the results. \texttt{BF} and \texttt{FF} solve only for placement while using the most accurate models, and thus the costs are constant. For the scenarios we examined, {\name} is near optimal across a range of accuracy targets. \texttt{JB} and \texttt{LB} eventually use the most accurate models, converging with \texttt{FF} for AICity.

\vspace{0.05in}
\noindent\textbf{Effect of model selection.} To examine the model selection strategy used in {\name} (\xref{sec:model}), Table~\ref{tab:ablation} illustrates an ablation study in which we substitute our model selection for either the most accurate models or a brute force selection. We also evaluate our model selection strategy for \texttt{PTc} and \texttt{SPc}. Results show that our proposed model selection is effective with our {\name} processor as well as other ML runtimes.

\begin{table}
    \caption{Ablation analysis of model selection on the AICity dataset.}
	\resizebox{0.98\columnwidth}{!}{
    \begin{tabular}{cccccccc}
        \toprule
        \multicolumn{2}{c}{Model} & \multicolumn{3}{c}{\texttt{medium}} & \multicolumn{3}{c}{\texttt{large}} \\
        \cmidrule(lr){3-5} 
        \cmidrule(lr){6-8}
        Select. & Assign. & Comp & Net & QO & Comp & Net & QO \\
        \midrule
        JB & JB & 12.7 & 8.4 & 6.5ms & 17.0 & 13.6 & 7.8ms\\
		Most acc. & JB  & 14.3 & 14.0 & 1.2ms & 17.5 & 19.1 & 1.3ms\\
		Brute f. & JB & 12.7 & 8.4 & 11.6ms & 17.0 & 13.6 & 15.0ms\\
        \midrule
		JB & PTc & 5.3 & 27.2 & N/A & 7.3 & 37.4 & N/A \\
		JB & SPc & 5.8 & 27.2 & N/A & 8.0 & 37.4 & N/A \\
		Most acc. & PTc & 7.6 & 27.2 & N/A & 10.4 & 37.4 & N/A \\
		Most acc. & SPc & 8.7 & 27.2 & N/A & 12.0 & 37.4 & N/A\\ 
        \bottomrule
    \end{tabular}
	}
    \label{tab:ablation}
\end{table}

\subsection{Sensitivity Analysis}\label{sec63}
We further study the robustness and flexibility of {\name} (\texttt{G4}) with the following sensitivity analysis experiments.

\begin{figure}
    \centering 
    \begin{subfigure}{0.48\columnwidth}
      \includegraphics[width=\linewidth]{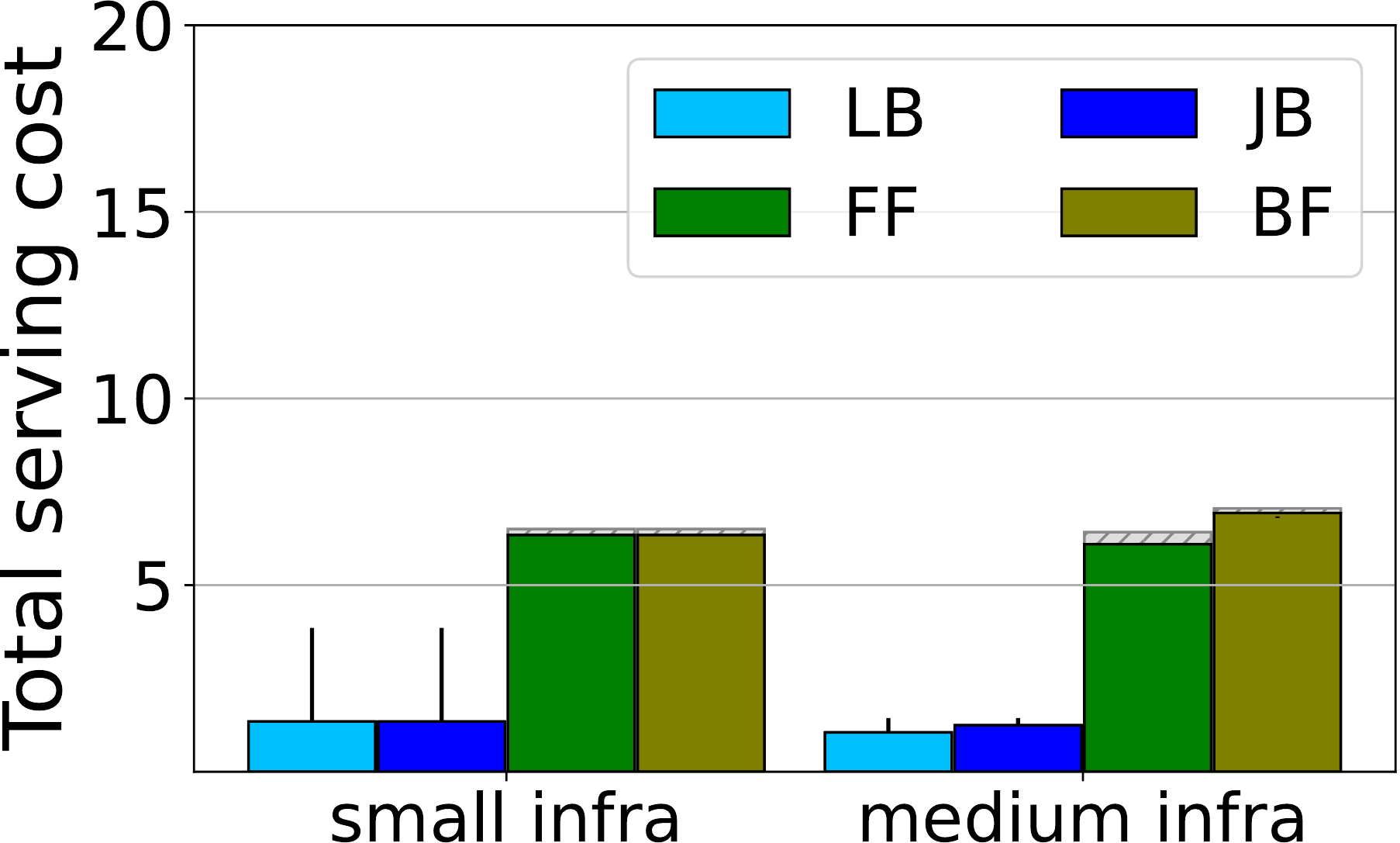}
      \caption{VQA}
    \end{subfigure} \hfil
    \begin{subfigure}{0.48\columnwidth}
      \includegraphics[width=\linewidth]{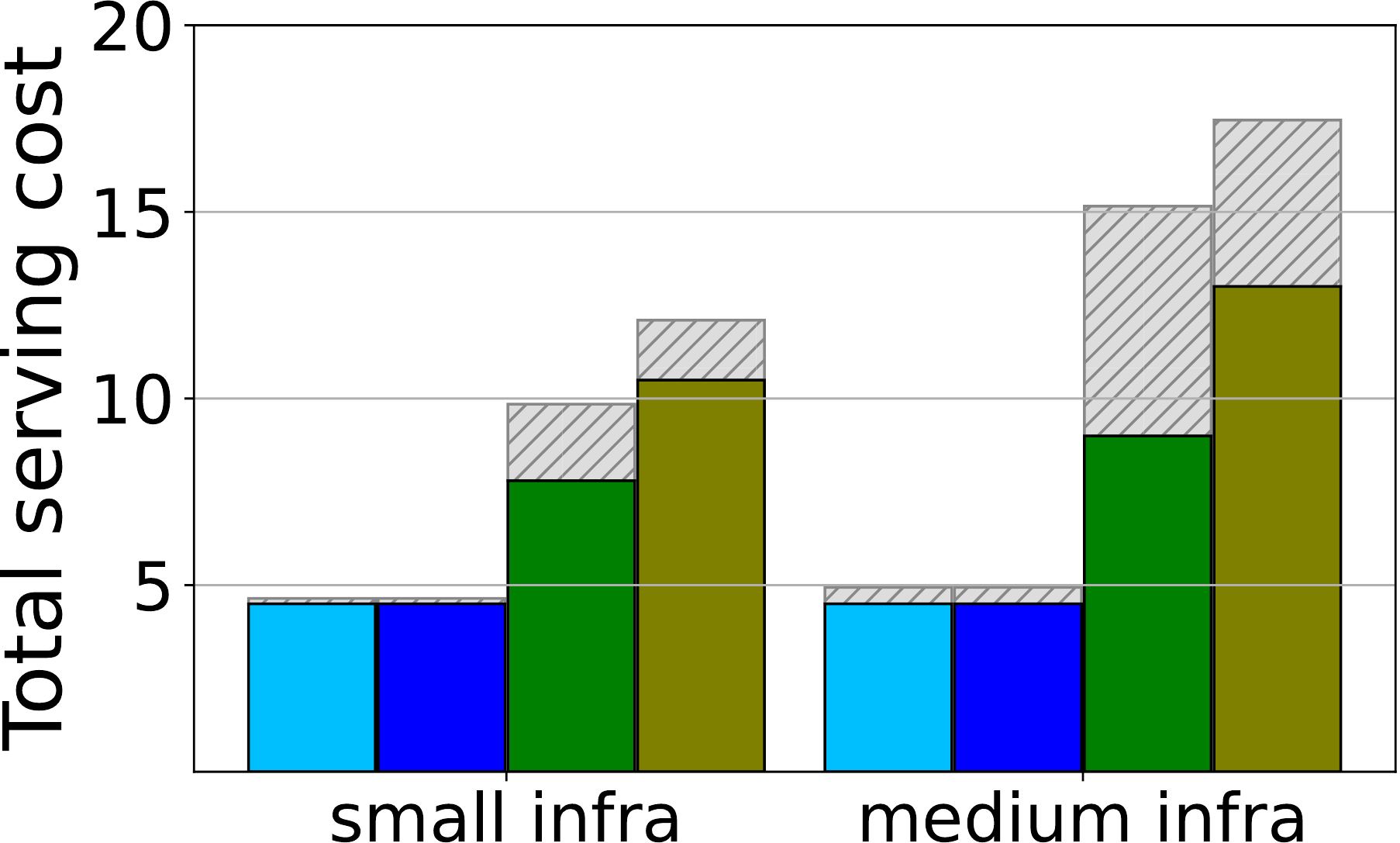}
      \caption{AICity}
    \end{subfigure}\vspace{-0.1in}
\caption{Applying the \texttt{small} workload setup on the \texttt{medium} infrastructures. {\name} uses the minimum available resource to achieve an optimal performance.}
\label{fig:sen1}
\end{figure}

\vspace{0.05in}
\noindent\textbf{Effect of resource over-subscriptions.} When there are more resources than needed, especially on the cloud, can {\name} handle the workloads without wasting resources? Also, how do the costs change? We answer these questions by deploying the \texttt{small} workload on the \texttt{medium} infrastructure (Table~\ref{tab:setups}). Figure~\ref{fig:sen1} illustrates the results. We observe that compared with using the \texttt{small} infrastructure, more resource availability will not significantly increase the serving cost for {\name} with a fixed workload. However, \texttt{BF} and \texttt{FF} cannot guarantee cost efficiency in such a scenario. This is largely due to their sub-optimal worker assignment strategies which disregard resource availability. With {\name}, users may use large cloud subscriptions without wasting resources.

\begin{figure}
    \centering 
    \begin{subfigure}{0.48\columnwidth}
      \includegraphics[width=\linewidth]{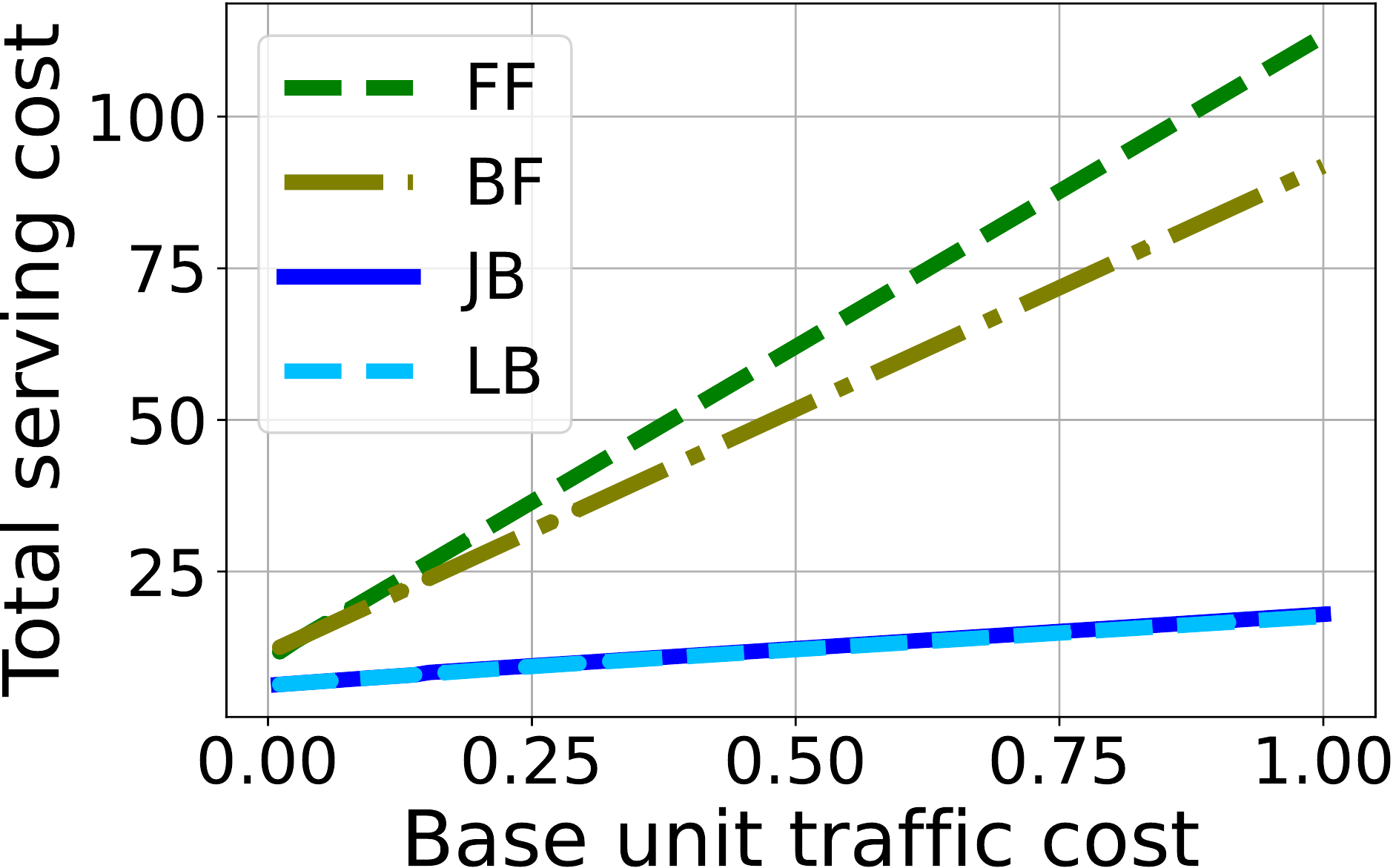}
      \caption{VQA}
    \end{subfigure} \hfil
    \begin{subfigure}{0.48\columnwidth}
      \includegraphics[width=\linewidth]{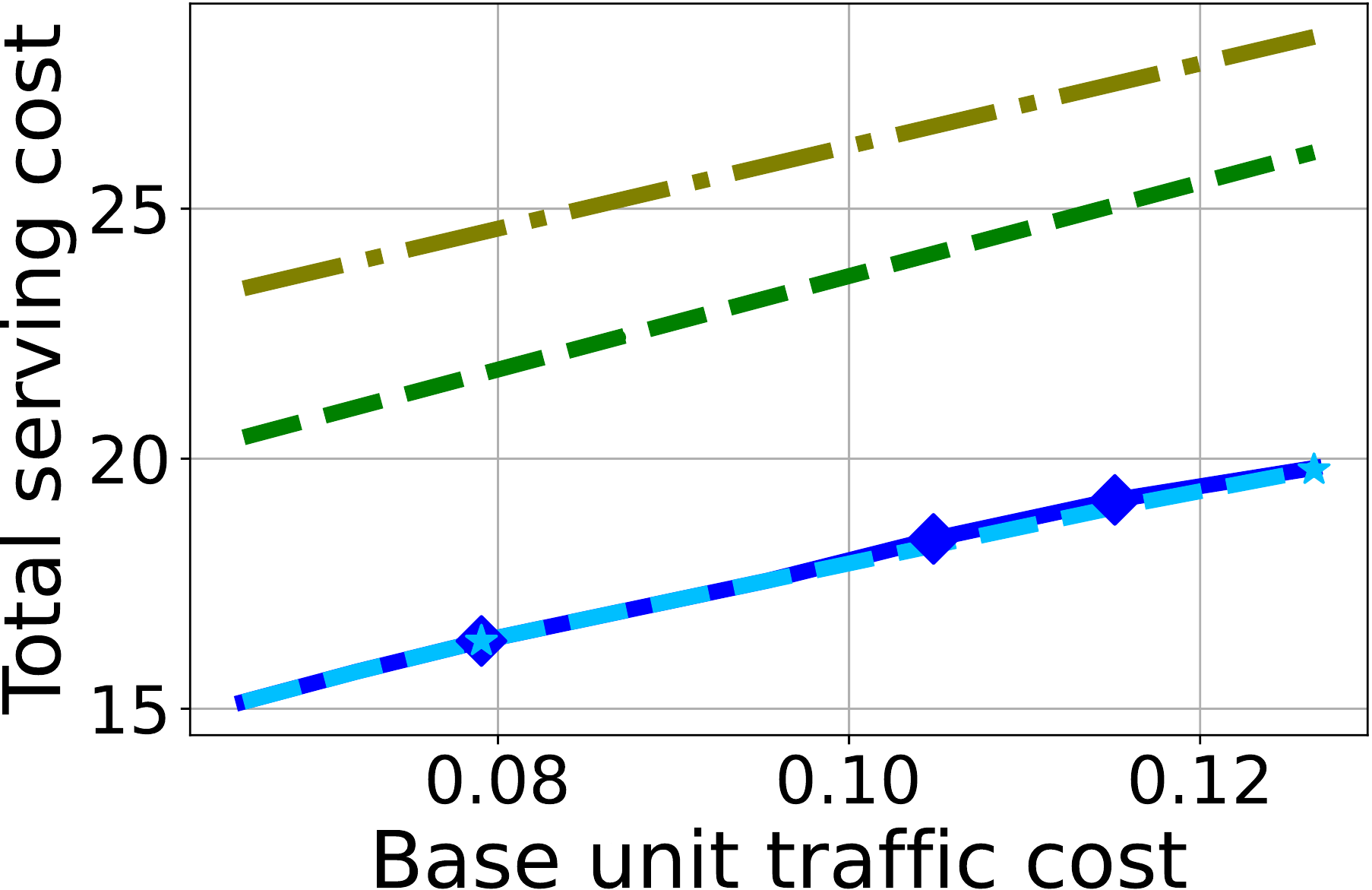}
      \caption{AICity}
    \end{subfigure}\vspace{-0.1in}
\caption{The overall serving costs w.r.t. base unit traffic cost. Dots indicate when the execution plan changes. On AICity, different lines are near linear after 0.12 and hence we show cropped results. }
\label{fig:sen2}
\end{figure}

\begin{figure}
    \centering 
     \includegraphics[width=\linewidth]{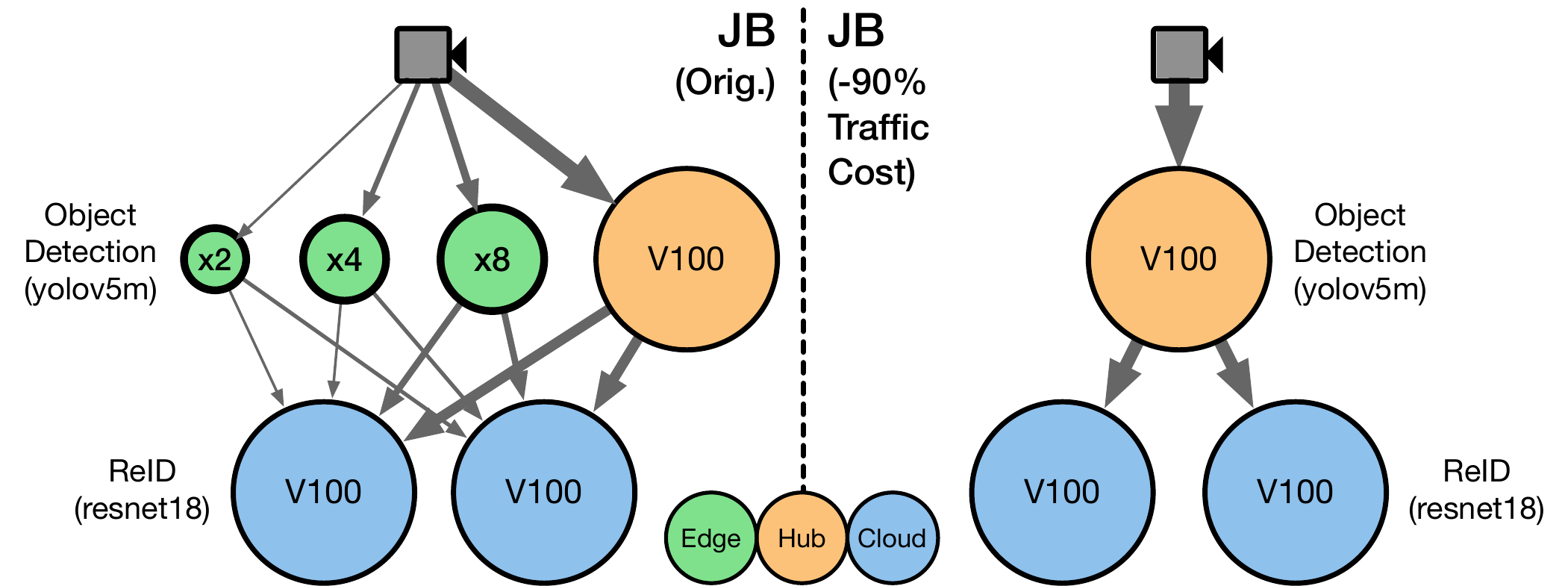}
\caption{Change of worker assignment when unit traffic cost is 10\% of the original traffic cost on \texttt{medium} setup for AICity. }
\label{fig:case_study_2}
\end{figure}

\vspace{0.05in}
\noindent\textbf{Base unit network costs}. We examine the effect of varying the network costs in a \texttt{medium} setup, which play a critical role in the total serving costs. Figure~\ref{fig:sen2} showcases a change in cost from 0 to 1 (per GB). Interestingly, for VQA, we found that the unit network costs actually have minor effects on the execution plans and the plan changes are subtle -- this is due to a relatively higher compute cost on the cloud, so the computation is kept at the edge. Meanwhile, on AICity, we use blue dots to show where the plan changes, though the total serving cost is near linear. We present actual query plans in Figure~\ref{fig:case_study_2} to show an example plan change when the network cost is reduced by 90\% and compute is shifted to higher-tier workers.

\begin{figure}
    \centering 
    \begin{subfigure}{0.48\columnwidth}
      \includegraphics[width=\linewidth]{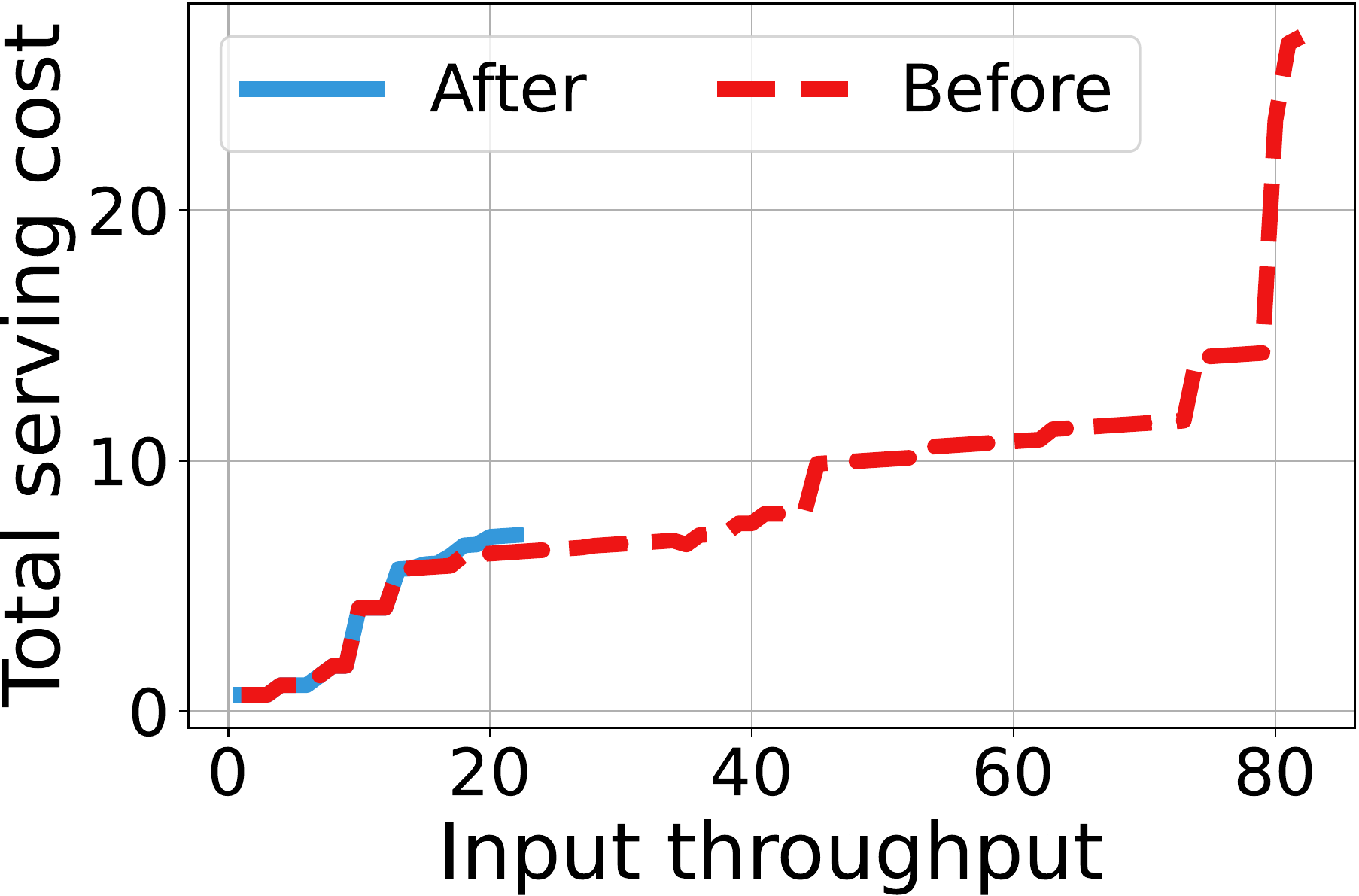}
      \caption{}
      \label{fig:sen3-a}
    \end{subfigure}
    \begin{subfigure}{0.48\columnwidth}
      \includegraphics[width=\linewidth]{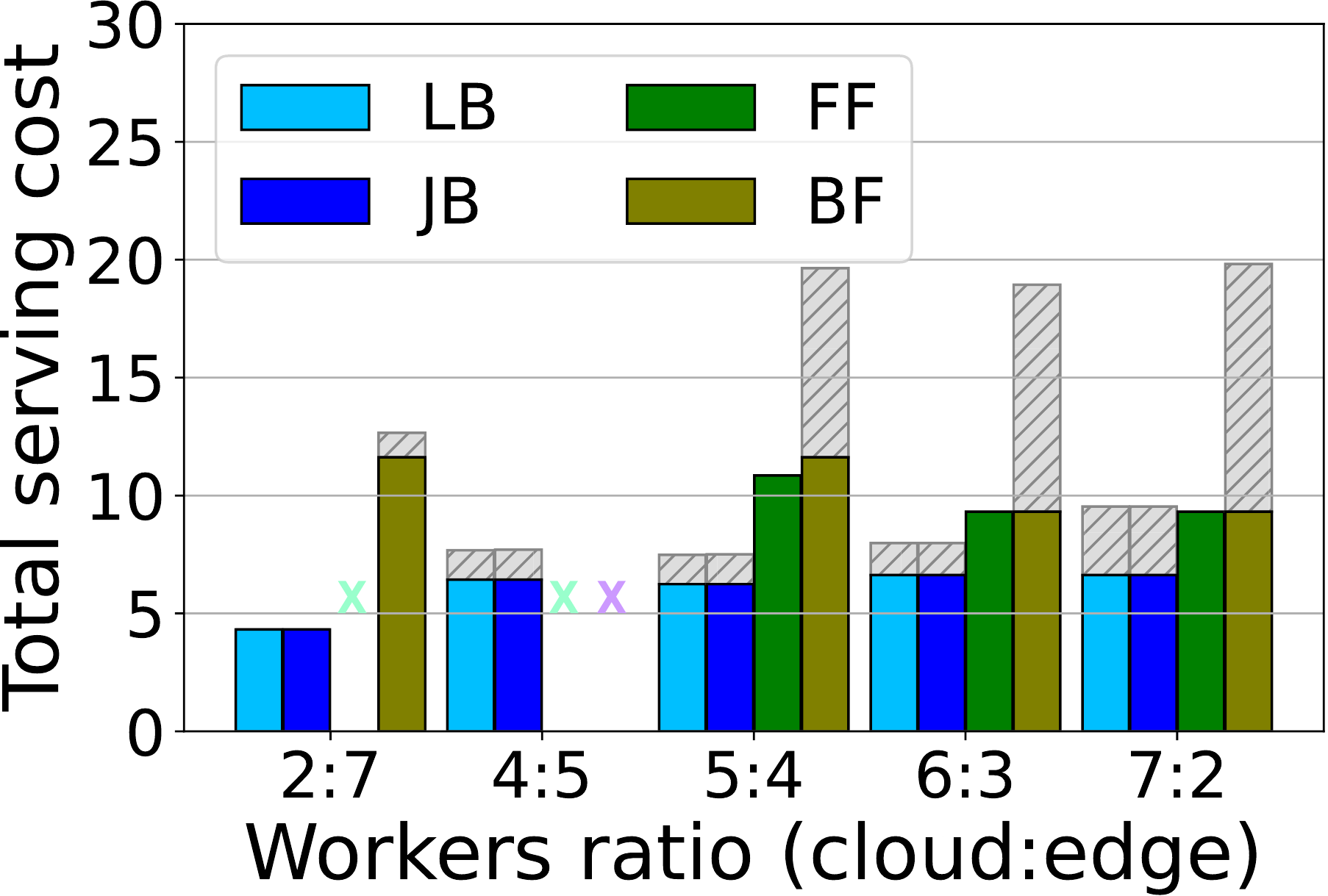}
      \caption{}
      \label{fig:sen3-b}
    \end{subfigure}\vspace{-0.1in}
\caption{Sensitivity study on VQA for (a) limiting total outbound bandwidth at 10Mbps on edge devices with the  \texttt{medium} setup, changing input throughput and (b) changing the worker ratios on different tiers; see text for details. `X' indicates unsolvable inputs.}
\label{fig:sen3}
\end{figure}

\vspace{0.05in}
\noindent\textbf{Effect of infrastructure changes.} We examine the flexibility of {\name} when there are additional resource caps on the \texttt{medium} setup. Specifically, we placed a 10Mbps bandwidth constraint over all edge devices, mimicking real-world scenarios with limited networking. The {\name} optimizer simply applies an additional constraint and limits the search space; there is no change on the processor and execution engine. Figure~\ref{fig:sen3-a} illustrates the results and our findings. To reduce network costs, the {\name} optimizer uses more compute resources on the edge. The blue curve ends early since no viable solution can be found.

We change the number of workers allocated to different tiers, and observe how the total serving cost changes (Figure~\ref{fig:sen3-b}). Since there are 9 total workers in the \texttt{medium} setting, we rank them according their cost and place those with higher costs on higher tiers (and vice versa). For instance, in the 5:4 case, the edge has 1x2, 1x4, 2x8 workers, and the cloud has 1x16, 1x48, 3xV100. Results show that {\name} successfully finds good execution plans in all the settings; more cloud resources does increase the network costs. 

\vspace{0.05in}
\noindent\textbf{Effect of profiling.} In our previous experiments in ~\xref{sec62}, we leveraged the profiling of P75 percentiles as inputs to our QO and report actual runtime numbers; doing so gives extra room for the QO to find valid plans. In Figure~\ref{fig:end_to_end}, we also explore the estimated costs using P50 and P90 efficiency profiles on the error bars. We observe a small variance -- the the actual runtime using P75 in most cases falls in the middle of estimates using P50 and P90. In some cases, the optimizer chooses different plans which leads to discontinuity of the costs. Overall, we observe that this has minor effects on our end-to-end solution.

\section{Discussion}
\label{sec:discussion}
\noindent\textbf{Obtaining diverse model choices.}
The user optionally provides a list of model choices for each operator in the workflow. Our current prototype depends on this provided model choices. However, in the future {\name} can also enrich the choices using off-the-shelf model quantization, pruning, and distillation tools. Several tools already exist today, and it is an active area of research in ML~\cite{jacob2018quantization, cheng2017survey, polino2018model,liu2017learning,sanh2020movement,esser2019learned} in order to democratize ML on weak edge devices.
To integrate these tools into {\name}, we can simply invoke them to derive cheaper models offline (similar to how we profile models for their accuracy profiles). We acknowledge that running these tools may require us to have access to the original training data and labels.

\vspace{0.05in}
\noindent\textbf{Limitations}. As discussed in~\xref{sec:qo} and~\xref{sec:qp}, we used one-to-one mapping between the workers and operators. Using one-to-multiple mapping to consolidate the operators may further improve the performance and can be an interesting further work to explore. Doing so may require automatic grouping of the operators. Nevertheless, we have shown in~\xref{sec62} that our processor already has low overhead. 

{\name} also assumes the heterogeneous infrastructures to have near constant input requests on the edge devices; this is true for the use cases discussed in~\xref{sec:bg} and in our experiments. Exploring use cases that do not fall into this category, such as security sensors or cluster telemetries which send only intermittent signals, can be an interesting future work. Besides, we used one-time profiling and fixed worker costs in our experiments; quickly adapting to changes in these aspects can also improve the usability of our system.
\section{Related Work}
\label{sec:related}

\vspace{0.05in}
\noindent\textbf{Edge-cloud systems.}
Moving compute to the edge can reduce the networking cost and is used in video analytics to eliminate the need to transfer raw video streams.
Chameleon~\cite{jiang2018chameleon} leverages temporal and spatial correlations to tune frame resolution, sampling rate, detector model configurations for an optimal resource-accuracy trade-off. In~\cite{wang2020joint}, a latency and energy consumption model is considered for choosing the configuration. Jain et al.~\cite{jain2019scaling} scale video analytics to large camera deployments using hand-crafted rules that leverage cross-camera correlation to improve cost efficiency and accuracy. Elf~\cite{zhang2021elf} applies a content-aware approach to offload smaller inference tasks in parallel to edge servers. These works considered a simple edge-cloud infrastructure and used workload-specific optimization techniques. We support optimizing and running arbitrary ML workflows on a wide range of infrastructures, both of which are inputs to our optimizer.

\vspace{0.05in}
\noindent\textbf{ML inference systems.}
Serving machine learning inference has attracted great attention. TensorFlow Serving~\cite{olston2017tensorflow} is one of the first serving systems for production environments. Clipper~\cite{crankshaw2017clipper} maximizes throughput under a user-specified latency service-level objective (SLO) and enables the use of different frameworks by hosting the model in a container. Model selection policies are also integrated to provide different cost-accuracy trade-offs. Nexus~\cite{shen2019nexus} automatically chooses the optimal batch size and the number of GPUs to use according to the request rate and latency SLO for a given model. Model DAGs are also considered in other works~\cite{crankshaw2020inferline, romero2019infaas, triton, romero2021llama, hu2021rim, hu2021scrooge}. Rim~\cite{hu2021rim} considers serving ML workflows at a cluster of edge GPUs. {\name} differs in two ways. First, we choose individual models based on input throughput and target accuracy for the entire ML workflow. Second, we target at deploying ML workflows on heterogeneous infrastructures, where prior works focused on either: a) homogeneous cloud datacenters or edge devices only, or b) heterogeneity within a single tier (i.e., datacenter).

\vspace{0.05in}
\noindent\textbf{Optimizing ML queries}
A number of works have been proposed in optimizing ML queries at either logical- or physical-level. Lu et al.~\cite{lu2018accelerating} filter data that does not satisfy the query predicate by using probabilistic predicates. BlazeIt~\cite{kang2018blazeit} optimizes aggregation and limit queries for videos. Yang et al.~\cite{yang2022optimizing} exploit predicate correlations to build proxy models online to avoid exhaustive offline filter construction. Optimization at physical execution-level is addressed in some of the ML serving systems that support model DAGs. For instance, Llama~\cite{romero2021llama} applies a greedy strategy that chooses cost-efficient worker configurations for video analytics pipelines.
These works did not consider network cost, because 
these systems target pure datacenter deployment scenarios.
{\name} optimizes general ML workflows jointly at logical and physical levels for a heterogeneous infrastructure across edges and the cloud.

\section{Conclusions}
The rise of smart home devices and the Internet of Things 
opens up the opportunity  for ML serving systems at the level of both the infrastructure and ML workflow to explore new trade-offs between accuracy and performance.
We build {\name}, an ML serving to optimize ML workflows which takes into account the cost, availability, and performance of the increasingly tiered and heterogeneous infrastructures.
{\name} significantly reduces the total serving cost of visual question answering and vehicle tracking from the NVIDIA AI City Challenge compared with state-of-the-art solutions.
{\name} will be open-sourced.

\bibliographystyle{ACM-Reference-Format}
\bibliography{confs_long, reference}

\appendix


\newpage
\section{Appendix}

\begin{figure}
	\begin{subfigure}{0.45\linewidth}
      \includegraphics[width=\linewidth]{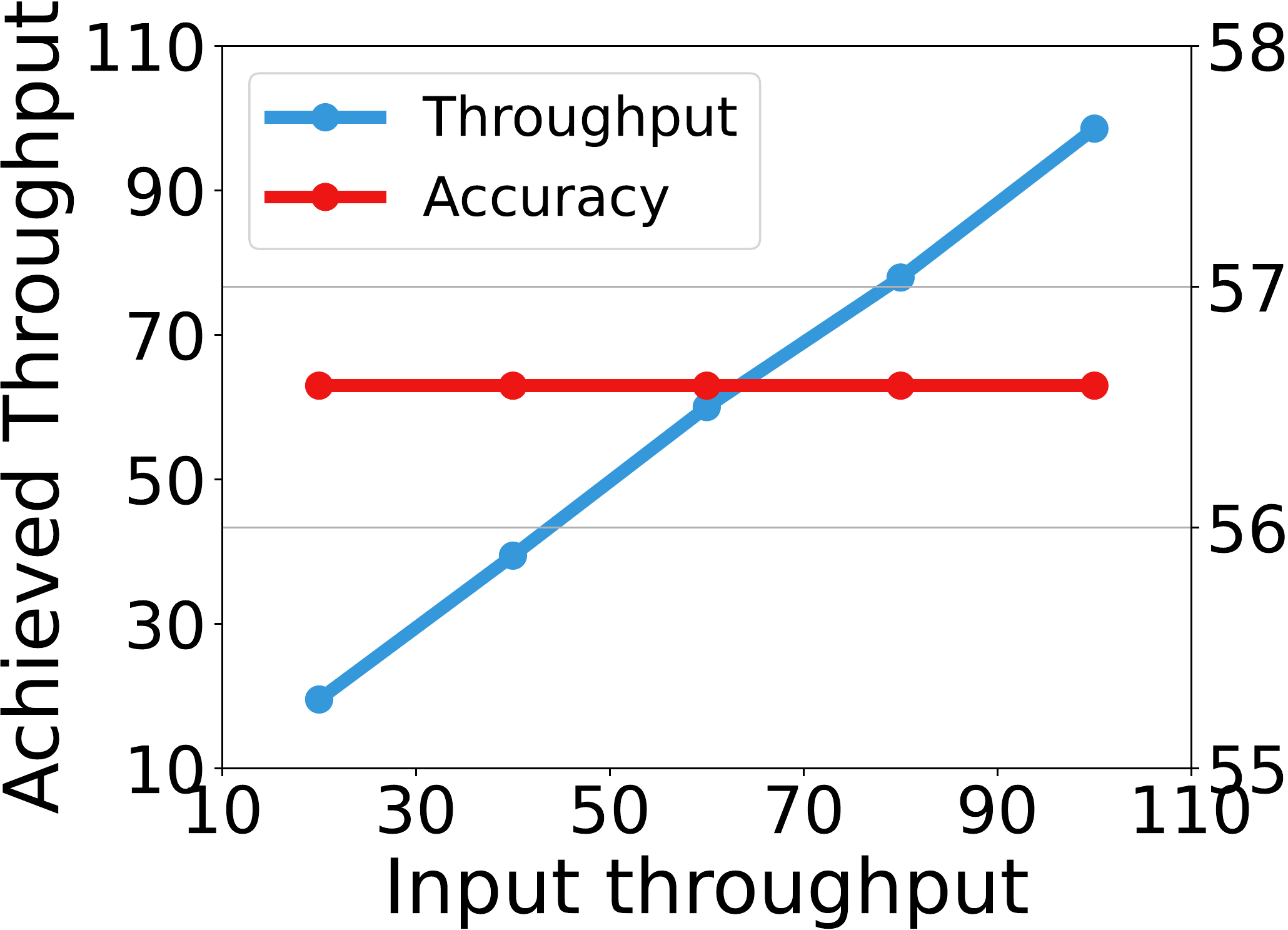}
       \caption{VQA}
    \end{subfigure} \hfil
	\begin{subfigure}{0.45\linewidth}
      \includegraphics[width=\linewidth]{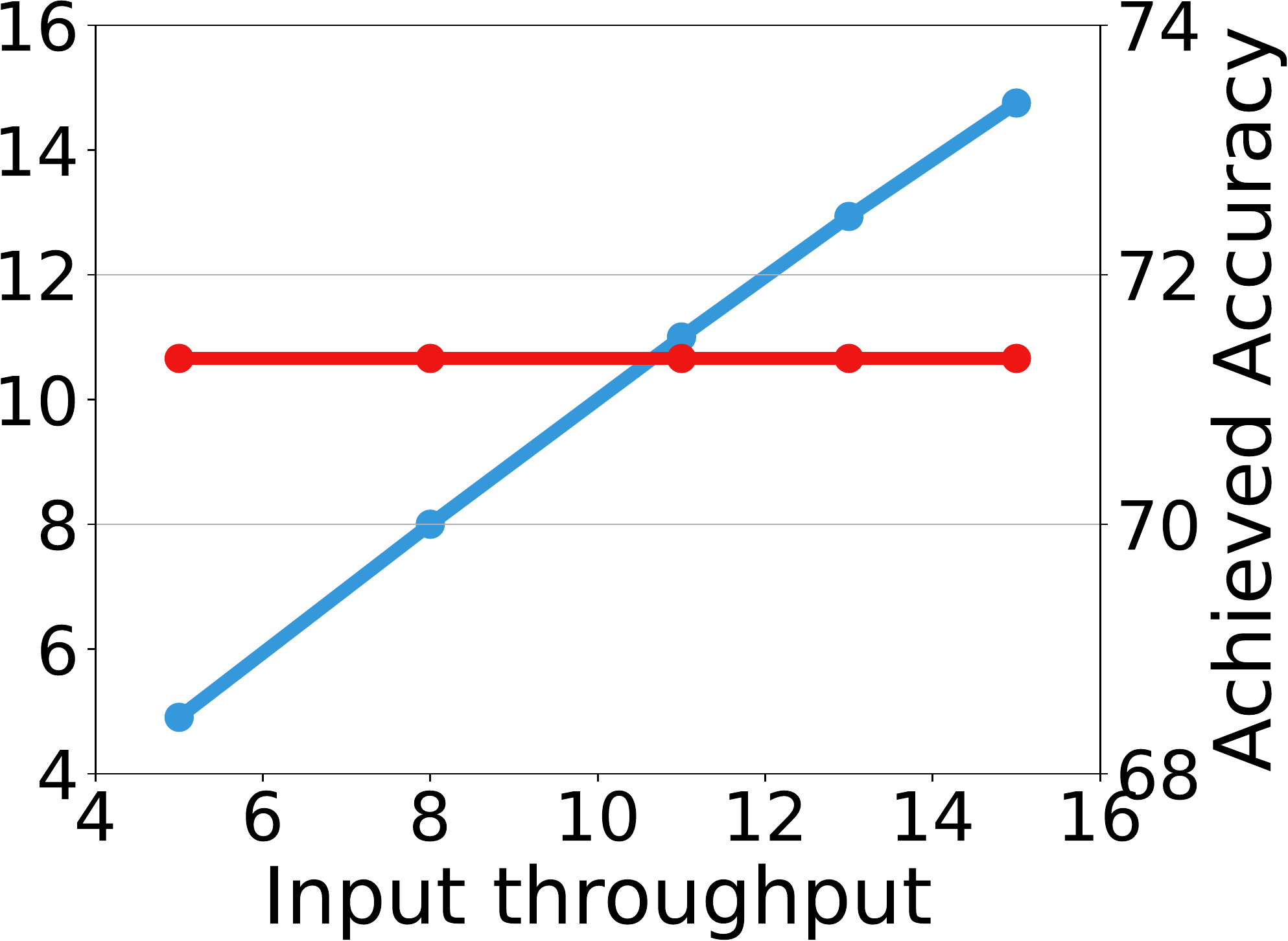}
      	  \caption{AICity}
    \end{subfigure}\vspace{-0.1in}
\caption{Achieved throughput and accuracy given input throughput on the \texttt{large} setup.}
\label{fig:extended_actual}
\end{figure}

\subsection{System Evaluation}
\label{sec:app_se}
We illustrate the actual throughput and achieved accuracy with varying input throughput on the \texttt{large} setting in Figure~\ref{fig:extended_actual}. Again, we find that \name's actual throughput mostly reaches the target throughput. In some cases, the achieved throughput is slightly lower than the target (<2\% lower). This might be contributed by the performance variation of the vCPUs on IBM cloud at different time-of-day, as we observed.

Here we also report the runtime variance on the \texttt{xlarge} setup. The standard deviation of the serving cost across 5 runs is 0.122\% for AICity, and 0.028\% for VQA.

\begin{figure}
	\begin{subfigure}{0.475\linewidth}
	\centering
      \includegraphics[width=\linewidth]{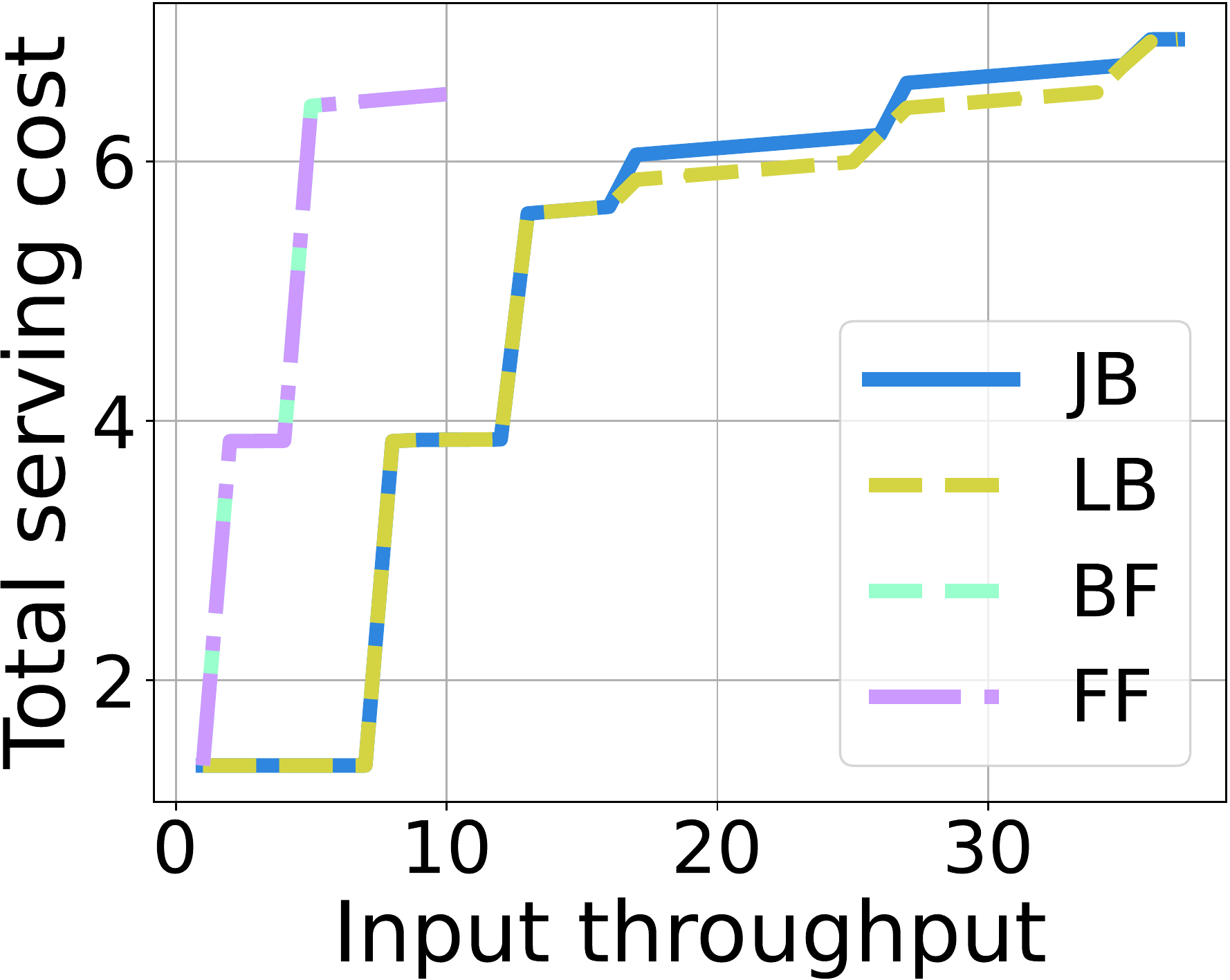}
     \caption{VQA}
    \end{subfigure} \hfil	
	\begin{subfigure}{0.475\linewidth}
	\centering
      \includegraphics[width=\linewidth]{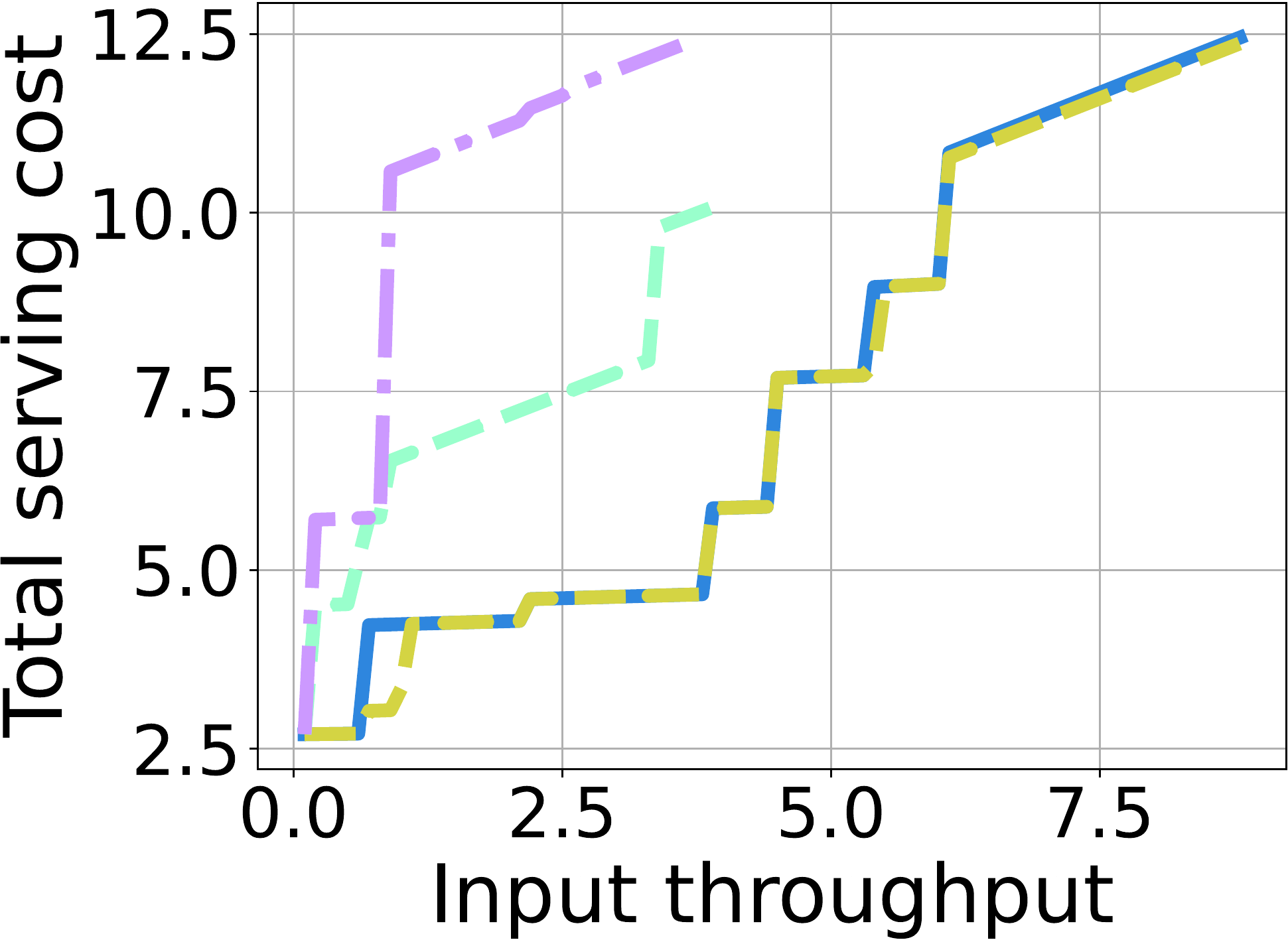}
	  \caption{AICity}
    \end{subfigure}\vspace{-0.1in}
\caption{Total serving cost w.r.t. input throughput in \texttt{JB} on the \texttt{small} setup.}
\label{fig:extended_ab1_a}
\end{figure}

\begin{figure}
	\begin{subfigure}{0.475\linewidth}
      \includegraphics[width=\linewidth]{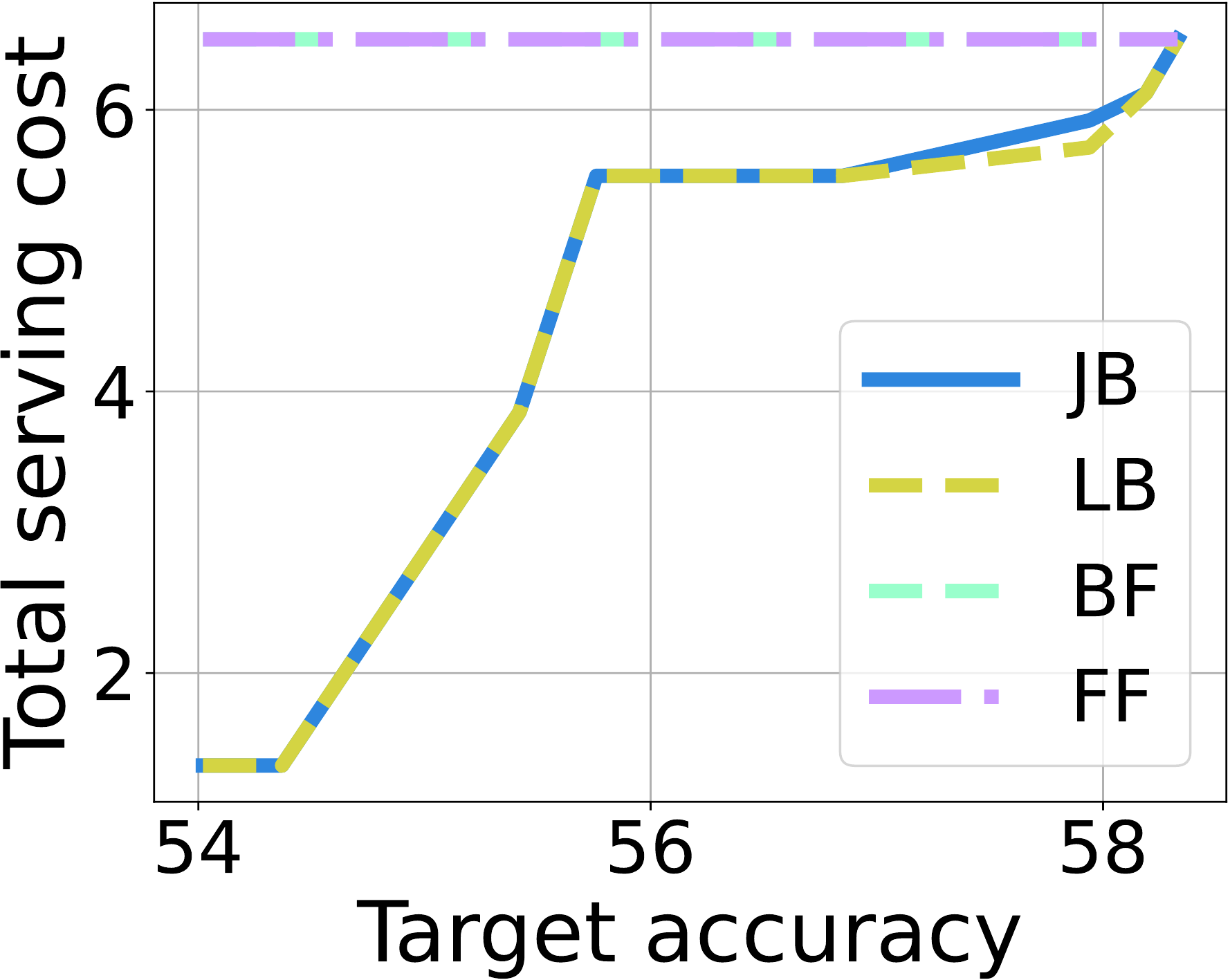}
          \caption{VQA}
    \end{subfigure} \hfil
	\begin{subfigure}{0.475\linewidth}
      \includegraphics[width=\linewidth]{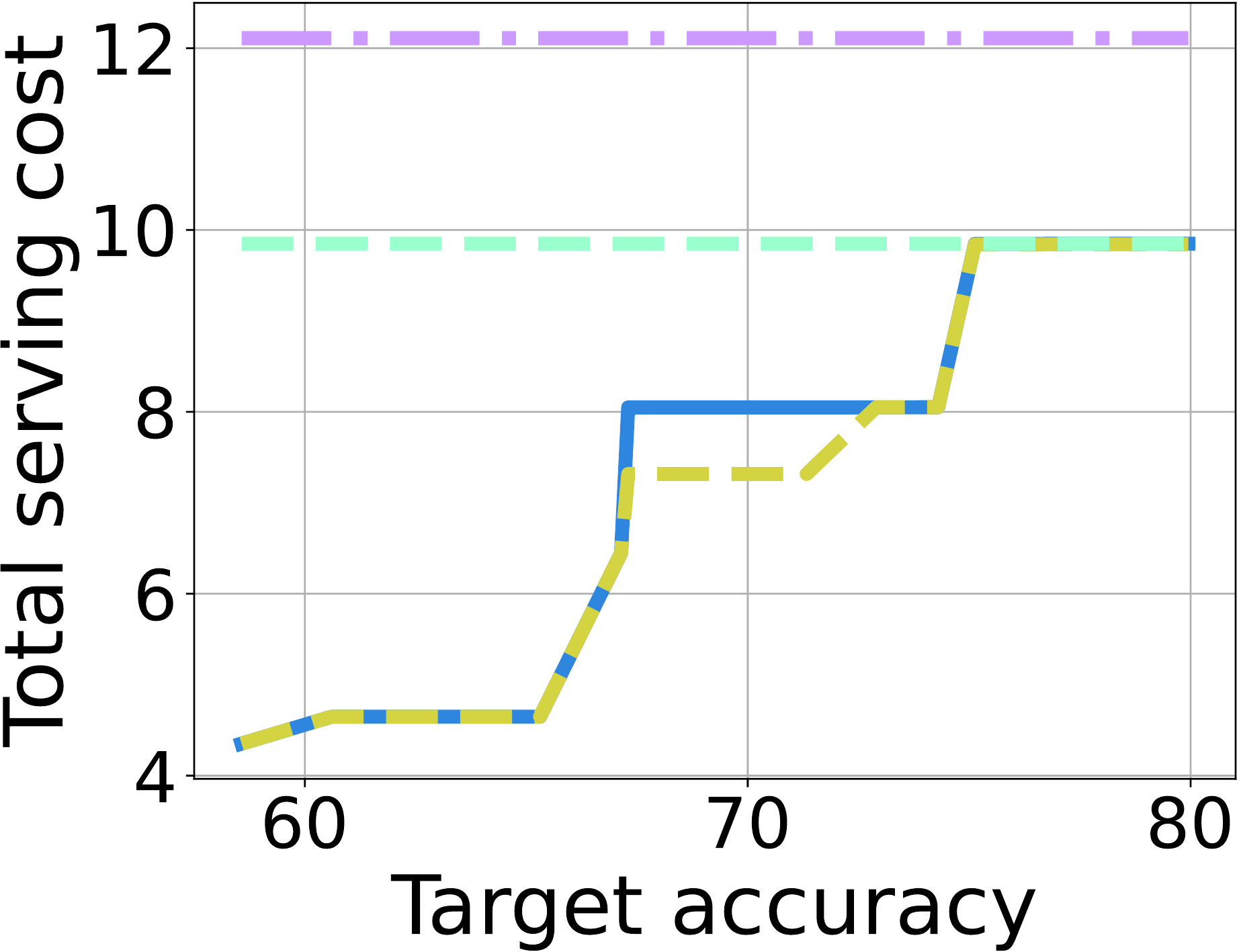}
     	  \caption{AICity} 
    \end{subfigure}\vspace{-0.1in}
\caption{Total serving cost w.r.t. target accuracy in \texttt{JB} on the \texttt{small} setup.}
\label{fig:extended_ab2_a}
\end{figure}

\begin{figure}
	\begin{subfigure}{0.475\linewidth}
	\centering
      \includegraphics[width=\linewidth]{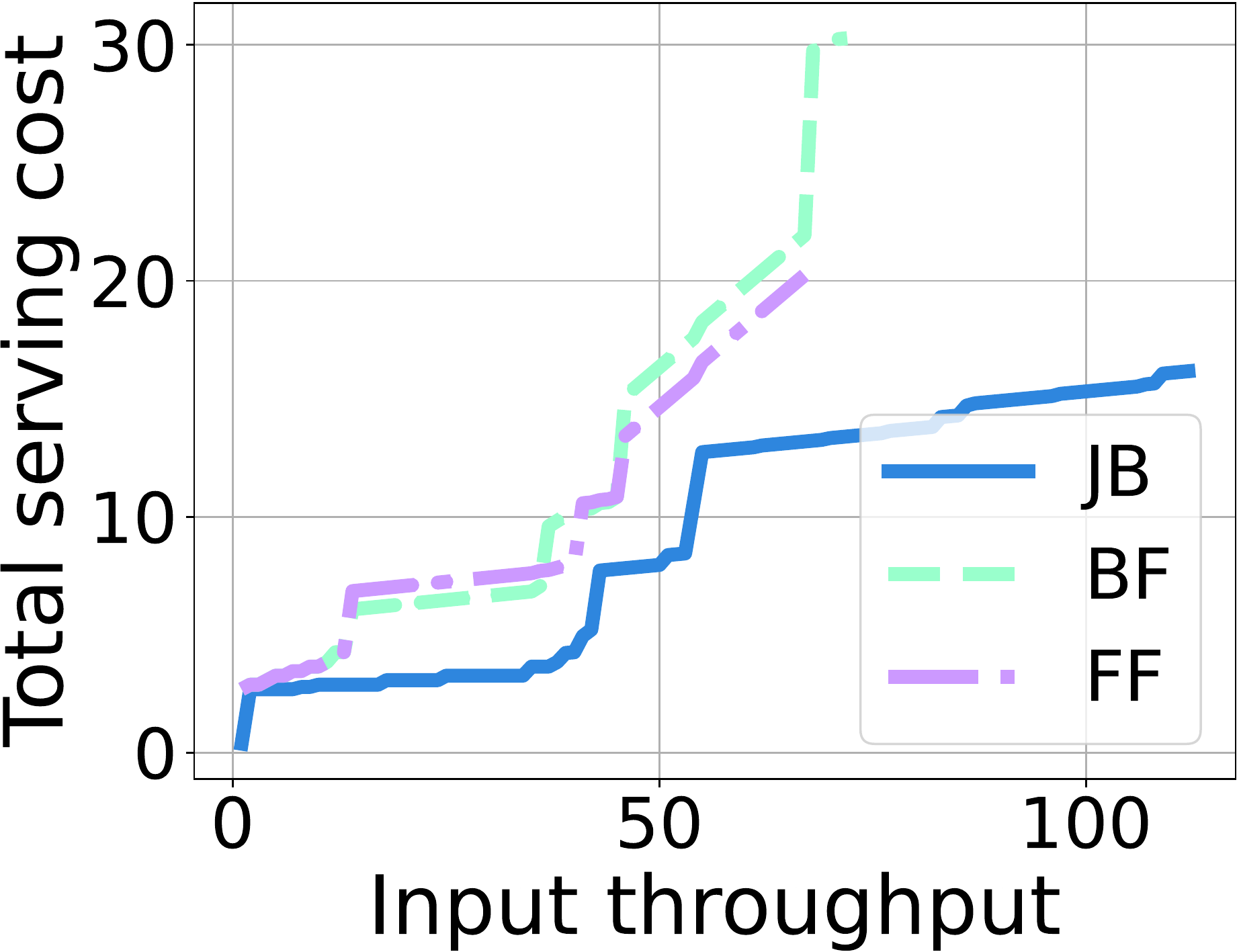}
     \caption{VQA}
    \end{subfigure} \hfil	
	\begin{subfigure}{0.475\linewidth}
	\centering
      \includegraphics[width=\linewidth]{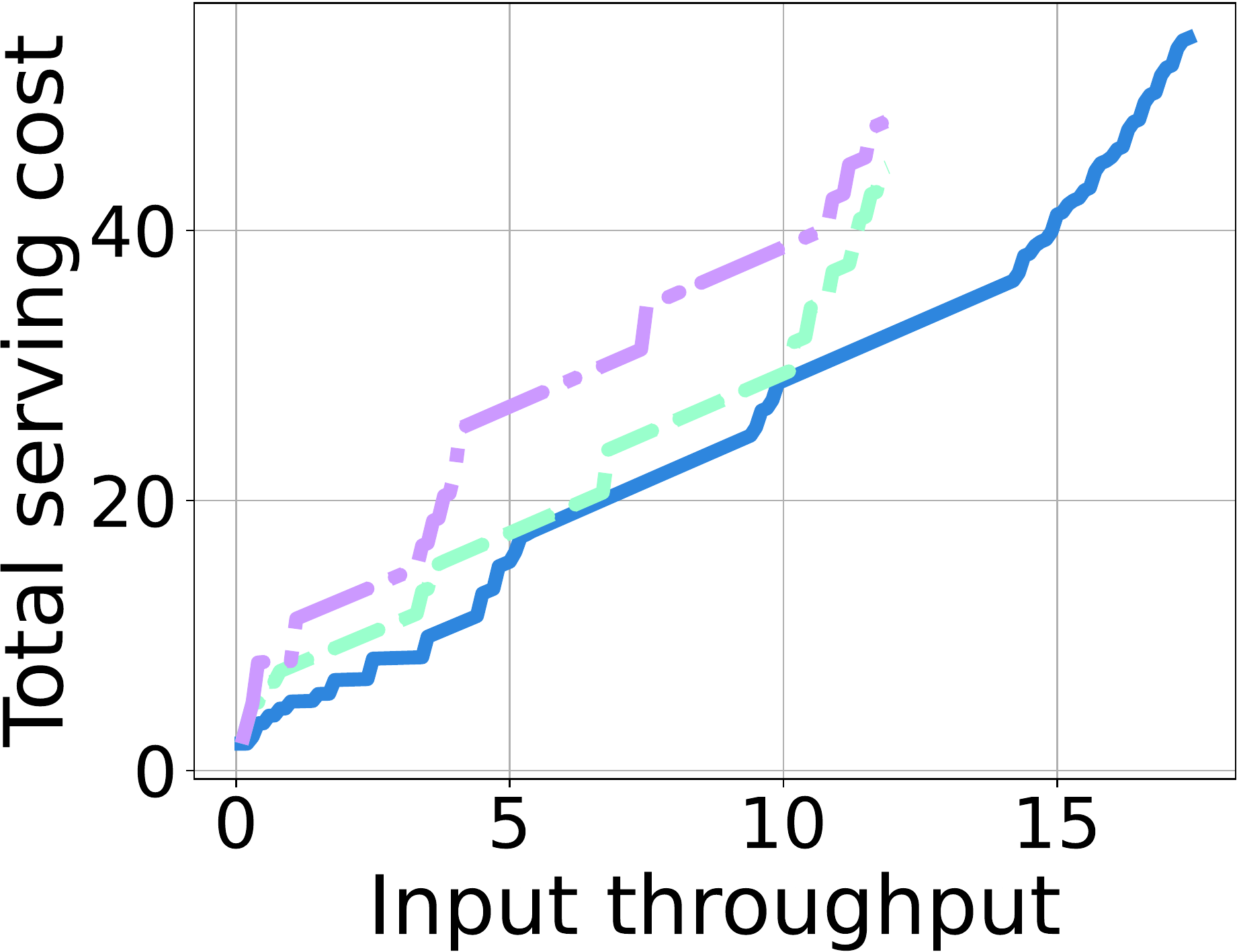}
	  \caption{AICity}
    \end{subfigure}\vspace{-0.1in}
\caption{Total serving cost w.r.t. input throughput in \texttt{JB} on the \texttt{large} setup.}
\label{fig:extended_ab1_c}
\end{figure}

\begin{figure}
	\begin{subfigure}{0.475\linewidth}
      \includegraphics[width=\linewidth]{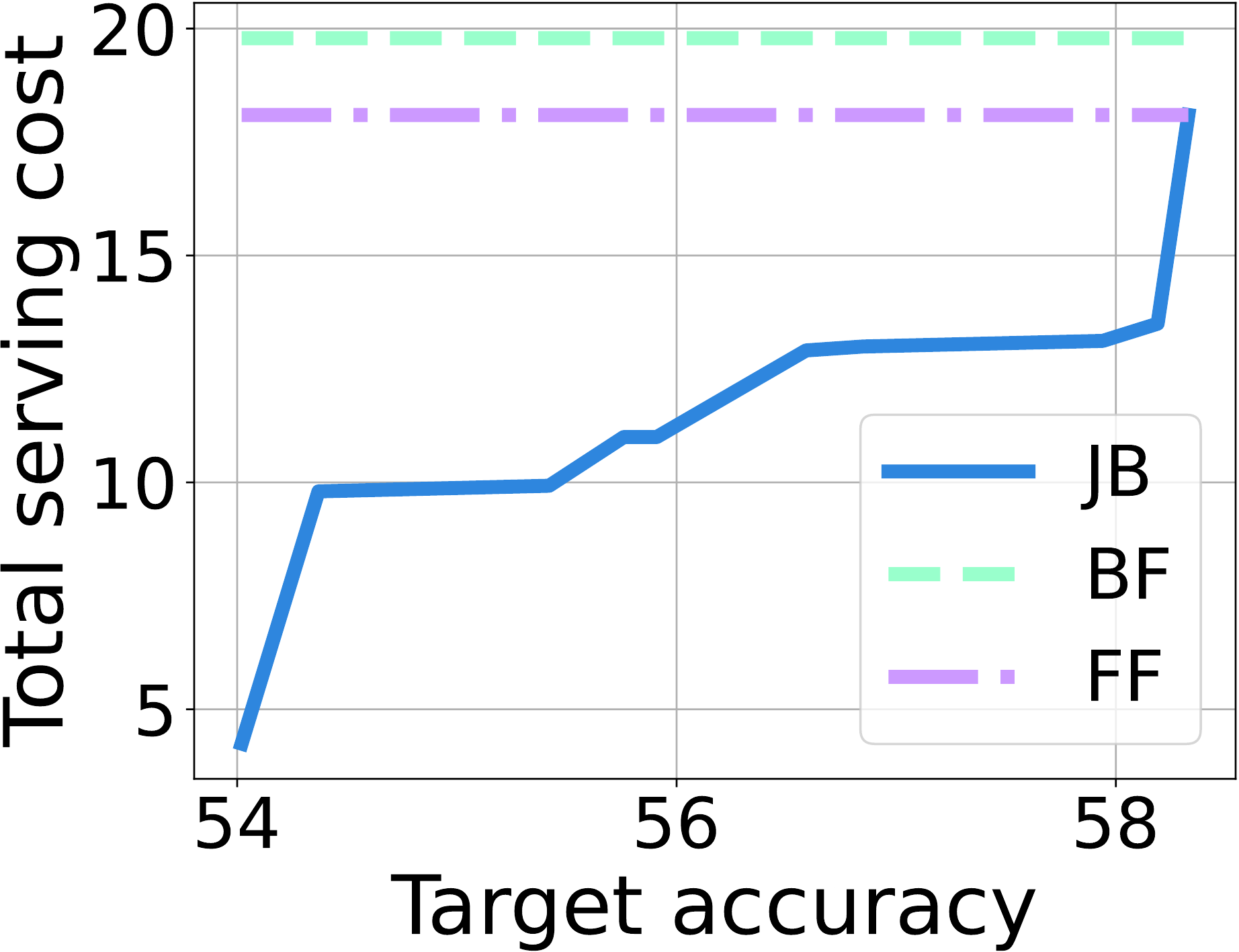}
          \caption{VQA}
    \end{subfigure} \hfil
	\begin{subfigure}{0.475\linewidth}
      \includegraphics[width=\linewidth]{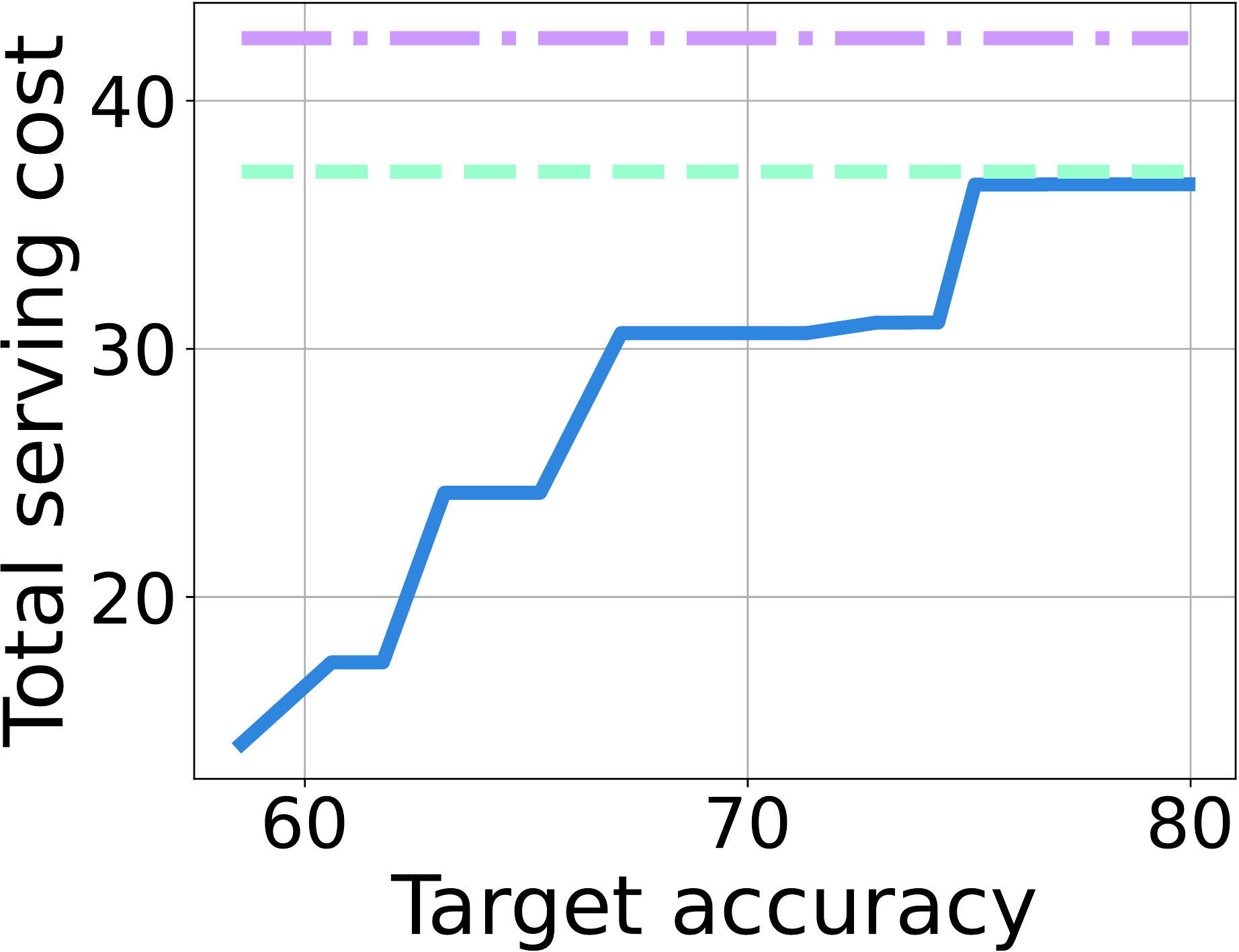}
     	  \caption{AICity} 
    \end{subfigure}\vspace{-0.1in}
\caption{Total serving cost w.r.t. target accuracy in \texttt{JB} on the \texttt{large} setup.}
\label{fig:extended_ab2_c}
\end{figure}

\subsection{Ablation Study}
\label{sec:app_ab}
Here we demonstrate how the serving costs change with respect to input throughput and target accuracy for \name and the compared baselines in \texttt{small} and \texttt{large} setups. The results on the \texttt{small} setup are shown in Figure~\ref{fig:extended_ab1_a} and ~\ref{fig:extended_ab2_a}, while the results on the \texttt{large} setup are shown in Figure~\ref{fig:extended_ab1_c} and \ref{fig:extended_ab2_c}. We do not report the serving costs for \texttt{LB} on the \texttt{large} setup, as they could not find the solution in a reasonable amount of time. From the figures, we observe that \texttt{JB} consistently outperforms \texttt{BF} and \texttt{FF}, and achieves the same or similar serving costs compared with \texttt{LB} on the \texttt{small} setup.

\begin{table}
    \caption{Ablation analysis of model selection on the VQA dataset.}
	\resizebox{0.98\columnwidth}{!}{
    \begin{tabular}{cccccccc}
        \toprule
        \multicolumn{2}{c}{Model} & \multicolumn{3}{c}{\texttt{medium}} & \multicolumn{3}{c}{\texttt{large}} \\
        \cmidrule(lr){3-5} 
        \cmidrule(lr){6-8}
        Select. & Assign. & Comp & Net & QO & Comp & Net & QO \\
        \midrule
        JB & JB & 6.2 & 1.3 & 20.8 ms & 11.3 & 1.6 & 29.3ms\\
		Most acc. & JB & 9.1 & 8.8 & 2.3ms & 12.1 & 6.0 & 3.1ms\\
		Brute f. & JB & 6.2 & 1.3 & 21.3ms & 11.3 & 1.6 & 30.0ms\\
        \midrule
		JB & PTc & 4.0 & 12.2 & N/A & 6.0 & 18.4 & N/A \\
		JB & SPc & 15.8 & 12.2 & N/A & 23.7 & 18.4 & N/A \\
		Most acc. & PTc & 5.8 & 12.2 & N/A & 8.8 & 18.4 & N/A \\
		Most acc. & SPc & 19.3 & 12.2 & N/A & 29.0 & 18.4 & N/A\\ 
        \bottomrule
    \end{tabular}
	}
    \label{tab:extended_ablation}
\end{table}

We also showcase the effect of model selection on the VQA dataset in Table~\ref{tab:extended_ablation}. The results again demonstrate the effectiveness of \texttt{JB}'s model selection strategy, as well as its effectiveness with other ML runtimes.

\subsection{Discussion on the Performance of \name}
\label{appendix:perf}

\begin{figure}
    \centering 
    \begin{subfigure}{0.48\columnwidth}
      \includegraphics[width=\linewidth]{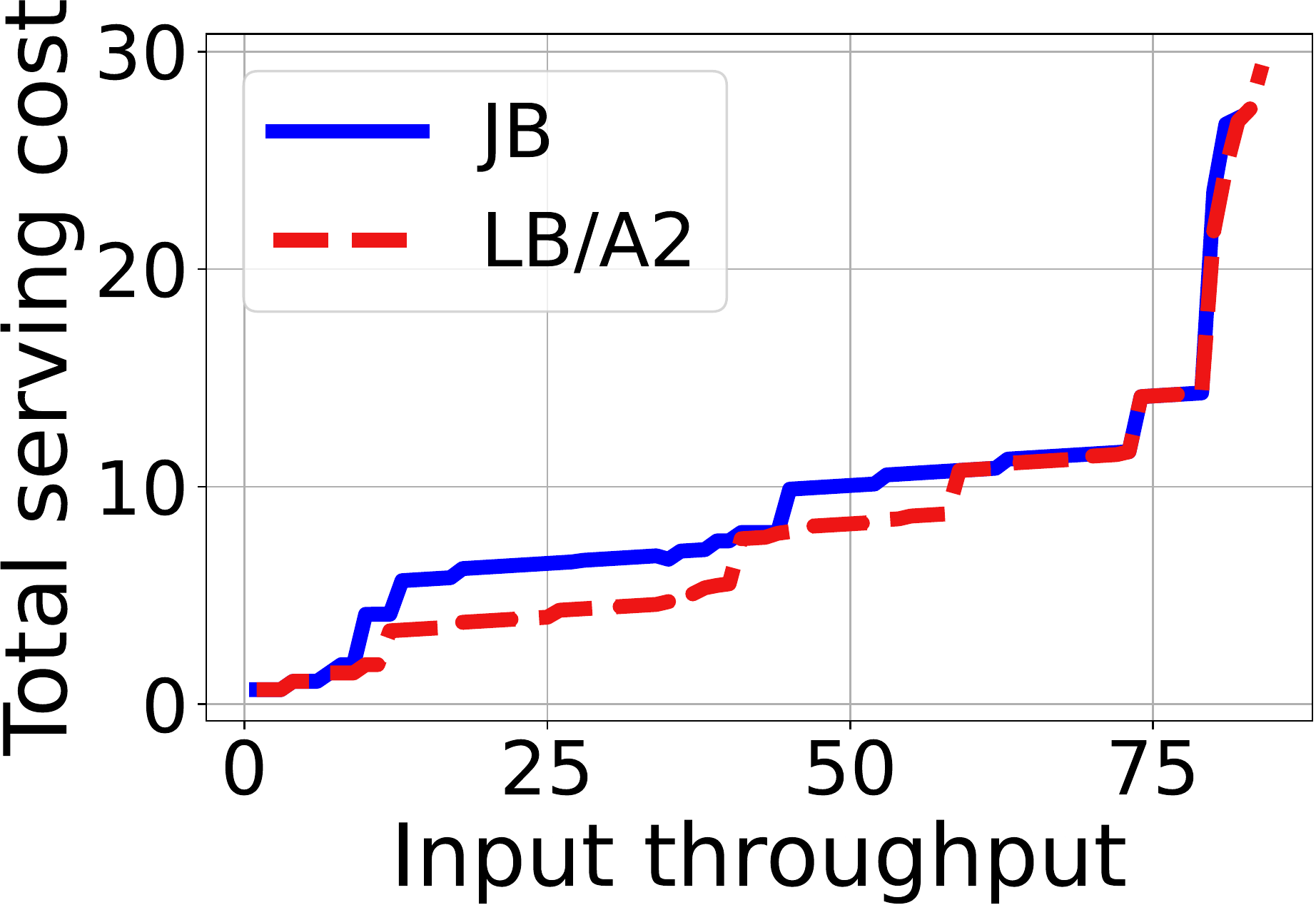}
      \caption{Remove Assumption A2}
      \label{fig:error_analysis_tput}
    \end{subfigure} \hfil
    \begin{subfigure}{0.48\columnwidth}
      \includegraphics[width=\linewidth]{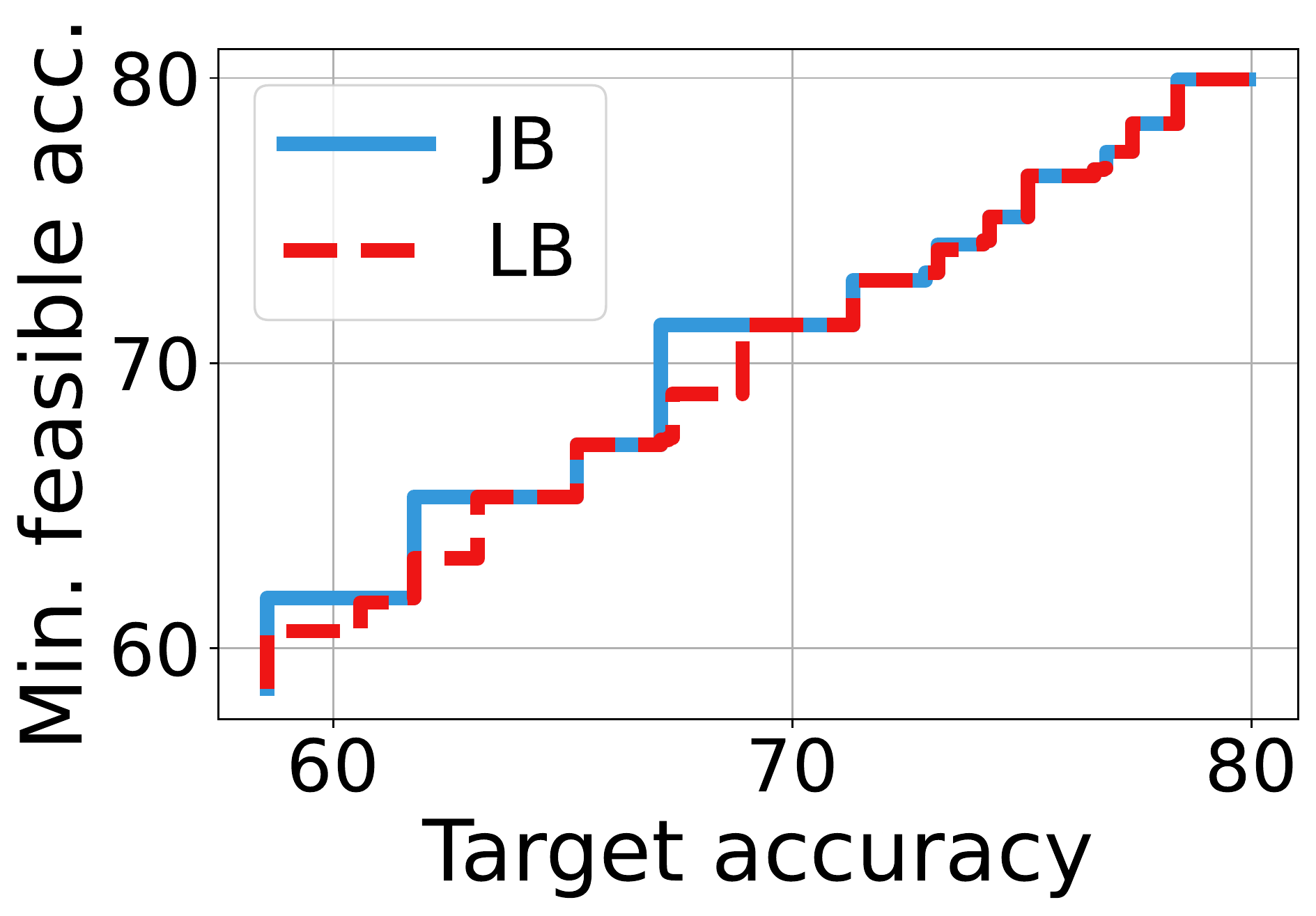}
      \caption{\texttt{JB}'s model selection}
      \label{fig:error_analysis_ms}
    \end{subfigure}\vspace{-0.1in}
\caption{To demonstrate failure cases in \name, we compare the total serving cost of \texttt{JB} with \texttt{LB} when Assumption A2 is relaxed. We also illustrate the minimal accuracy of the feasible model assignments found \texttt{JB}'s model selection algorithm.}
\label{fig:error_analysis}
\end{figure}

\begin{figure}
    \centering 
     \includegraphics[width=\linewidth]{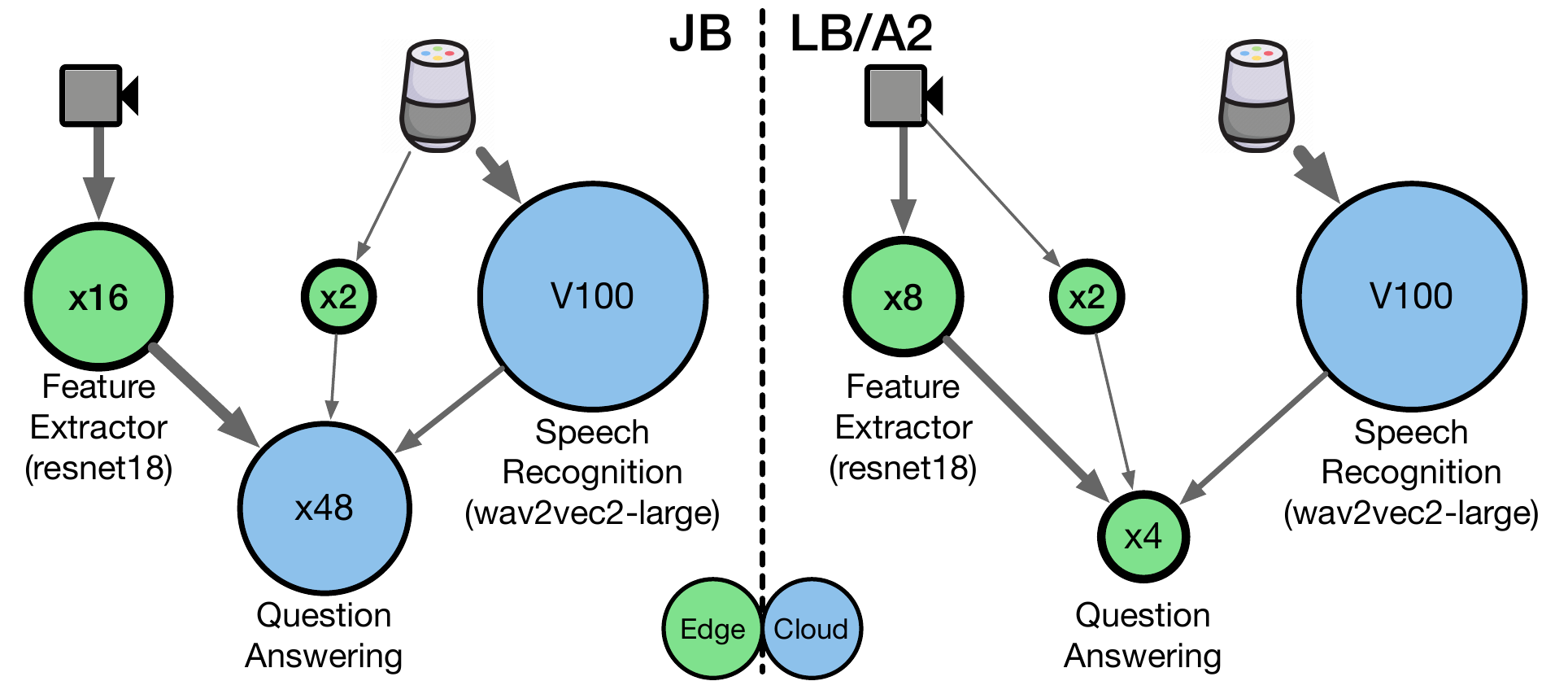}
\caption{Comparison of the execution plans of JB and LB on VQA using the \texttt{medium} setup modified to 25 rps, when Assumption A2 is removed.}
\label{fig:case_4}
\end{figure}

We employ greedy approaches in the \name query optimizer for both model selection and worker assignment.
In this section, we further analyze: 1) why \name often provides execution plans that are competitive with the lower bound in practical configurations, and 2) properties of configurations for which \name may generate worse execution plans. 

\subsubsection{Why \name performs well in practice}

For most cases in Section~\xref{s:eval}, \texttt{JB} has a total serving cost close to \texttt{LB}, even though \texttt{JB} utilizes a greedy strategy.
The small performance gap is due in part to the workflow properties and infrastructure configurations. 
On the infrastructure side, lower infrastructure tiers generally have fewer, less-capable workers than higher tiers; additionally, lower tiers have a reduced communication cost compared to higher tiers due to their proximity to the data sources. With Assumption A2, our worker assignment algorithm starts to assign workers from lower tiers considering both compute and communication costs. 
On the workflow side, the output accuracy of ML operators generally increases monotonically with respect to the improvement of the input quality; our beam-search leverages this property to find feasible model assignments.

\begin{figure}
    \centering 
    \begin{subfigure}{0.48\columnwidth}
      \includegraphics[width=\linewidth]{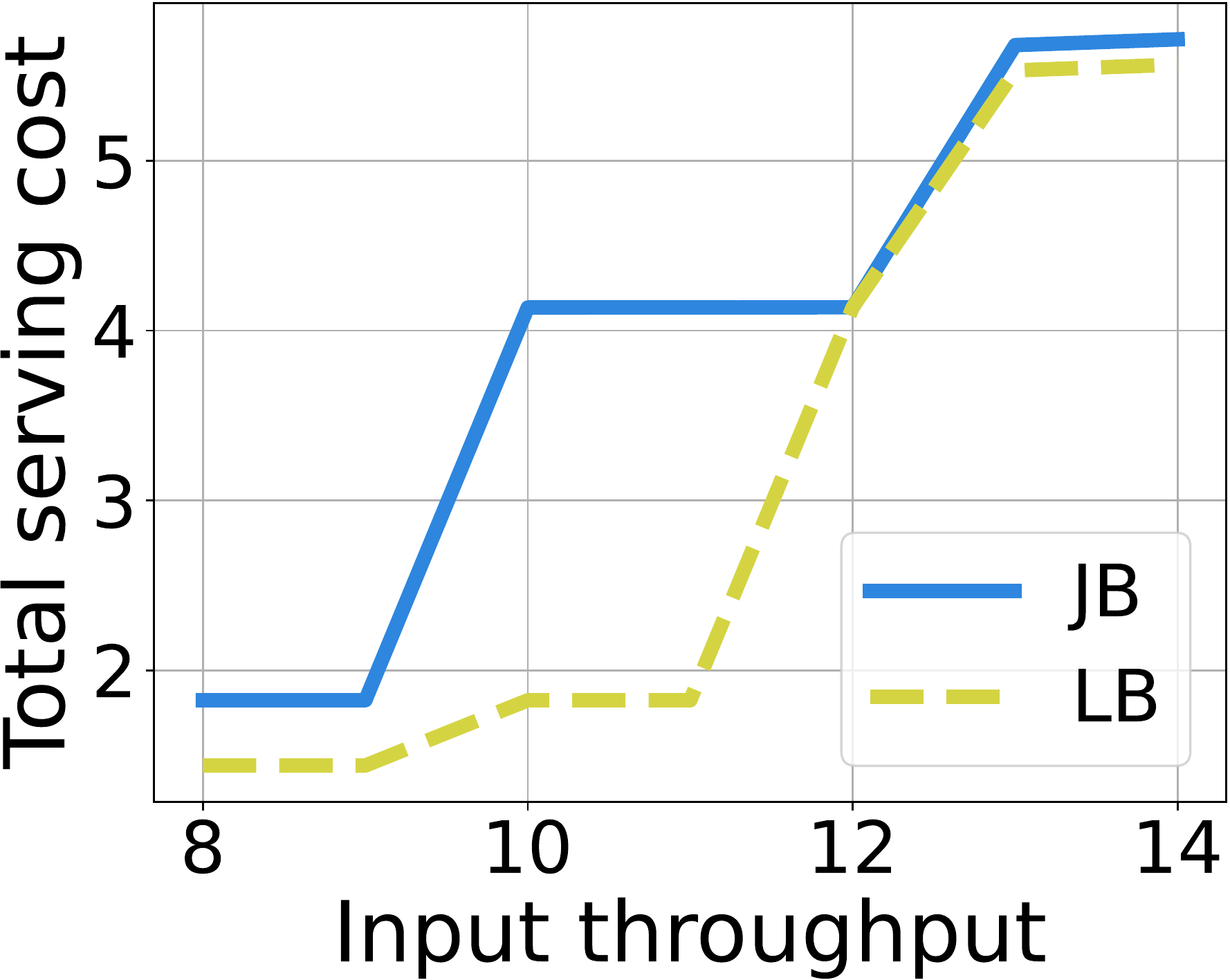}
      \caption{VQA}
    \end{subfigure} \hfil
    \begin{subfigure}{0.48\columnwidth}
      \includegraphics[width=\linewidth]{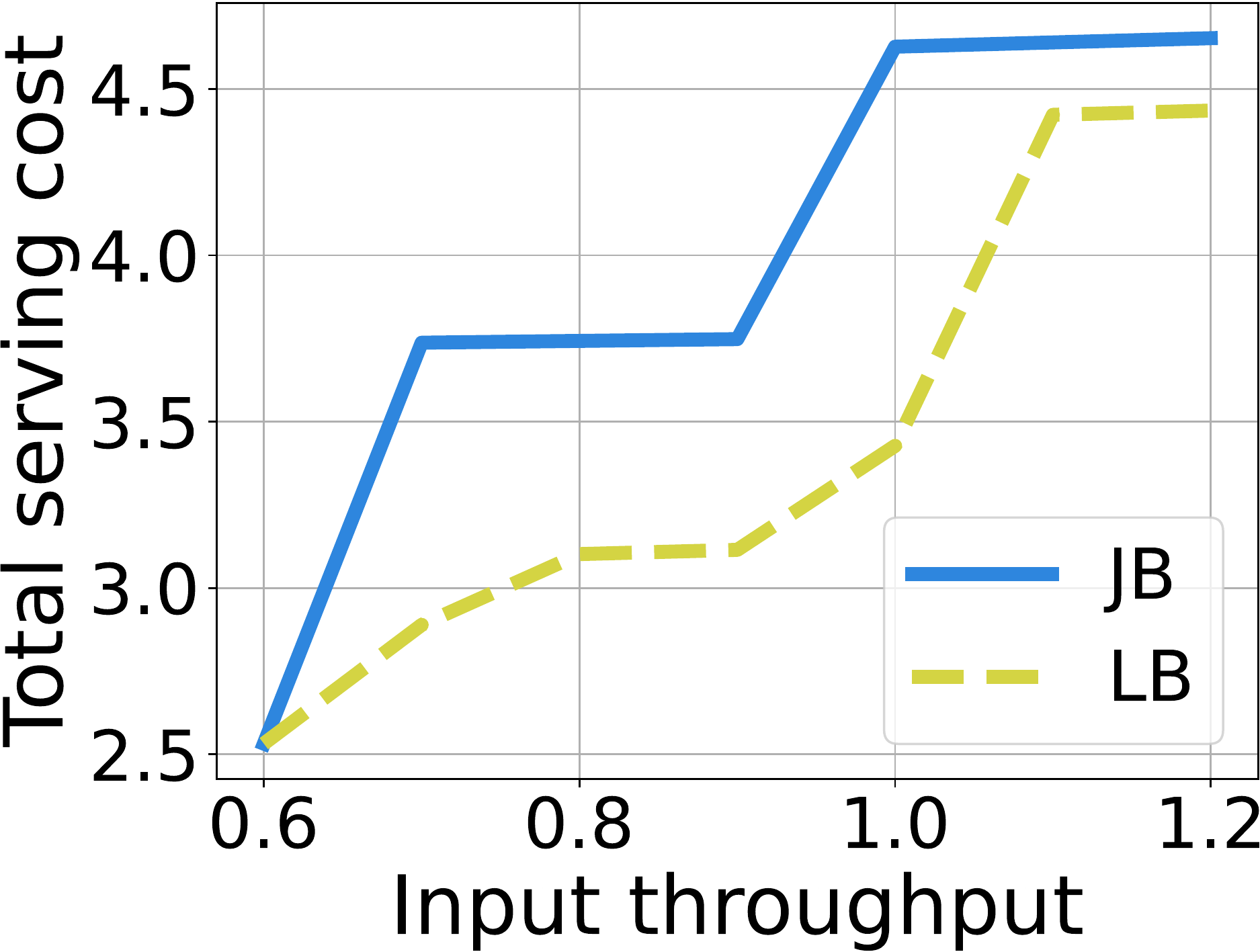}
      \caption{AICity}
    \end{subfigure}\vspace{-0.1in}
\caption{Zoom-in of Figure~\ref{fig:ab1} over the regions \texttt{JB} has a noticeably higher cost than \texttt{LB}}
\label{fig:cost_zoomed}
\end{figure}

\begin{figure}
    \centering 
     \includegraphics[width=\linewidth]{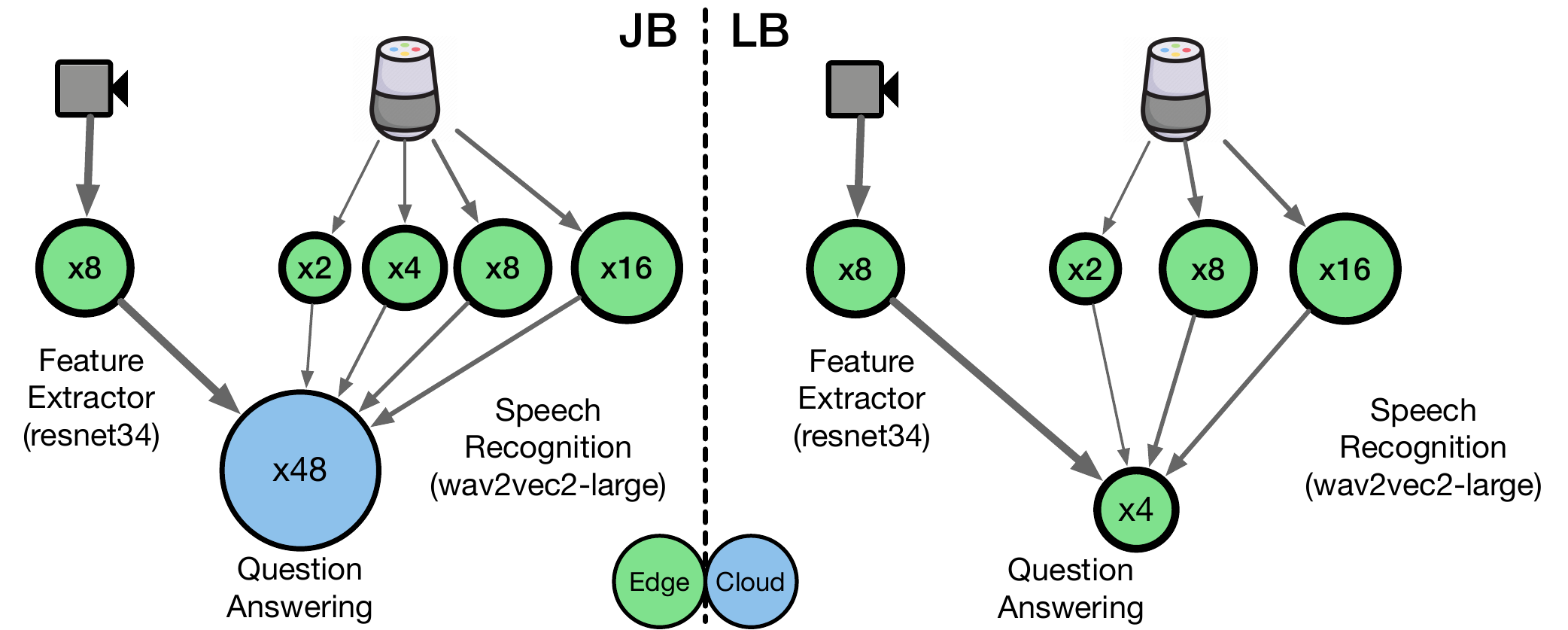}
\caption{Comparison of the execution plans of JB and LB on VQA using the \texttt{medium} setup modified to 10 rps.}
\label{fig:case_3}
\end{figure}

\subsubsection{Failure cases in \name}

Here we discuss properties of workflow scenarios and infrastructure configurations that can lead to our greedy strategies falling short and thus delivering less-than-optimal execution plans.

First, in some workflows, downstream operators may require less compute than upstream operators (i.e., those closer to the data sources). If the communication cost is not prohibitive, either as a result of the infrastructure cost or the bandwidth requirements, placing the upstream operators on the cloud and downstream operators on the edge may make sense. However, due to our Assumption A2, this placement is prohibited as we assume that information only flows in one direction from lower to higher tiers.
Given this assumption, workers on higher tiers should be assigned conservatively as their assignment restricts possible assignments for downstream operators in the workflow.
Our greedy strategy shares this philosophy, as it only considers higher tiers when the overall unit cost is cheaper (or when workers on lower tiers are exhausted).
In these cases, we find that removing A2 can lead to a reduction in total serving cost for the lower bound.
In Figure~\ref{fig:error_analysis_tput}, we analyzed the total serving cost of JellyBean (\texttt{JB}) compared to a lower bound without A2 (\texttt{LB/A2}) using the \texttt{medium} setting for VQA.
Comparing to Figure~\ref{fig:ab1-vqa}, we see a larger gap between \texttt{JB} and \texttt{LB}, as \texttt{JB} has up to 66\% higher cost than \texttt{LB} (among the cases that \texttt{LB} outputs different execution plan after removing Assumption A2).
We also compare the execution plans generated by \texttt{JB} and \texttt{LB}, when the assumption is removed. The plans on \texttt{medium} setting at 25 rps input throughput are shown in Figure~\ref{fig:case_4}. As we can see, \texttt{LB} places the QA operator backward using a x4 worker on edge.
Nevertheless, Assumption A2 is still valid in most of the test cases in our main experiments.

Second, in some edge cases, larger models for an operator may not necessarily lead to accuracy improvements. 
Due to the constant expansion factor in our model selection beam search, \texttt{JB} may fail to discover the model with accuracy right above the user-specified threshold.
We explore the minimal accuracy of feasible model selections made by \texttt{JB} on AICity in Figure~\ref{fig:error_analysis_ms}; we find that in the AICity workflow, the largest ReID model variant's accuracy is consistently lower than some smaller variants.

Third, in some infrastructure setups and workflows, smaller workers may also have lower unit compute cost.  \texttt{JB} may then over-provision workers to operators, as it assigns workers by greedily picking the one with the lowest unit cost. 
In Figure~\ref{fig:cost_zoomed}, we zoom in on Figure~\ref{fig:ab1} to focus on cases where \texttt{JB} has a higher cost than the lower bound (which are less obvious in the original figure).
We found that in these scenarios, \texttt{JB} over-provisions workers to some nodes due to its greedy strategy of iterative  worker assignments.
A visualization of the assignment in Figure~\ref{fig:case_3} shows that
\texttt{JB} uses one more worker than necessary for speech recognition, since it assigns the x2, x4, and x8 workers before the x16 (thus forcing the question answering operator to execute on the cloud).
For CPU workers, this is due to the fact that while the unit cost scales directly with increasing number of cores (i.e., x4 is half the cost of x8), the speedup in execution time does not.

There are some other uncommon scenarios in which \texttt{JB} could generate sub-optimal execution plans. For instance, there might be some workflows where an operator's output accuracy may not monotonically increase with the input accuracy. Under this circumstance, the beam-search strategy in model assignment may fail to discover feasible assignments. Some models may not have stable performance ranking across different workers (e.g., a model executes the fastest compared to other variants on GPU, while being the slowest on CPU); in this case, using a worker-agnostic cost in the model assignment is not sufficient.

\end{document}